\documentclass[lettersize,journal]{IEEEtran}
\usepackage[dvipsnames]{xcolor}
\usepackage{colortbl}
\usepackage{amsmath,amsfonts}
\usepackage{algorithmic}
\usepackage{algorithm}
\usepackage{array}

\usepackage{caption,subcaption}

\usepackage{tabularray}
\usepackage{nicematrix}

\usepackage{textcomp}
\usepackage{stfloats}
\usepackage{url}
\usepackage{verbatim}
\usepackage{graphicx}
\usepackage{cite}

\usepackage{mfirstuc}
\usepackage{hyperref}
\usepackage{url}
\usepackage{todonotes}
\usepackage{tikz}
\usepackage{pgfplots}
\pgfplotsset{compat=1.17}

\usepackage{pgfplotstable}
\usepackage{pgffor} 

\usepackage{filecontents}
\usepgfplotslibrary{fillbetween}
\usepackage{tcolorbox}
\usepackage{fontawesome}

\usepgfplotslibrary{groupplots}
\usepgfplotslibrary{colormaps}%
\usepgfplotslibrary{colorbrewer}

\pgfplotsset{compat=1.17}

\pgfmathdeclarefunction{lg10}{1}{%
    \pgfmathparse{ln(#1)/ln(10)}%
}

\pgfplotsset{
    colormap/bluecolormap/.style={colormap={bluecolormap}{
            rgb255=(255,255,255);
            rgb255=(192,192,255));
            rgb255=(0,0,255);
            rgb255=(0,0,0);
        }
    },
    colormap/redcolormap/.style={colormap={redcolormap}{
            rgb255=(255,255,255);
            rgb255=(255,192,192);
            rgb255=(255,0,0);
            rgb255=(0,0,102);
        }
    },
}

\usepgfplotslibrary{external}
\tikzstyle{mybox} = [draw=black, fill=black!5, thick,
    rectangle, rounded corners, inner sep=10pt, inner ysep=10pt]
\tikzstyle{fancytitle} =[fill=black, text=white]

\usepackage{csvsimple}

\usetikzlibrary{calc}
\usetikzlibrary{pgfplots.groupplots}

\usepackage{multirow}
\usepackage{booktabs}
\usepackage{tabularx}

\usepackage{siunitx}
\usepackage{wrapfig}

\usepackage{tkz-kiviat} 
\usetikzlibrary{arrows}
\usepackage[acronym]{glossaries}
\makeglossaries
\newacronym{DE}{DE}{deep ensemble}
\newacronym{RDE}{RDE}{repulsive deep ensemble}

\newacronym{POVI}{POVI}{particle-optimization variational inference}
\newacronym{BNN}{BNN}{bayesian neural network}
\newacronym{NN}{NN}{neural network}

\newacronym{AU}{AU}{aleatoric uncertainty}
\newacronym{EU}{EU}{epistemic uncertainty}
\newacronym{PU}{PU}{predictive uncertainty}
\newacronym{TU}{TU}{total uncertainty}

\newacronym{PDE}{PDE}{partial differential equation}
\newacronym{SVGD}{SVGD}{Stein variational gradient descent}
\newacronym{KDE}{KDE}{kernel density estimation}
\newacronym{MAP}{MAP}{maximum a posteriori}

\newacronym{fLLPOVI}{fs-RLL-E}{repulsive last-layer ensemble in function-space}
\newacronym{RLLPOVI}{RLL-E}{repulsive last-layer ensemble}
\newacronym{LLPOVI}{LL-E}{last-layer ensemble}
\newacronym{fs}{fs}{function space}

\newacronym{OOD}{OOD}{out-of-distribution}
\newacronym{ID}{ID}{in-distribution}
\newacronym{DDU}{DDU}{Deep Deterministic Uncertainties}
\newacronym{SNGP}{SNGP}{Spectral-normalized Neural Gaussian Processes}

\usepackage{soul}
\usepackage{ulem}
\usepackage{pifont}
\newcommand{\cmark}{\ding{51}}%
\newcommand{\xmark}{\ding{55}}%

\usepackage{amsmath,amsfonts,bm}

\def\eqref#1{equation~\ref{#1}}

\def\1{\bm{1}}
\newcommand{\train}{\mathcal{D}}

\newcommand{\repset}{\mathcal{D_{\mathrm{rep}}}}

\def\rvx{{\mathbf{x}}}
\def\rvy{{\mathbf{y}}}

\def\ervv{{\textnormal{v}}}

\def\rmX{{\mathbf{X}}}
\def\rmY{{\mathbf{Y}}}

\DeclareMathAlphabet{\mathsfit}{\encodingdefault}{\sfdefault}{m}{sl}
\SetMathAlphabet{\mathsfit}{bold}{\encodingdefault}{\sfdefault}{bx}{n}

\def\gD{{\mathcal{D}}}

\def\gX{{\mathcal{X}}}
\def\gY{{\mathcal{Y}}}

\def\sH{{\mathbb{H}}}
\def\sI{{\mathbb{I}}}

\newcommand{\E}{\mathbb{E}}

\newcommand{\KL}{D_{\mathrm{KL}}}

\newcommand{\rep}{\rvx_{rep}}

\usepackage[nameinlink,capitalize]{cleveref}
\usepackage{booktabs}
\usepackage{multirow}
\usepackage{graphicx}
\newcommand{\underlineitalic}[1]{\underline{#1}}

\begin{document}
\MFUnocap{the}\MFUnocap{and}\MFUnocap{in}\MFUnocap{for}\MFUnocap{are}\MFUnocap{via}

\title{\capitalisewords{Last layer repulsive ensembles: Function space diversity for improved uncertainty quantification}}

\title{\capitalisewords{Repulsive Last-Layer ensembles: \\ Uncertainty via function space diversity}}
\title{Function Space Diversity for Uncertainty Prediction via Repulsive Last-Layer Ensembles}

\author{Sophie Steger, 
Christian Knoll, 
Bernhard Klein, 
Holger Fr\"oning, 
Franz Pernkopf
\thanks{Sophie Steger is with the Signal Processing and Speech Communication Laboratory, Graz University of Technology, Austria, E-mail: sophie.steger@tugraz.at}%
\thanks{Franz Pernkopf is with the Signal Processing and Speech Communication Laboratory, Graz University of Technology, Austria, and Christian Doppler Laboratory for Dependable Intelligent Systems in Harsh Environments, Graz, Austria, E-mail: pernkopf@tugraz.at}%
\thanks{Christian Knoll is with Levata, Graz, Austria, knoll@levata.at}
\thanks{Bernhard Klein and Holger Fr\"oning are with the Institute of Computer Engineering (ZITI), Heidelberg University, Germany}
}

\maketitle

\begin{abstract}
Bayesian inference in function space has gained attention due to its robustness against overparameterization in neural networks. However, approximating the infinite-dimensional function space introduces several challenges. In this work, we discuss function space inference via particle optimization and present practical modifications that improve uncertainty estimation and, most importantly, make it applicable for large and pretrained networks. First, we demonstrate that the input samples, where particle predictions are enforced to be diverse, are detrimental to the model performance. While diversity on training data itself can lead to underfitting, the use of label-destroying data augmentation, or unlabeled out-of-distribution data can improve prediction diversity and uncertainty estimates.
Furthermore, we take advantage of the function space formulation, which imposes no restrictions on network parameterization other than sufficient flexibility. Instead of using full deep ensembles to represent particles, we propose a single multi-headed network that introduces a minimal increase in parameters and computation. This allows seamless integration to pretrained networks, where this repulsive last-layer ensemble can be used for uncertainty aware fine-tuning at minimal additional cost. We achieve competitive results in disentangling aleatoric and epistemic uncertainty for active learning, detecting out-of-domain data, and providing calibrated uncertainty estimates under distribution shifts with minimal compute and memory.
\end{abstract}


\glsresetall
\section{Introduction}

\IEEEPARstart{E}{pistemic}
uncertainty is often estimated by \glspl{DE} \cite{lakshminarayanan2017simple}. 
For the estimate to be accurate, each ensemble member must entail a sufficiently different optimum of the posterior distribution.
\Gls{POVI} \cite{liu2016stein, liu2017stein, liu2019understanding} achieves this diversity among ensemble members (i.e., the particles) by incorporating a repulsion term during parameter optimization.
However, parameter diversity alone is insufficient.
This is because neural networks with different parameters can still represent similar functions.
It is thus advisable to circumvent this issue by performing inference directly in the function space, enforcing diversity therein \cite{wang2019function, dangelo2021repulsive}. 
Yet -- despite its theoretical appeal – function space \gls{POVI} often performs worse than standard \glspl{DE} in both accuracy and quality of uncertainty estimation \cite{dangelo2021repulsive, trinh2023input, yashima2022feature}.
In this work, we reveal the underlying reason for this surprising performance gap; it is not because of the function space diversity but because of the challenges in accurately approximating the infinite-dimensional function space.

\IEEEpubidadjcol 

It remains practically infeasible to achieve function space diversity over the whole input domain (particularly for high dimensional input data).
Therefore, good {repulsion samples} must not only be \emph{diverse} but also capture the most \emph{relevant parts} of the input domain.
The training data itself is generally not rich enough and, as such, insufficient for accurate uncertainty estimation \cite{dangelo2021repulsive,trinh2023input}.
We demonstrate how to improve upon this without sacrificing accuracy: 
the key is to utilize unlabeled  \gls{OOD} data.
If \gls{OOD} data is unavailable, label-destroying data augmentation\footnote{Modification of input samples such that the original labels do not apply, e.g., shuffling of random image patches.} achieves similar quality.
Our evaluation confirms that repulsion samples satisfying both properties successfully suppress spurious features, improve uncertainty estimation, and achieve reliable \gls{OOD} detection.

\begin{figure}[t!]
    \centering
    \includegraphics[width=1\linewidth]{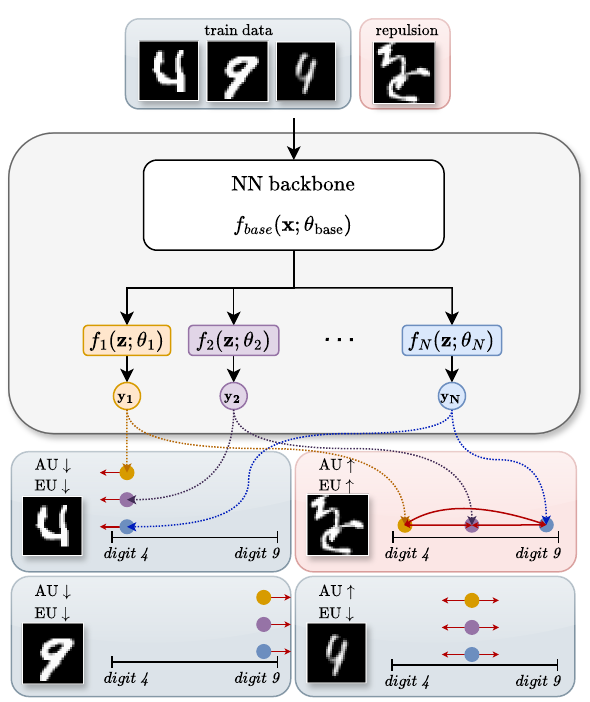}
    \caption{\Acrfull{fLLPOVI}, with $N$ particles. Colored dots correspond to the prediction of a particle. Unlabeled data points from a different distribution are used as repulsion samples for the function space repulsion loss. Epistemic uncertainty (EU) is the lowest, when all particle predictions agree, and increases with the spread of the particles. 
    The aleatoric uncertainty (AU) increases with ambiguous samples, e.g. the digit on the lower right belonging to both classes, resulting in particle predictions centered in the probability region.
    }
    \label{fig:illustration}
\end{figure}

Training and storing multiple ensemble members requires substantial computational resources, especially since the entire ensemble must be optimized jointly to maintain diversity.
Fortunately, function space \Gls{POVI} comes with few restrictions.
Drawing inspiration from ensemble distillation \cite{tran2020hydra}, we propose a multi-headed architecture (see Fig. \ref{fig:illustration}).
That is, we first learn a single base network that we subsequently equip with multiple heads – each representing one particle. We then enforce function space diversity among the heads, thus drastically reducing computational demands.

Moreover, our multi-headed approach opens the door for seamless integration with pretrained networks and provides several benefits:
First, it eliminates the need to train all ensemble members from scratch;
Second, unlike classical DEs, which require diverse weight initializations and degrade in combination with pretrained models \cite{schweighofer2024quantification}, function space \Gls{POVI} circumvents this problem and can provide better quality uncertainty estimates.
Thus, our \gls{fLLPOVI} allows for uncertainty-aware fine-tuning of the model.\\

\paragraph*{Contributions}
In summary, we propose a simple and scalable uncertainty estimation method based on function space \gls{POVI} methods. 
\begin{itemize}
    \item We highlight the importance of the repulsion term in function space \gls{POVI} methods to accurately estimate epistemic uncertainty, especially for a limited number of particles (Section \ref{sec:finite_particles}, \ref{sec:where_diversity}).
    \item We propose specific approximations of the function space to improve the quality of uncertainty estimates and to reduce the computational overhead. Instead of using full \glspl{DE}, we utilize a \gls{RLLPOVI} to be parameter-effective (Section \ref{sec:approx1_para}). Diverse predictions are ensured by choosing an appropriate set of repulsion samples for the function space repulsion term (Section \ref{sec:approx2_repulsion}). 
    \item We show that our method retrospectively provides uncertainty estimates for pretrained models. If the pretrained network is regularized to avoid feature collapse, it is sufficient to retrain the \gls{fLLPOVI} with function space repulsion (Section \ref{sec:approx3_finetuning}).
    \item We empirically evaluate our method for regression and classification tasks on synthetic and real-world datasets. Our method can (i) disentangle aleatoric and epistemic uncertainty for active learning, (ii) improve detection of both near and far \gls{OOD} data, and (iii) provide calibrated uncertainty estimates under distribution shifts (Section \ref{sec:experiments}). 
\end{itemize}

\section{Preliminaries}

We consider supervised learning tasks. Let $\gD=\{ \rvx_i, \rvy_i \}_{i=1}^N=(\rmX, \rmY)$ denote the training data set consisting of $N$ i.i.d. data samples with inputs $\rvx_i \in \gX$ and targets $\rvy_i \in \gY$.
We define a likelihood model $p(\rvy|\rvx,\theta)$ with the mapping $f(\cdot; \theta) : \gX \to \mathbb{R}^K$ parameterized by a \gls{NN}.

\subsection{Bayesian neural networks}

The core concept of \Glspl{BNN} involves the treatment of network parameters $\theta$ as random variables instead of point estimates.
This entails defining a prior distribution of the parameters $p(\theta)$ to infer the posterior distribution of the parameters $p(\theta | \train) \propto p(\theta) p(\rmY|\rmX,\theta)$. Predictions for a test data point $\rvx_{\text{test}}$ are obtained by marginalizing over all possible parameters: $p(\rvy_{\text{test}} | \rvx_{\text{test}}, \train) = \int_\theta p(\rvy_{\text{test}} | \rvx_{\text{test}}, \theta) p(\theta|\train) d\theta$. 
This integral, however, is generally intractable. 

\subsection{\Acrlong{POVI}} 
Variational inference approximates the posterior $p(\theta | \train)$ by a simpler parametric distribution $q(\theta)$. 
\Gls{POVI} methods \cite{liu2016stein, chen2018unified} aim to provide more flexibility by considering a non-parametric distribution, specified by a discrete set of particles $\{ \theta^{(i)} \}_{i=1}^n$ according to
$q(\theta) \approx \frac{1}{n} \sum_{i=1}^n \delta(\theta-\theta^{(i)}),$
where $\delta(\cdot)$ is the Dirac function. The particles can then be optimized iteratively via
\begin{align}
    \theta_{l+1}^{(i)} \leftarrow \theta_{l}^{(i)} + \epsilon_l \ervv(\theta_l^{(i)})
\end{align}
where $\epsilon_l$ is the step size at time step $l$. 
By viewing the particle optimization as a gradient flow in Wasserstein space, \cite{dangelo2021repulsive} derive the following update rule that decomposes into an attraction and repulsion term
\begin{align}
\begin{split}
\ervv(\theta_l^{(i)}) = &~ \underbrace{\nabla_{\theta_l^{(i)}} \log p(\theta_l^{(i)} | \train)}_{\textsc{attraction}} \\
&-\underbrace{\frac{\sum_{j=1}^n \nabla_{\theta_l^{(i)}} k\left(\theta_l^{(i)}, \theta_l^{(j)}\right)}{\sum_{j=1}^n k\left(\theta_l^{(i)}, \theta_l^{(j)}\right)}}_{\textsc{repulsion}} \label{eq:kdeWGD}
\end{split}
\end{align}
where $k(\cdot, \cdot)$ denotes a kernel function.  
The attraction term drives particles into high-density regions of the posterior distribution, while the repulsion term induces diversity by preventing particles from collapsing into the same optimum.        
For a single particle, this training procedure reduces to \gls{MAP} training;
for $n\to\infty$ and a properly defined kernel, it converges to the true posterior distribution \cite{dangelo2021repulsive}.

\section{Why repulsion matters for a finite number of particles} \label{sec:finite_particles}
Importantly, the guarantee of convergence to the posterior distribution $p(\theta| \train)$ is only valid in the limit of an infinite number of particles. Although fascinating from a theoretical perspective, practical importance lies in the analysis of the behavior for a finite number of particles. 
In this section, we discuss the importance of the repulsion term in the regime where the number of particles is significantly smaller than the number of local minima. \\

\paragraph*{Estimating epistemic uncertainty}
Typically, the disagreement between the predictions of ensemble members is used for uncertainty estimation. Following \cite{depeweg2018decomposition,wimmer2023quantifying,schweighofer2024quantification}, predictive uncertainty can be decomposed into aleatoric and epistemic components, represented as conditional entropy and mutual information, respectively:
\begin{align}
\begin{split}
   \underbrace{\sH [ \E_{p(\theta| \train)} [ p( \rvy | \rvx , \theta) ] ]}_{\textit{Total}} &=
    \underbrace{\E_{p(\theta| \train)} [ \sH [ p( \rvy | \rvx , \theta) ] ]}_{\textit{Aleatoric}} \\ 
   &~~~+ \underbrace{\sI [ \rvy; \theta | \rvx, \train]}_{\textit{Epistemic}}
\end{split} \label{eq:UQ_decomp}
\end{align}

The total uncertainty is given by the entropy of the model's predictions, aleatoric uncertainty represents the variability in outcomes due to inherent randomness in the data, and epistemic uncertainty reflects our lack of knowledge about which model generated the data.
If a test sample $\rvx$ can be explained by many disagreeing models $p( \rvy | \rvx , \theta)$, each plausible under the posterior distribution $p(\theta | \train)$, epistemic uncertainty is high.
By acquiring additional training data close to $\rvx$, the space of plausible models and thus inconsistent predictions is decreased. \\

\paragraph*{Finite particle approximation}
If the posterior distribution is approximated by a finite set of discrete particles $\theta^{(i)}$, epistemic uncertainty estimation simplifies to a Monte Carlo integration \cite{wimmer2023quantifying}:
\begin{align}
\begin{split}
    & \sI [ \rvy; \theta | \rvx, \train] \approx  \\ \textstyle
   &\frac{1}{n} \sum_{i=1}^n \KL \Big( p\big( \rvy | \rvx , \theta^{(i)}\big) \Big\vert \Big\vert  \frac{1}{n} \sum_{j=1}^n p\left( \rvy | \rvx , \theta^{(j)}\right) \Big) 
\end{split} \label{eq:EU_KL}
\end{align}

Given practical constraints on the number of particles (typically five to ten), many posterior modes remain unexplored and the estimate of the epistemic uncertainty is shaped largely by a small number of posterior modes \cite{wimmer2023quantifying}. This limitation stresses the need for guiding particles towards representative and \textit{diverse posterior modes} to avoid underestimation of epistemic uncertainty. 
\\

\paragraph*{Repulsion in deep ensembles}
Deep ensembles can be viewed as an unregularized case of \cref{eq:kdeWGD}, lacking a repulsion term. Particles move according to the gradient flow towards high-density posterior modes $p(\theta_l^{(j)}|\train)$, with diversity stemming from their random initial positions $\theta_{l=0}^{(j)}$ in the loss landscape. 
Recent research has raised concerns about the effectiveness of this approach in achieving diverse posterior modes. The loss landscape, heavily influenced by input features correlated with the target, may render diverse posterior modes inaccessible to the unregularized gradient flow \cite{schweighofer2024quantification}.
In addition, empirical evidence has demonstrated that the epistemic uncertainty provided by \glspl{DE} does not reliably identify distribution shifts. In several cases, the aleatoric uncertainty of a single model has been more effective in detecting \gls{OOD} data \cite{schweighofer2024quantification,xia2022usefulness}. 

\Gls{POVI} methods introduce a repulsion kernel, $k\left(\theta^{(i)}, \theta^{(j)}\right)$, to prevent particles from converging to identical posterior modes. 
Theoretically, when the number of particles approaches infinity, this repulsion mechanism ensures convergence to the posterior distribution \cite{dangelo2021repulsive, wild2024rigorous}. 
In practical applications, where the number of particles is finite and vastly smaller than the number of local optima, studies show that random initialization is sufficient to prevent the particles from collapsing into the same local optimum \cite{wild2024rigorous}. Still, it is not guaranteed that those distinct local optima result in diverse prediction functions. Consequently, the repulsion kernel needs to actively direct particles towards such diverse posterior modes.\\

In the following sections, we discuss current practices of employing \Gls{POVI} methods. 
We propose the following desiderata for the repulsion term in order to achieve reliable epistemic uncertainty estimates and practical applicability. \\
\begin{itemize}
    \item[D1] \textit{The repulsion term should steer particles towards diverse posterior modes, which provide a useful approximation for the epistemic uncertainty in \cref{eq:EU_KL}. } \\
    \item[D2] \textit{Particles should reach diverse posterior modes from the same initial parameters through the use of the repulsion term. This enables the fine-tuning of pretrained models to better approximate epistemic uncertainty.}
\end{itemize}

\section{Where should we enforce diversity?} \label{sec:where_diversity}

\subsection{Diversity of network parameters}
Traditionally, \glspl{BNN}  have focused on inferring the distribution over network parameters $\theta$. However, enforcing repulsion directly in the parameter space presents several practical challenges, particularly in defining a useful distance metric. Previous work utilized Euclidean ($\ell 2$ norm) or Manhattan ($\ell 2$ norm) repulsive kernels \cite{wang2019function, dangelo2021repulsive}:
$$k\left(\theta^{(i)}, \theta^{(j)}\right) = \exp\left(-\frac{\|\theta^{(i)} - \theta^{(j)}\|_p}{\nu} \right)$$
However, this regularization term does not ensure \textit{diversity of prediction functions} as stated in desideratum (D1). Due to the overparameterized nature of \glspl{NN}, different network configurations can yield identical prediction functions. Two notable invariances in \glspl{NN} are the permutation of hidden nodes (permutation symmetry) and the arbitrary scaling afforded by ReLU activations (scaling symmetry) \cite{pourzanjani2017identifiability}. 
In the extreme case, particles can settle into local minima that are distant under the considered kernel, but yield functionally equivalent predictions. 
Moreover, the immense number of \gls{NN} parameters, ranging from thousands to billions, complicates the use of distance-based kernels in parameter space. In such high-dimensional spaces, traditional measures of proximity and similarity become less effective \cite{aggarwal2001surprising}.

\subsection{Diversity of prediction functions}
To effectively capture the diversity of particle predictions, it is crucial for the distance kernel to be informative about the variations in these predictions. This objective can be achieved by conducting inference directly in the function space and incorporating these predictions into the kernel function itself \cite{wang2019function, dangelo2021repulsive}. 
By reformulating posterior inference in the space of prediction functions, the $n$ particles represent functions $f^{(1)}(\gX), \dots, f^{(n)}(\gX)$, updated as follows:
\begin{align}
    f_{l+1}^{(i)}(\gX) \leftarrow f_{l}^{(i)}(\gX) + \epsilon_l \ervv(f_l^{(i)}(\gX)). 
\end{align}
Performing inference in the function space directly enforces diverse prediction functions, and addresses issues arising with overparameterization. 
Optimizing the objective in function space ensures that the particles converge towards diverse posterior modes that cover large values of the KL-divergence in \cref{eq:EU_KL}. Through
\begin{align*}
\begin{split}
  k\Big( f^{(i)}(\gX)&, f^{(j)}(\gX) \Big) =  \\
  &\exp\left(-\frac{\|f^{(i)}(\gX) - f^{(j)}(\gX)\|_p}{\nu} \right)
\end{split} \label{eq:fs-kernel}
\end{align*}
we enforce diverse predictions and our desideratum (D1) is fullfilled. 
However, to solve the problem we must rely on gradient based optimization procedures that in turn require a parameterized representation of the particles. \\

\subsubsection{Function parameterization}
Each particle $f^{(i)}(\gX)$ is represented by a specific parameterization $f^{(i)}(\gX;\theta^{(i)})$. The parameterization $f^{(i)}(\gX;\theta^{(i)})$ must be sufficiently flexible to effectively approximate the underlying function space \cite{wang2019function}. \\

\subsubsection{Repulsion samples} 
Moreover, it remains prohibitive to evaluate $f^{(i)}(\gX;\theta^{(i)})$ across the entire input domain $\gX$. Instead, prior work \cite{wang2019function} adopted a mini-batch approximation, where the evaluation over the full set $\gX$ is replaced with $B$ \emph{repulsion samples} drawn from an arbitrary distribution $\rep \sim \mu$ with support on $\gX^B$. The variational distribution is shown to converge to the true posterior if the posterior can be determined by almost all $B$-dimensional marginals
$\{p(f(\mathbf{x}) | \mathbf{X}, \mathbf{Y}): \mathbf{x} \in \operatorname{supp}(\mu)\}$ \cite{wang2019function}. 

\section{Improving function space approximations: Practical choices and implications}
\label{sec:my_method}

The function space formulation effectively circumvents issues regarding overparameterization and identifiability \cite{kirsch2024bridging}. It directly enforces local minima that correspond to diverse functional predictions, mitigating underestimation of uncertainty. Still, empirical results lack improvement over unregularized \glspl{DE}, especially for large scale image tasks \cite{wang2019function, dangelo2021repulsive, trinh2023input}. In the following section, we show that these results are expected given previous choices for the approximations of the function space. We provide practical improvements and justifications for our choices, demonstrating how we can effectively meet both outlined desiderata D1 and D2.
\subsection{Choice of function parameterization} \label{sec:approx1_para}

A key benefit of function space inference entails the theoretical justification to use any flexible network parameterization. 
Still, prior work has utilized the same \gls{DE} structure, where each particle is parameterized by a separate neural network \cite{wang2019function, dangelo2021repulsive}. This choice limits the number of particles in large scale problems, where it might be difficult to train and store several networks. Additionally, due to the interaction of all particles through the repulsion kernel, the training procedure is further complicated by requiring parallel training. Empirically, it has not been analyzed how parameter-efficient network structures perform as alternative parametric approximations. Thus, we propose to use a shared base network with multiple heads that represent the particles in function space, i.e.
$$f^{(i)}(\mathbf{x}; \theta_{\text{base}}, \theta_{\text{head}}^{(i)}) = f_{\text{head}}^{(i)}(f_{\text{base}}(\mathbf{x}; \theta_{\text{base}}); \theta_{\text{head}}^{(i)}).$$
By sharing the latent representation of the base model $f_{\text{base}}(\mathbf{x}; \theta_{\text{base}})$, our model is highly parameter-effective. We demonstrate in several experiments that an ensemble of linear layer is sufficient to improve uncertainty estimates of a single network.  \\

\paragraph*{Justification}
Multi-headed network architectures have been used successfully to distill \glspl{DE} and replicate their functional behavior \cite{tran2020hydra}, demonstrating sufficient flexibility of single networks \cite{hinton2015distilling}. Performing particle optimization in function space mitigates the need for training a full \gls{DE} prior to distillation. 
The use of a shared deterministic base network aligns with partially stochastic \glspl{BNN}, where a subnetwork of the parameters is treated probabilistically. Most prominently, Bayesian last-layer networks are employed as practical means to reduce computational demands \cite{sharma2023bayesian}. 

\tikzexternaldisable
\vspace*{.1cm}
\begin{tikzpicture}
\node [mybox] (box){%
\begin{minipage}{0.4\textwidth}
    \textit{Deep ensembles are not necessary: A single neural network with multiple heads is sufficiently flexible to provide diverse predictions for uncertainty estimation.}
\end{minipage}
};
\end{tikzpicture}%
\tikzexternalenable

\subsection{Choice of repulsion evaluation samples}  \label{sec:approx2_repulsion}

\begin{figure}[t!]
     \centering
     \begin{subfigure}[b]{0.22\columnwidth}
         \centering
         \includegraphics[width=1\columnwidth]{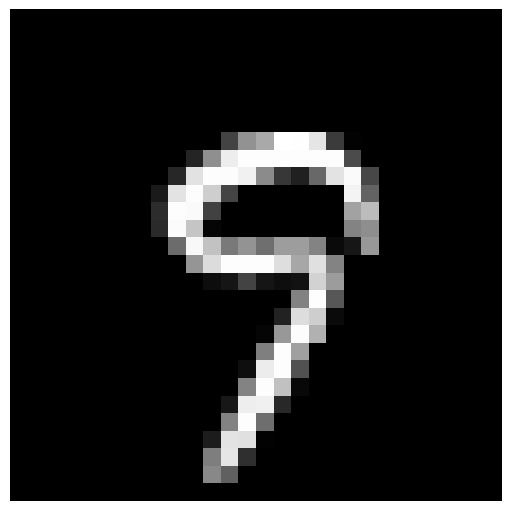}
         \caption*{MNIST}
         \label{fig:y equals x}
     \end{subfigure}
     \begin{subfigure}[b]{0.22\columnwidth}
         \centering
         \includegraphics[width=1\columnwidth]{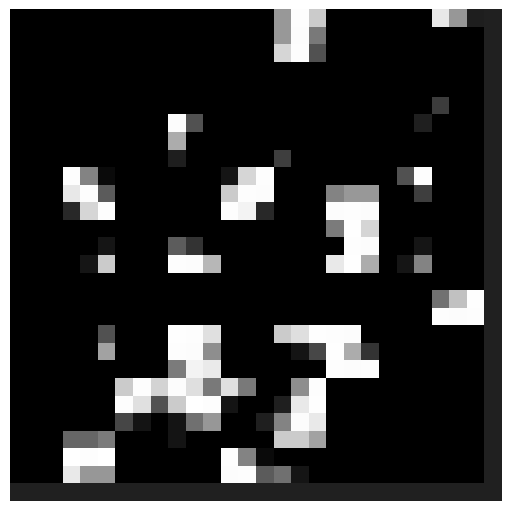}
         \caption*{Patches-8}
         \label{fig:three sin x}
     \end{subfigure}
     \begin{subfigure}[b]{0.22\columnwidth}
         \centering
         \includegraphics[width=1\columnwidth]{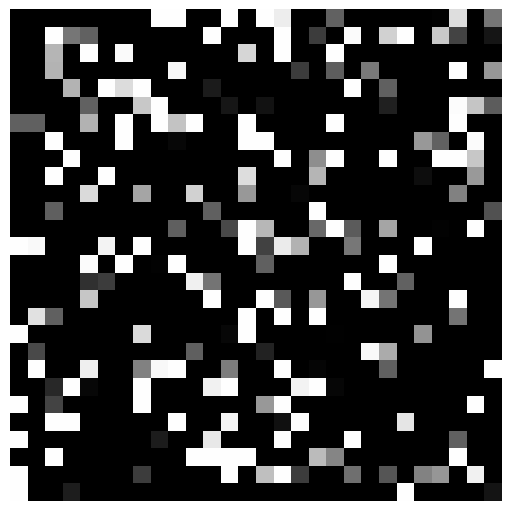}
         \caption*{Patches-16}
         \label{fig:five over x}
     \end{subfigure}
     \begin{subfigure}[b]{0.22\columnwidth}
         \centering
         \includegraphics[width=1\columnwidth]{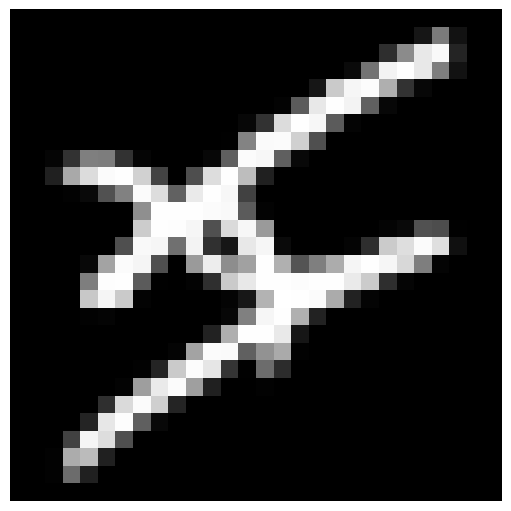}
         \caption*{eMNIST}
         \label{fig:five over x}
     \end{subfigure}

    \begin{subfigure}[b]{0.22\columnwidth}
         \centering
         \includegraphics[width=1\columnwidth]{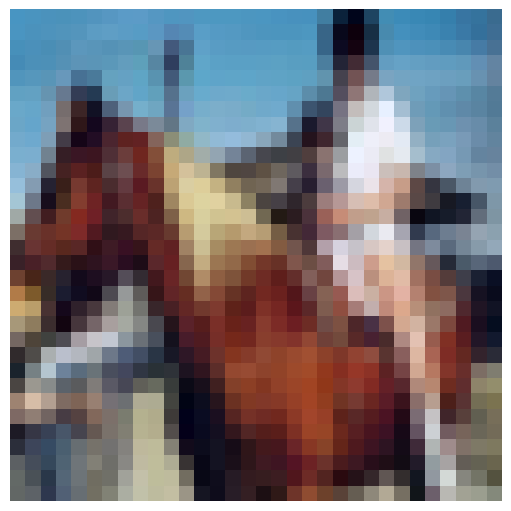}
         \caption*{CIFAR10}
         \label{fig:y equals x}
     \end{subfigure}
     \begin{subfigure}[b]{0.22\columnwidth}
         \centering
         \includegraphics[width=1\columnwidth]{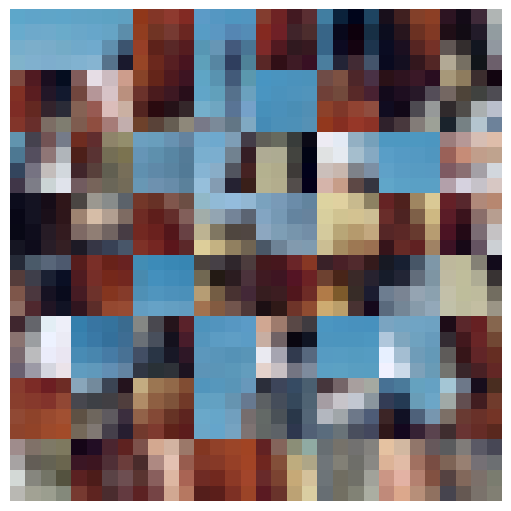}
         \caption*{Patches-8}
         \label{fig:three sin x}
     \end{subfigure}
     \begin{subfigure}[b]{0.22\columnwidth}
         \centering
         \includegraphics[width=1\columnwidth]{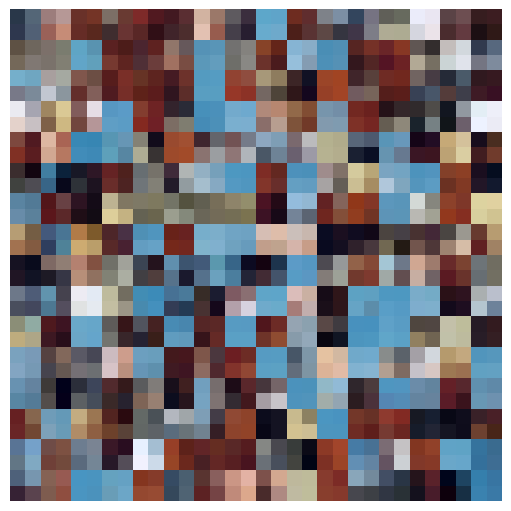}
         \caption*{Patches-16}
         \label{fig:five over x}
     \end{subfigure}
     \begin{subfigure}[b]{0.22\columnwidth}
         \centering
         \includegraphics[width=1\columnwidth]{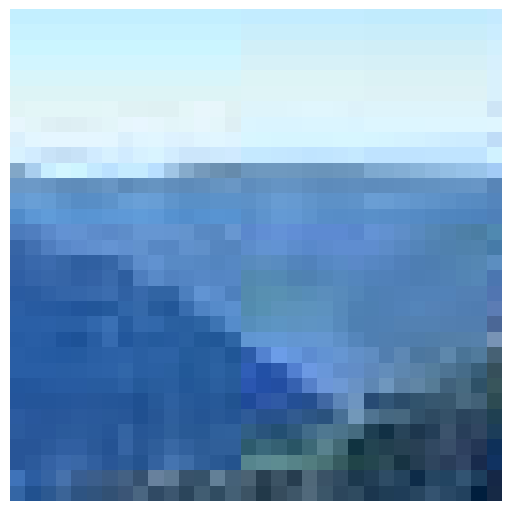}
         \caption*{TinyImagenet}
         \label{fig:five over x}
     \end{subfigure}
        \caption{Example of repulsion samples for DirtyMNIST (top row) and CIFAR10/100 (bottom row).}
        \label{fig:repulsion}
\end{figure}

Evaluation of the function-space repulsion term requires selecting a set of \emph{repulsion samples} $\rep \in \repset$. 
This choice significantly impacts the valid input domain for uncertainty estimates of fs-\gls{POVI} methods.
Prior work proposed to draw repulsion samples from the kernel density estimate over the training data \cite{wang2019function}, or take samples from the training data directly \cite{dangelo2021repulsive}.
This choice, however, limits the applicability of the \gls{BNN} approximation to situations where samples are coming from the same distribution as the training data. 
Selecting good repulsion samples becomes particularly challenging for high-dimensional spaces, for which drawing random samples from the entire input domain is simply infeasible.
Instead, one must restrict the selection to an informative subset that covers the domain of interest from the input space.
In image classification, this often includes natural images from varying distributions. 
We can thus exploit the abundance of available unlabeled image data. For example, using {\mbox{eMNIST}} as repulsion samples for models trained on  MNIST, or TinyImagenet for models trained on CIFAR10/100, leverages natural variability across different image sets.
If unlabeled OOD data is unavailable, repulsion samples can be generated from the training data by label-destroying data augmentation techniques.
One such effective method is the random shuffling of image patches to destroy the shape information of objects that is crucial for human perception (see Fig.~\ref{fig:repulsion}). \\

\paragraph*{Justification}
Enforcing diversity directly on the training data has been shown to degrade performance by artificially inflating epistemic uncertainty at data points where independent training would yield confident predictions \cite{abe2022best, jeffares2024joint}. 
This approach often fails to detect \gls{OOD} data, which may be characterized by spurious features present in the training data or features that are completely absent in the training set.
Using unlabeled \gls{OOD} data as repulsion samples provides an effective solution to these challenges. These samples may contain features that are problematic or absent in the training data, allowing the models to meaningfully enforce diversity and improve \gls{OOD} detection capabilities. 
Similarly, label-destroying data augmentation mitigates robust features that are indicative of the class label. 
By promoting diversity on those augmented images, it can prevent the model from depending on features that are not truly indicative of the class, thus mitigating the risk of overconfident and erroneous predictions. 
Compared to methods that rely on feature density to detect \gls{OOD} data, repulsion samples offer the benefit of learning to ignore spurious features that may be present in the training data. 

\tikzexternaldisable
\vspace*{.5cm}
\begin{tikzpicture}
\node [mybox] (box){%
\begin{minipage}{0.4\textwidth}
    \textit{Encouraging diverse predictions on training data itself is not sufficient to improve epistemic uncertainty estimation -- we need random sampling, label-destroying data augmentation, or \gls{OOD} data as repulsion samples.    }
\end{minipage}
};
\end{tikzpicture}%
\tikzexternalenable

\subsection{Retrospective uncertainties for pretrained models} \label{sec:approx3_finetuning}
The multi-headed network approach provides a principled approach to computing retrospective uncertainties for a pretrained base network.
We replace the last layer of the base network with an ensemble of linear layers trained with the function space repulsion term. 
In this way, representation learning and function space inference are decoupled, allowing the computation of diverse decision boundaries while leveraging pre-learned representations. 
However, if the function space repulsion term is included post hoc, features indicative of \gls{OOD} data may not be extracted by the base network due to feature collapse, where data points far apart in input space collapse into indistinguishable parts in feature space \cite{van2020uncertainty}. 
Fortunately, pretrained deep neural networks are often trained with mechanisms that help to mitigate feature collapse. In the following, we discuss techniques commonly used in training deep networks that lead to feature space regularization of the learned representations. 
In Section \ref{sec:experiments}, we evaluate whether the last-layer retraining is sufficiently flexible to enforce diverse predictions on \gls{OOD} data. This directly addresses our second goal (D2) of obtaining diverse posterior modes from the same initial parameters.
\\

\paragraph*{Spectral normalization and residual connections}
Distance-aware representations can be achieved by imposing bi-Lipschitz constraints 
\begin{align*}
K_L d_I(\rvx_1, \rvx_2) \leq d_F(f_{\text{base}}(\rvx_1), &f_{\text{base}}(\rvx_2)) \\
&\leq K_U d_I(\rvx_1, \rvx_2) .
\end{align*}
Here, \(d_I\) and \(d_F\) represent distance measures in the input and feature spaces, while \(K_L\) and \(K_U\) are the lower and upper Lipschitz constants, respectively. 
These constraints enforce a bounded relationship between distances in the input (\(d_I\)) and feature (\(d_F\)) space. 
Models with constrained Lipschitz constants have demonstrated improved generalization and adversarial risk mitigation \cite{miyato2018spectral}.
Spectral normalization and residual connections serve as effective techniques to impose these constraints \cite{miyato2018spectral}. They prevent feature collapse while introducing smoothness (upper Lipschitz constant) in the feature space. While most pretrained models are not trained with spectral normalization, employing a network structure with residual connections alone often suffices to maintain distance awareness in feature space. 
\\

\paragraph*{Data augmentation}
Data augmentation is an important training technique that aims to enrich the data set by generating different variations of the original samples. Techniques such as Mixup \cite{zhang2017mixup} or CutMix \cite{yun2019cutmix} are often used for this purpose. 
They introduce variations by combining or interpolating between different samples, thereby expanding the model's exposure to a broader range of data distributions.
Suppose we train a classifier to distinguish between the digits "0" and "1". Initially, the model might rely on simple features, such as the presence or absence of a straight line, to make accurate predictions. While this approach may be sufficient for distinguishing between "0" and "1", it may struggle when faced with more complex tasks, such as identifying the digit "7" as \gls{OOD}. This difficulty stems from the model's limited exposure to diverse features during training.
Next, consider CutMix augmentation, where parts of different images are combined to create new synthetic samples. For example, by merging the top half of a "0" image with the bottom half of a "1" image, we create a synthetic sample that resembles a "7". Incorporating such augmented samples into the training data forces the model to learn more nuanced features that distinguish not only between "0" and "1", but also between "7" and the other digits. 
This process enriches the model's representation of the data and promotes the extraction of more diverse features.

\tikzexternaldisable
\vspace*{.5cm}
\begin{tikzpicture}
\node [mybox] (box){%
\begin{minipage}{0.4\textwidth}
    \textit{
    Pretrained neural networks are often trained with methods that avoid feature collapse. In this case we can decouple the problem into two stages: representation learning and uncertainty-aware fine-tuning using function-space inference.
    }
\end{minipage}
};
\end{tikzpicture}%
\tikzexternalenable

\section{Implementation and computational cost}
\paragraph*{Memory}
The feature space of a pretrained base network serves as the input space for our last-layer ensemble. For all image classification tasks, the parameters of the base network are frozen during the training of our \gls{fLLPOVI}. As we use an ensemble of linear layers, the number of trainable parameters is given by 
$N_{total} = (d \times K + K)\times n$, where $d$ is the dimension of the feature space of the base network, $K$ is the number of classes, and $n$ the number of particles, i.e. ensemble heads. As argued in Section \ref{sec:where_diversity}, function space repulsion is useful in the small particle regime to drive particles towards representative modes of the posterior. 
\paragraph*{Training and inference}
During each gradient step, we draw a batch of repulsion samples to obtain the particle predictions for the repulsion term in function space. During inference, uncertainty estimates are obtained from the disagreement of the particle predictions.

\section{Related Work}\label{sec:rel_work}

\subsection*{Scalable Bayesian Neural Networks}
\Acrlong{BNN} provide a principled way to quantify uncertainty in neural networks. 
However, the computational cost of training and inference is often prohibitive.
Gradient based Monte Carlo methods, such as Hamiltonian Monte Carlo \cite{neal1995bayesian}, are powerful tools for sampling from complex distributions. However, they are computationally expensive and require careful tuning of hyperparameters. Thus, much research has been invested in finding efficient methods to approximate the posterior distribution and to make \glspl{BNN} scalable, including variational inference \cite{blundell2015weight}, dropout as variational inference \cite{gal2016dropout}, and Laplace approximation \cite{daxberger2021laplace}.
Partially stochastic networks reduce the computational demands of \glspl{BNN} by treating only a subset of the parameters probabilistically \cite{daxberger2021laplace, sharma2023bayesian}. In particular, last-layer approaches have been shown to be effective in reducing overconfidence \cite{dusenberry2020efficient,kristiadi2020being, harrison2024variational}. Our multi-headed structure can be interpreted as a partially stochastic network where the last layer is trained using particle optimization in function space.

\subsection*{(Repulsive) Deep Ensembles}
Deep ensembles combine the predictions of several deep neural networks, where each network is initialized randomly and trained independently.
Originally considered an uncertainty heuristic, \glspl{DE} have been shown to outperform Bayesian methods in empirical evaluations regarding prediction accuracy, uncertainty calibration, and out-of-distribution detection \cite{lakshminarayanan2017simple, gustafsson2020evaluating, ovadia2019can}. 
Subsequently, there has been considerable research on \glspl{DE} and the conditions under which they be considered a Bayesian method \cite{wilson2020case, dangelo2021repulsive, wild2024rigorous}. 
Repulsive \glspl{DE} introduce a kernelized repulsion term that prevents ensemble members from collapsing to the same local optimum. They differ in the space in which diversity is enforced: network parameters \cite{wang2019function, dangelo2021repulsive}, feature representations \cite{yashima2022feature}, input gradients \cite{trinh2023input}, or function space \cite{wang2019function, dangelo2021repulsive}.

\subsection*{Function-space Inference}
A number of inference methods for \glspl{BNN} consider the shift from inference in the space of network parameters to the function space \cite{sun2019functional, ma2019variational, burt2020understanding, wang2019function, ma2021functional, rudner2022tractable}. This allows to specify meaningful prior distributions over the network parameters. 
Recent work proposed a tractable variational inference method by linearizing the function mapping of the neural network around a Gaussian distribution \cite{rudner2022tractable, rudner2023function}.
\gls{POVI} methods approximate the posterior distribution using a set of discrete particles to capture its multimodal structure \cite{wang2019function, dangelo2021repulsive}. 

\subsection*{Auxiliary out-of-distribution data}
Function space inference methods enforce the function prior on a set of input points, in some work referred to as measurement \cite{sun2019functional,wang2019function,ma2021functional} or context samples \cite{rudner2022tractable,rudner2023function}. In low-dimensional problems, such samples can be obtained by drawing from a distribution with support over the domain of interest \cite{sun2019functional,wang2019function,ma2021functional}. For high-dimensional problems with structured data, such as natural images, samples from an \gls{OOD} data set have shown improvements \cite{rudner2022tractable,rudner2023function}. Similar work on \gls{OOD} detection methods for single networks has used auxiliary \gls{OOD} datasets to maximize softmax entropy \cite{hendrycks2018deep}.

\subsection*{Multi-headed architectures} 
Various approaches have used multi-headed network architectures to reduce memory requirements by sharing parameters of a base network \cite{song2018collaborative, sercu2016very, lee2015m}. In reinforcement learning, bootstrapping using a multi-headed network has been employed to improve exploration tasks \cite{osband2016deep}. In addition, multi-headed networks were used to perform online distillation of a teacher model \cite{zhu2018knowledge}, and to replicate functional behavior of deep ensembles \cite{tran2020hydra}. The trade-off between ensembling the whole network and a selection of specific layers has been analyzed in \cite{valdenegro2023sub}.

\subsection*{Distance based uncertainty methods}

Distance-based uncertainty methods consider the epistemic uncertainty of a given test input to be proportional to the distance {to} the support of the training data. If a test sample is close to the training distribution, the predictions are considered trustworthy. If the distance is large, the model should abstain from making predictions. Computing distances directly in high-dimensional input spaces, however, is often impractical. Thus, most methods depend on well-informed latent representations of the network and estimate epistemic uncertainty by considering feature space densities \cite{charpentier2020posterior, postels2020quantifying, mukhoti2023deep, winkens2020contrastive} or distances \cite{liu2020simple, van2020uncertainty, tagasovska2019single}. 
This requires appropriate regularization of the feature space to avoid feature collapse and to ensure that densities and distances are meaningful \cite{van2021feature}.
Common methods to achieve bi-Lipschitz conditions include gradient penalties \cite{gulrajani2017improved, van2020uncertainty}, and spectral normalization \cite{miyato2018spectral, liu2020simple}. 
In extensive experiments, \cite{postels2021practicality} demonstrate that relying solely on the feature space density of a model is not sufficient to indicate correctness of a prediction and results in poor calibration under distribution shifts.

\section{Experiments}\label{sec:experiments}

\subsubsection*{Datasets} 
We evaluate our method on several benchmark experiments. We start with an illustrative evaluation on a synthetic regression and classification problem. 
Next, we test the effectiveness of disentangling aleatoric and epistemic uncertainty in an image classification task.
We perform experiments on the DirtyMNIST dataset \cite{mukhoti2023deep}, which consists of MNIST digits that belong to a unique class and artificial ambiguous digits with several class labels. Further, we use epistemic uncertainty estimates for an active learning task on the DirtyMNIST dataset. We test out-of-distribution (OOD) detection on CIFAR10/CIFAR100 \cite{Krizhevsky09learningmultiple} for both far (Places365 \cite{places365}, SVHN \cite{Netzer2011}, Texture \cite{texture}, FakeData\footnote{Randomly generated noise images.}) and near (CIFAR100, TinyImagenet \cite{le2015tiny}) OOD data. 
Finally, we evaluate calibration of uncertainty estimates under various synthetic corruptions for the CIFAR10-C/100-C datasets \cite{hendrycks2018benchmarking}. See supplementary material, Appendix A, for a summary of the training details. 

\subsubsection*{Uncertainty baselines} 
A single neural network (MAP) serves as the base model and backbone for all post-hoc uncertainty techniques. For unregularized \glspl{DE}, we retrain the base network 5 times with random initializations (DE-5). We have selected two baselines for deterministic uncertainty methods: \acrshort{DDU} \cite{mukhoti2023deep} and \acrshort{SNGP} \cite{liu2023simple}. As a representative of single-mode Bayesian methods, we use the last-layer Laplace approximation (LL-Laplace) \cite{kristiadi2020being}. For our method, we compare the unregularized last-layer ensemble (\acrshort{LLPOVI}), repulsion in parameter space (\acrshort{RLLPOVI}), and repulsion in function space (\acrshort{fLLPOVI}) with varying repulsion samples. The \acrshort{LLPOVI} consists of 10 particles with linear layers, introducing minimal computational overhead. During training, we use a batch size of 128 for both training data and repulsion samples.
To ensure a fair and meaningful comparison between deterministic distance-based methods (DDU, SNGP) and last-layer Bayesian methods (LL-Laplace, \acrshort{LLPOVI}, \acrshort{RLLPOVI}, \acrshort{fLLPOVI}), we always use the same pre-trained network as the base feature extractor. 
To quantify epistemic uncertainty we use the softmax entropy (MAP, SNGP), mutual information as given in \autoref{eq:UQ_decomp} (LL-Laplace, \acrshort{LLPOVI}, \gls{DE}-5), and GMM density (DDU). 
\tikzexternaldisable
\begin{figure}[t!]
    \centering
    \begin{subfigure}[b]{0.4\textwidth}
        \includegraphics[height=0.33\linewidth, trim={1.5cm 1.5cm .5cm .5cm},clip]{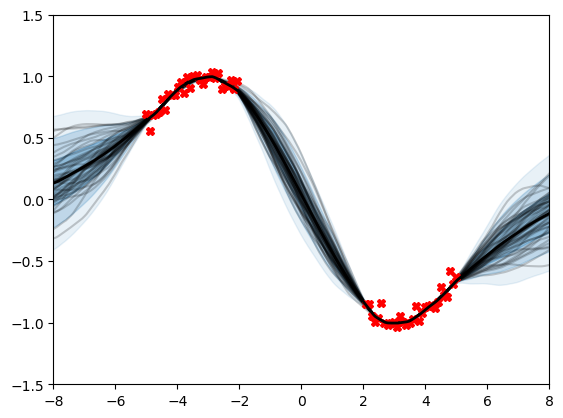}
        \includegraphics[height=0.33\linewidth, trim={0 0 0 0},clip]{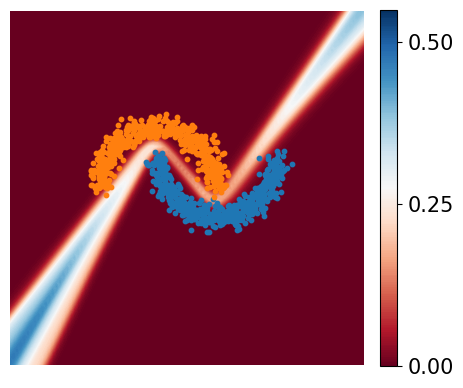}
        \caption{Unregularized \Glspl{DE}}
    \end{subfigure}
    \begin{subfigure}[b]{0.4\textwidth}
        \includegraphics[height=0.33\linewidth, trim={1.5cm 1.5cm .5cm .5cm},clip]{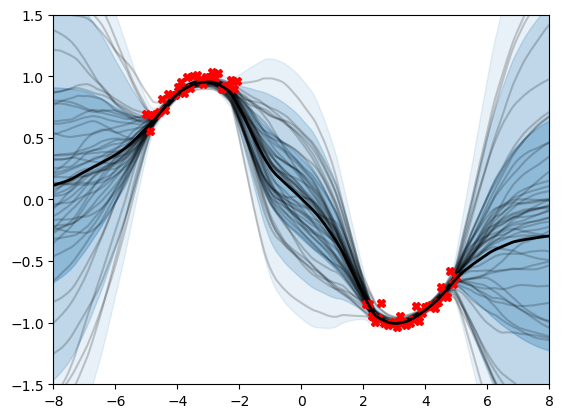}
        \includegraphics[height=0.33\linewidth, trim={0 0 0 0},clip]{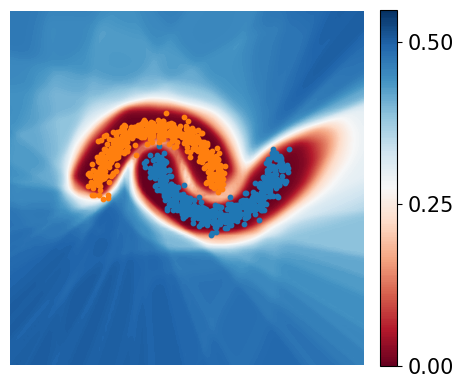}
        \caption{Function space \acrshort{RLLPOVI}}
    \end{subfigure}
    \label{fig:toy}
    \caption{
    For regression, we show the prediction of individual particles, the mean and the standard deviation. For classification, we show the standard deviation of $p(\rvy|\rvx,\theta)$. \glspl{DE} are highly confident in regions distant from training data, while \acrshort{fLLPOVI} predictions are enforced to be diverse outside of the training data.}
    \label{fig:toy}
\end{figure}
\tikzexternalenable

\subsubsection*{Evaluation metrics} 
We evaluate the reliability of uncertainty predictions using calibration on \gls{ID} data and distribution shifts, and the ability to discriminate \gls{OOD} data. 
For classification performance, we report accuracy (\textsc{Acc.}) and negative log-likelihood (\textsc{NLL}), 
$
\text{NLL} = -\frac{1}{N} \sum_{i=1}^{N} \log p(y_i | x_i)
$
with \(N\) being the number of samples, \(x_i\) the input data points, and \(y_i\) the true labels.
For calibration, we use the Expected Calibration Error (ECE) \cite{naeini2015obtaining}, which measures the agreement between predicted probabilities and actual outcomes. The ECE is computed by dividing predictions into \(M\) bins, 
$
\text{ECE} = \sum_{m=1}^{M} \frac{|B_m|}{N} \left| \text{acc}(B_m) - \text{conf}(B_m) \right|
$
where \( |B_m| \) is the number of samples in bin \( m \).
For OOD detection, we report the Area Under the ROC Curve (AUROC) of a classifier using the estimated epistemic uncertainty to distinguish ID and OOD samples.

\subsection{Synthetic Data}
On two toy examples, we illustrate the effectiveness of the multi-head architecture as a lightweight parameterization and the advantages of performing inference in function space.  
We estimate the epistemic uncertainty for a one-dimensional regression and a two-dimensional classification problem using full \acrshort{DE}s and \acrshort{fLLPOVI}. 
A feed-forward neural network with 3 hidden layers and 128 neurons is used as the base network. 
The repulsive head consists of 30 particles with linear layers. Results are shown in Figure \ref{fig:toy}. 
Deep ensemble predictions show low uncertainty far from the training data. By performing particle inference in function space, we can enforce diverse predictions outside the training distribution even with a simpler network structure.
\pgfplotstableread[col sep=comma]{sections/graphs/OOD_MNIST/OOD_net18_dirty_mnist_auroc.csv}\tableOOD
\pgfplotstableread[col sep=comma]{sections/graphs/OOD_MNIST/ID_net18_dirty_mnist_accuracy.csv}\tableIDaccuracy
\pgfplotstableread[col sep=comma]{sections/graphs/OOD_MNIST/ID_net18_dirty_mnist_nll.csv}\tableIDnll
\pgfplotstableread[col sep=comma]{sections/graphs/OOD_MNIST/ID_net18_dirty_mnist_ece.csv}\tableIDece

\pgfplotstableread[col sep=comma]{sections/graphs/OOD_MNIST/OOD_dirty_mnist_auroc_POVI_SGD.csv}\tableOODSGD
\pgfplotstableread[col sep=comma]{sections/graphs/OOD_MNIST/ID_dirty_mnist_accuracy_POVI_SGD.csv}\tableIDaccuracySGD
\pgfplotstableread[col sep=comma]{sections/graphs/OOD_MNIST/ID_dirty_mnist_nll_POVI_SGD.csv}\tableIDnllSGD
\pgfplotstableread[col sep=comma]{sections/graphs/OOD_MNIST/ID_dirty_mnist_ece_POVI_SGD.csv}\tableIDeceSGD

\pgfplotstableread[col sep=comma]{sections/graphs/OOD_MNIST/OOD_dirty_mnist_auroc_POVI_kde_WGD.csv}\tableOODWGD
\pgfplotstableread[col sep=comma]{sections/graphs/OOD_MNIST/ID_dirty_mnist_accuracy_POVI_kde_WGD.csv}\tableIDaccuracyWGD
\pgfplotstableread[col sep=comma]{sections/graphs/OOD_MNIST/ID_dirty_mnist_nll_POVI_kde_WGD.csv}\tableIDnllWGD
\pgfplotstableread[col sep=comma]{sections/graphs/OOD_MNIST/ID_dirty_mnist_ece_POVI_kde_WGD.csv}\tableIDeceWGD

\pgfplotstableread[col sep=comma]{sections/graphs/OOD_MNIST/OOD_dirty_mnist_auroc_POVI_kde_f_WGD_coll_emnist_128.csv}\tableOODemnist
\pgfplotstableread[col sep=comma]{sections/graphs/OOD_MNIST/ID_dirty_mnist_accuracy_POVI_kde_f_WGD_coll_emnist_128.csv}\tableIDaccuracyemnist
\pgfplotstableread[col sep=comma]{sections/graphs/OOD_MNIST/ID_dirty_mnist_nll_POVI_kde_f_WGD_coll_emnist_128.csv}\tableIDnllemnist
\pgfplotstableread[col sep=comma]{sections/graphs/OOD_MNIST/ID_dirty_mnist_ece_POVI_kde_f_WGD_coll_emnist_128.csv}\tableIDeceemnist

\pgfplotstableread[col sep=comma]{sections/graphs/OOD_MNIST/OOD_dirty_mnist_auroc_POVI_patches_std_16.csv}\tableOODPatchsixteen
\pgfplotstableread[col sep=comma]{sections/graphs/OOD_MNIST/ID_dirty_mnist_accuracy_POVI_patches_std_16.csv}\tableIDaccuracyPatchsixteen
\pgfplotstableread[col sep=comma]{sections/graphs/OOD_MNIST/ID_dirty_mnist_nll_POVI_patches_std_16.csv}\tableIDnllPatchsixteen
\pgfplotstableread[col sep=comma]{sections/graphs/OOD_MNIST/ID_dirty_mnist_ece_POVI_patches_std_16.csv}\tableIDecePatchsixteen

\pgfplotstableread[col sep=comma]{sections/graphs/OOD_MNIST/OOD_dirty_mnist_auroc_POVI_patches_std_8.csv}\tableOODPatcheight
\pgfplotstableread[col sep=comma]{sections/graphs/OOD_MNIST/ID_dirty_mnist_accuracy_POVI_patches_std_8.csv}\tableIDaccuracyPatcheight
\pgfplotstableread[col sep=comma]{sections/graphs/OOD_MNIST/ID_dirty_mnist_nll_POVI_patches_std_8.csv}\tableIDnllPatcheight
\pgfplotstableread[col sep=comma]{sections/graphs/OOD_MNIST/ID_dirty_mnist_ece_POVI_patches_std_8.csv}\tableIDecePatcheight

\pgfplotstableread[col sep=comma]{sections/graphs/OOD_MNIST/OOD_dirty_mnist_auroc_POVI_kde_f_WGD_patches_std_4.csv}\tableOODPatchfour
\pgfplotstableread[col sep=comma]{sections/graphs/OOD_MNIST/ID_dirty_mnist_accuracy_POVI_kde_f_WGD_patches_std_4.csv}\tableIDaccuracyPatchfour
\pgfplotstableread[col sep=comma]{sections/graphs/OOD_MNIST/ID_dirty_mnist_nll_POVI_kde_f_WGD_patches_std_4.csv}\tableIDnllPatchfour
\pgfplotstableread[col sep=comma]{sections/graphs/OOD_MNIST/ID_dirty_mnist_ece_POVI_kde_f_WGD_patches_std_4.csv}\tableIDecePatchfour

\pgfplotstableread[col sep=comma]{sections/graphs/OOD_MNIST/OOD_dirty_mnist_auroc_POVI_kde_f_WGD_coll_dirty_mnist_128.csv}\tableOODdirty
\pgfplotstableread[col sep=comma]{sections/graphs/OOD_MNIST/ID_dirty_mnist_accuracy_POVI_kde_f_WGD_coll_dirty_mnist_128.csv}\tableIDaccuracydirty
\pgfplotstableread[col sep=comma]{sections/graphs/OOD_MNIST/ID_dirty_mnist_nll_POVI_kde_f_WGD_coll_dirty_mnist_128.csv}\tableIDnlldirty
\pgfplotstableread[col sep=comma]{sections/graphs/OOD_MNIST/ID_dirty_mnist_ece_POVI_kde_f_WGD_coll_dirty_mnist_128.csv}\tableIDecedirty

\newcommand{\fetchAndStore}[4]{%
    \pgfplotstablegetelem{#1}{#2}\of{#3}
    \expandafter\xdef\csname #2Mean#1#4\endcsname{\pgfplotsretval}
    \pgfplotstablegetelem{#1}{#2_std}\of{#3}
    \expandafter\xdef\csname #2Std#1#4\endcsname{\pgfplotsretval}
}

\foreach \method in {softmax,logits,logits_GMM,softmax_laplace_MI,softmax_laplace_AU,softmax_SNGP_PE, logits_SNGP,softmax_en_AU,softmax_en_MI} {
    \foreach \intensity in {0,1,2} { 
        \fetchAndStore{\intensity}{\method}{\tableOOD}{OOD}
    }
}

\foreach \method in {softmax,logits,logits_GMM,softmax_laplace_MI,softmax_laplace_AU,softmax_SNGP_PE, logits_SNGP,softmax_en_AU,softmax_en_MI} {
    \fetchAndStore{0}{\method}{\tableIDaccuracy}{accuracy}
    \fetchAndStore{0}{\method}{\tableIDnll}{nll}
    \fetchAndStore{0}{\method}{\tableIDece}{ece}
}

\foreach \method in {softmax_POVI_MI, softmax_POVI_AU} {
    \foreach \intensity in {0,1,2} { 
        \fetchAndStore{\intensity}{\method}{\tableOODSGD}{OODSGD}
        \fetchAndStore{\intensity}{\method}{\tableOODWGD}{OODWGD}
        \fetchAndStore{\intensity}{\method}{\tableOODemnist}{OODemnist}
        \fetchAndStore{\intensity}{\method}{\tableOODPatchsixteen}{OODPatchsixteen}
        \fetchAndStore{\intensity}{\method}{\tableOODPatcheight}{OODPatcheight}
        \fetchAndStore{\intensity}{\method}{\tableOODPatchfour}{OODPatchfour}
        \fetchAndStore{\intensity}{\method}{\tableOODdirty}{OODdirty}
    
    }
}
\foreach \method in {softmax_POVI_MI, softmax_POVI_AU} {
    \fetchAndStore{0}{\method}{\tableIDaccuracySGD}{accuracySGD}
    \fetchAndStore{0}{\method}{\tableIDnllSGD}{nllSGD}
    \fetchAndStore{0}{\method}{\tableIDeceSGD}{eceSGD}

    \fetchAndStore{0}{\method}{\tableIDaccuracyWGD}{accuracyWGD}
    \fetchAndStore{0}{\method}{\tableIDnllWGD}{nllWGD}
    \fetchAndStore{0}{\method}{\tableIDeceWGD}{eceWGD}
    
    \fetchAndStore{0}{\method}{\tableIDaccuracyemnist}{accuracyemnist}
    \fetchAndStore{0}{\method}{\tableIDnllemnist}{nllemnist}
    \fetchAndStore{0}{\method}{\tableIDeceemnist}{eceemnist}

    \fetchAndStore{0}{\method}{\tableIDaccuracyPatchsixteen}{accuracyPatchsixteen}
    \fetchAndStore{0}{\method}{\tableIDnllPatchsixteen}{nllPatchsixteen}
    \fetchAndStore{0}{\method}{\tableIDecePatchsixteen}{ecePatchsixteen}
    
    \fetchAndStore{0}{\method}{\tableIDaccuracyPatcheight}{accuracyPatcheight}
    \fetchAndStore{0}{\method}{\tableIDnllPatcheight}{nllPatcheight}
    \fetchAndStore{0}{\method}{\tableIDecePatcheight}{ecePatcheight}
    
    \fetchAndStore{0}{\method}{\tableIDaccuracyPatchfour}{accuracyPatchfour}
    \fetchAndStore{0}{\method}{\tableIDnllPatchfour}{nllPatchfour}
    \fetchAndStore{0}{\method}{\tableIDecePatchfour}{ecePatchfour}
    
    \fetchAndStore{0}{\method}{\tableIDaccuracydirty}{accuracydirty}
    \fetchAndStore{0}{\method}{\tableIDnlldirty}{nlldirty}
    \fetchAndStore{0}{\method}{\tableIDecedirty}{ecedirty}
}

\begin{table*}[t!]
\centering
\caption{Comparison of uncertainty decomposition on DirtyMNIST. Aleatoric uncertainty (AU), and epistemic uncertainty (EU) are used to detect ambiguous, and \acrshort{OOD} samples. Mean and standard deviation are computed over 10 runs. Best results are in bold, second best are underlined.}
\label{tab:my-table}
\resizebox{\textwidth}{!}{%
\begin{tabular}{@{}lcccccc@{}} 
\toprule
\multirow{2}{*}{\textbf{Method}} & \multirow{2}{*}{\textsc{Acc. $\uparrow$ [\%]}} & \multirow{2}{*}{\textsc{NLL $\downarrow$ [\%]}} & \multirow{2}{*}{\textsc{ECE $\downarrow$ [\%]}} & \multicolumn{3}{c}{\textsc{OOD Auroc $\uparrow$ [\%] }} \\ \cmidrule(l){5-7} 
                                 &                                                           &                                                               &                                                           & MNIST vs ambig. (AU) & MNIST vs. OOD (EU) & ambig. vs OOD (EU) \\ \midrule
MAP & ${\csname softmaxMean0accuracy\endcsname _{\pm \csname softmaxStd0accuracy\endcsname}}$ & ${\csname softmaxMean0nll\endcsname _{\pm \csname softmaxStd0nll\endcsname}}$ & ${\csname softmaxMean0ece\endcsname _{\pm \csname softmaxStd0ece\endcsname}}$ & ${\csname softmaxMean0OOD\endcsname _{\pm \csname softmaxStd0OOD\endcsname}}$ & ${\csname softmaxMean1OOD\endcsname _{\pm \csname softmaxStd1OOD\endcsname}}$ & ${\csname softmaxMean2OOD\endcsname _{\pm \csname softmaxStd2OOD\endcsname}}$ \\ \midrule
DDU & ${\csname softmaxMean0accuracy\endcsname _{\pm \csname softmaxStd0accuracy\endcsname}}$ & ${\csname softmaxMean0nll\endcsname _{\pm \csname softmaxStd0nll\endcsname}}$ & ${\csname softmaxMean0ece\endcsname _{\pm \csname softmaxStd0ece\endcsname}}$ & ${\csname softmaxMean0OOD\endcsname _{\pm \csname softmaxStd0OOD\endcsname}}$ & $\mathbf{\csname logits_GMMMean1OOD\endcsname _{\pm \csname logits_GMMStd1OOD\endcsname}}$ & $\mathbf{\csname logits_GMMMean2OOD\endcsname _{\pm \csname logits_GMMStd2OOD\endcsname}}$ \\
SNGP & $\underlineitalic{\csname softmax_SNGP_PEMean0accuracy\endcsname _{\pm \csname softmax_SNGP_PEStd0accuracy\endcsname}}$ & ${\csname softmax_SNGP_PEMean0nll\endcsname _{\pm \csname softmax_SNGP_PEStd0nll\endcsname}}$ & ${\csname softmax_SNGP_PEMean0ece\endcsname _{\pm \csname softmax_SNGP_PEStd0ece\endcsname}}$ & ${\csname softmax_SNGP_PEMean0OOD\endcsname _{\pm \csname softmax_SNGP_PEStd0OOD\endcsname}}$ & ${\csname softmax_SNGP_PEMean1OOD\endcsname _{\pm \csname softmax_SNGP_PEStd1OOD\endcsname}}$ & ${\csname softmax_SNGP_PEMean2OOD\endcsname _{\pm \csname logits_SNGPStd2OOD\endcsname}}$ \\
LL-Laplace & ${\csname softmax_laplace_MIMean0accuracy\endcsname _{\pm \csname softmax_laplace_MIStd0accuracy\endcsname}}$ & ${\csname softmax_laplace_MIMean0nll\endcsname _{\pm \csname softmax_laplace_MIStd0nll\endcsname}}$ & ${\csname softmax_laplace_MIMean0ece\endcsname _{\pm \csname softmax_laplace_MIStd0ece\endcsname}}$ & ${\csname softmax_laplace_AUMean0OOD\endcsname _{\pm \csname softmax_laplace_AUStd0OOD\endcsname}}$ & ${\csname softmax_laplace_MIMean1OOD\endcsname _{\pm \csname softmax_laplace_MIStd1OOD\endcsname}}$ & ${\csname softmax_laplace_MIMean2OOD\endcsname _{\pm \csname softmax_laplace_MIStd2OOD\endcsname}}$ \\ \midrule
\acrshort{LLPOVI} \emph{(ours)} & $\mathbf{\csname softmax_POVI_MIMean0accuracySGD\endcsname _{\pm \csname softmax_POVI_MIStd0accuracySGD\endcsname}}$ & $\mathbf{\csname softmax_POVI_MIMean0nllSGD\endcsname _{\pm \csname softmax_POVI_MIStd0nllSGD\endcsname}}$ & $\mathbf{\csname softmax_POVI_MIMean0eceSGD\endcsname _{\pm \csname softmax_POVI_MIStd0eceSGD\endcsname}}$ & $\mathbf{\csname softmax_POVI_AUMean0OODSGD\endcsname _{\pm \csname softmax_POVI_MIStd0OODSGD\endcsname}}$ & ${\csname softmax_POVI_MIMean1OODSGD\endcsname _{\pm \csname softmax_POVI_MIStd1OODSGD\endcsname}}$ & ${\csname softmax_POVI_MIMean2OODSGD\endcsname _{\pm \csname softmax_POVI_MIStd2OODSGD\endcsname}}$ \\ 
\acrshort{RLLPOVI} \emph{(ours)}  & $\mathbf{\csname softmax_POVI_MIMean0accuracyWGD\endcsname _{\pm \csname softmax_POVI_MIStd0accuracyWGD\endcsname}}$ & $\mathbf{\csname softmax_POVI_MIMean0nllWGD\endcsname _{\pm \csname softmax_POVI_MIStd0nllWGD\endcsname}}$ & $\mathbf{\csname softmax_POVI_MIMean0eceWGD\endcsname _{\pm \csname softmax_POVI_MIStd0eceWGD\endcsname}}$ & $\mathbf{\csname softmax_POVI_AUMean0OODWGD\endcsname _{\pm \csname softmax_POVI_MIStd0OODWGD\endcsname}}$ & ${\csname softmax_POVI_MIMean1OODWGD\endcsname _{\pm \csname softmax_POVI_MIStd1OODWGD\endcsname}}$ & ${\csname softmax_POVI_MIMean2OODWGD\endcsname _{\pm \csname softmax_POVI_MIStd2OODWGD\endcsname}}$ \\ 
\acrshort{fLLPOVI} \emph{(ours)}      &                                                           &                                                               &                                                           &                       &              &                     \\
\textit{~~~+ dirtyMNIST}  & ${\csname softmax_POVI_MIMean0accuracydirty\endcsname _{\pm \csname softmax_POVI_MIStd0accuracydirty\endcsname}}$ & ${\csname softmax_POVI_MIMean0nlldirty\endcsname _{\pm \csname softmax_POVI_MIStd0nlldirty\endcsname}}$ & ${\csname softmax_POVI_MIMean0ecedirty\endcsname _{\pm \csname softmax_POVI_MIStd0ecedirty\endcsname}}$ & ${\csname softmax_POVI_AUMean0OODdirty\endcsname _{\pm \csname softmax_POVI_MIStd0OODdirty\endcsname}}$ & ${\csname softmax_POVI_MIMean1OODdirty\endcsname _{\pm \csname softmax_POVI_MIStd1OODdirty\endcsname}}$ & ${\csname softmax_POVI_MIMean2OODdirty\endcsname _{\pm \csname softmax_POVI_MIStd2OODdirty\endcsname}}$ \\
\textit{~~~+ eMNIST}  & $\underlineitalic{\csname softmax_POVI_MIMean0accuracyemnist\endcsname _{\pm \csname softmax_POVI_MIStd0accuracyemnist\endcsname}}$ & ${\csname softmax_POVI_MIMean0nllemnist\endcsname _{\pm \csname softmax_POVI_MIStd0nllemnist\endcsname}}$ & ${\csname softmax_POVI_MIMean0eceemnist\endcsname _{\pm \csname softmax_POVI_MIStd0eceemnist\endcsname}}$ & ${\csname softmax_POVI_AUMean0OODemnist\endcsname _{\pm \csname softmax_POVI_MIStd0OODemnist\endcsname}}$ & ${\csname softmax_POVI_MIMean1OODemnist\endcsname _{\pm \csname softmax_POVI_MIStd1OODemnist\endcsname}}$ & $\underlineitalic{\csname softmax_POVI_MIMean2OODemnist\endcsname _{\pm \csname softmax_POVI_MIStd2OODemnist\endcsname}}$ \\
\textit{~~~+ Patches-16} & $\underlineitalic{\csname softmax_POVI_MIMean0accuracyPatchsixteen\endcsname _{\pm \csname softmax_POVI_MIStd0accuracyPatchsixteen\endcsname}}$ & $\underlineitalic{\csname softmax_POVI_MIMean0nllPatchsixteen\endcsname _{\pm \csname softmax_POVI_MIStd0nllPatchsixteen\endcsname}}$ & $\underlineitalic{\csname softmax_POVI_MIMean0ecePatchsixteen\endcsname _{\pm \csname softmax_POVI_MIStd0ecePatchsixteen\endcsname}}$ & $\underlineitalic{\csname softmax_POVI_AUMean0OODPatchsixteen\endcsname _{\pm \csname softmax_POVI_MIStd0OODPatchsixteen\endcsname}}$ & $\underlineitalic{\csname softmax_POVI_MIMean1OODPatchsixteen\endcsname _{\pm \csname softmax_POVI_MIStd1OODPatchsixteen\endcsname}}$ & ${\csname softmax_POVI_MIMean2OODPatchsixteen\endcsname _{\pm \csname softmax_POVI_MIStd2OODPatchsixteen\endcsname}}$ \\
\textit{~~~+ Patches-8} & $\underlineitalic{\csname softmax_POVI_MIMean0accuracyPatcheight\endcsname _{\pm \csname softmax_POVI_MIStd0accuracyPatcheight\endcsname}}$ & ${\csname softmax_POVI_MIMean0nllPatcheight\endcsname _{\pm \csname softmax_POVI_MIStd0nllPatcheight\endcsname}}$ & $\underlineitalic{\csname softmax_POVI_MIMean0ecePatcheight\endcsname _{\pm \csname softmax_POVI_MIStd0ecePatcheight\endcsname}}$ & ${\csname softmax_POVI_AUMean0OODPatcheight\endcsname _{\pm \csname softmax_POVI_MIStd0OODPatcheight\endcsname}}$ & ${\csname softmax_POVI_MIMean1OODPatcheight\endcsname _{\pm \csname softmax_POVI_MIStd1OODPatcheight\endcsname}}$ & ${\csname softmax_POVI_MIMean2OODPatcheight\endcsname _{\pm \csname softmax_POVI_MIStd2OODPatcheight\endcsname}}$ \\ 
\textit{~~~+ Patches-4} & $\underlineitalic{\csname softmax_POVI_MIMean0accuracyPatchfour\endcsname _{\pm \csname softmax_POVI_MIStd0accuracyPatchfour\endcsname}}$ & ${\csname softmax_POVI_MIMean0nllPatchfour\endcsname _{\pm \csname softmax_POVI_MIStd0nllPatchfour\endcsname}}$ & $\underlineitalic{\csname softmax_POVI_MIMean0ecePatchfour\endcsname _{\pm \csname softmax_POVI_MIStd0ecePatchfour\endcsname}}$ & ${\csname softmax_POVI_AUMean0OODPatchfour\endcsname _{\pm \csname softmax_POVI_MIStd0OODPatchfour\endcsname}}$ & ${\csname softmax_POVI_MIMean1OODPatchfour\endcsname _{\pm \csname softmax_POVI_MIStd1OODPatchfour\endcsname}}$ & ${\csname softmax_POVI_MIMean2OODPatchfour\endcsname _{\pm \csname softmax_POVI_MIStd2OODPatchfour\endcsname}}$ \\ \midrule

DE-5 & ${\csname softmax_en_MIMean0accuracy\endcsname _{\pm \csname softmax_en_MIStd0accuracy\endcsname}}$ & ${\csname softmax_en_MIMean0nll\endcsname _{\pm \csname softmax_en_MIStd0nll\endcsname}}$ & ${\csname softmax_en_MIMean0ece\endcsname _{\pm \csname softmax_en_MIStd0ece\endcsname}}$ & ${\csname softmax_en_AUMean0OOD\endcsname _{\pm \csname softmax_en_AUStd0OOD\endcsname}}$ & ${\csname softmax_en_MIMean1OOD\endcsname _{\pm \csname softmax_en_MIStd1OOD\endcsname}}$ & ${\csname softmax_en_MIMean2OOD\endcsname _{\pm \csname softmax_en_MIStd2OOD\endcsname}}$ \\ \bottomrule
\end{tabular}}
\label{tab:OOD:mnist}
\end{table*}

\subsection{Uncertainty decomposition}\label{sec:uncertainty_decomp}
Common image classification datasets do not contain data points that are inherently ambiguous, i.e., data points that correspond to multiple classes and result in irreducible aleatoric uncertainty. 
Therefore, we use the \mbox{DirtyMNIST} dataset proposed by \cite{mukhoti2023deep} to evaluate the ability of our method to distinguish between ambiguous data, reflected in aleatoric uncertainty, and \gls{OOD} data, reflected in epistemic uncertainty. The \mbox{DirtyMNIST} dataset consists of MNIST digits with a unique class and artificial ambiguous digits that belong to multiple classes. For \gls{OOD} data we use kMNIST \cite{kmnist}, \mbox{fashionMNIST} \cite{xiao2017fashion}, and Omniglot \cite{omniglot}.
A Resnet-18 architecture serves as our base network. For our function space method, we evaluate different sets of repulsion samples $\repset$. 
These include training data samples (\textit{\mbox{DirtyMNIST}}), images from a related dataset (\textit{eMNIST}), and randomly shuffled image patches from the training data samples (\textit{Patches-$x$}), as illustrated in Fig. \ref{fig:repulsion}.

Histograms of the aleatoric versus epistemic uncertainty estimates for \gls{fLLPOVI}, \acrshort{RLLPOVI}, and \gls{DE} on different test sets are shown in Fig. \ref{fig:2dhist}. With both MNIST and ambiguous MNIST in the training data, epistemic uncertainty should be low on samples from the respective test sets. We expect high aleatoric uncertainty for ambiguous MNIST and high epistemic uncertainty for the unknown \gls{OOD} \mbox{fashionMNIST} test samples. 
Recent research \cite{xia2022usefulness, schweighofer2024quantification} has raised concerns about the quality of uncertainty estimates for \glspl{DE}, showing that aleatoric uncertainty may detect \gls{OOD} more reliably than epistemic uncertainty. We find that \glspl{DE} (Fig. \ref{fig:2dhist} (a)) indeed exhibit increased aleatoric uncertainty for both ambiguous and \gls{OOD} data, making it difficult to accurately discriminate between the two. 
The same applies to our unregularized \acrshort{LLPOVI} (Fig. \ref{fig:2dhist} (b)), where \gls{OOD} data yields high aleatoric uncertainty. In the case of the function space method, \gls{fLLPOVI} (Fig. \ref{fig:2dhist} (c)), the training procedure is modified by directly forcing diverse predictions on augmented training data. This leads to higher epistemic uncertainty for \mbox{fashionMNIST} and better separation of ambiguous and \gls{OOD}. Retraining the last layer using function space repulsion can be sufficiently expressive for learning diverse predictions on \gls{OOD} data without the need for a full \gls{DE}.
\pgfplotsset{compat=1.17}

\pgfplotsset{
    colormap/bluecolormap/.style={colormap={bluecolormap}{
            rgb255=(255,255,255);
            rgb255=(192,192,255));
            rgb255=(0,0,255);
            rgb255=(0,0,0);
        }
    },
    colormap/redcolormap/.style={colormap={redcolormap}{
            rgb255=(255,255,255);
            rgb255=(255,192,192);
            rgb255=(255,0,0);
            rgb255=(0,0,102);
        }
    },
}

\tikzexternaldisable


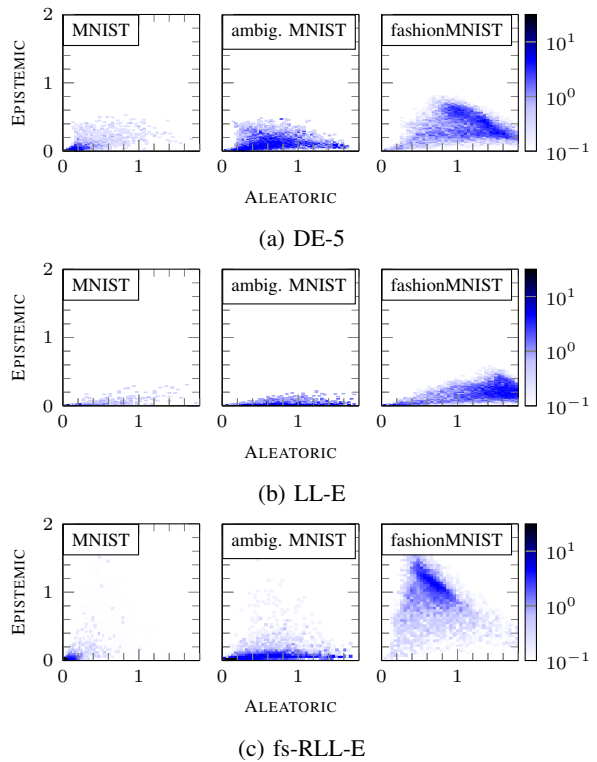
\begin{figure}[t!]
\centering

\begin{subfigure}{.5\textwidth}
\centering

\begin{tikzpicture}

\begin{groupplot}[
    group style={
        group size=3 by 1,
        horizontal sep=0.3cm,
    },
    height=3.4cm,
    width=3.0cm,
    legend style={
        at={(0.2,0)}, 
        anchor=west,
        draw=none,      
        fill=none,       
        legend columns=8
    },
]

\nextgroupplot[
            view={0}{90},
            ylabel={\textsc{Epistemic}},
            minor tick num=4,
            colormap/bluecolormap,
            point meta min=-1,
            point meta max=1.5,
            colorbar style={
                ylabel={},
                ytick={-1,0,...,1},
                width=0.03\linewidth, 
                at={(1.05,1)}, 
                font=\scriptsize,
                yticklabel=$10^{\pgfmathparse{\tick}\pgfmathprintnumber\pgfmathresult}$,
                ylabel style={yshift=0pt,},
            },
            width=3.4cm, 
            height=3.4cm, 
            xmin=0, xmax=1.8, ymin=0, ymax=2,  
            title={MNIST},every axis title/.style={below right,at={(0,1)},draw=black,fill=white},
            font=\scriptsize,
                ]

            \addplot3[
                surf,
                shader=flat corner,
                mesh/cols=50,
                mesh/ordering=rowwise,
                point meta={log10(abs(\thisrow{weight}))},
            ] table[x=xcenter, y=ycenter] {sections/graphs/data/2d_histograms/2d_histogram_mnist_softmax_en.csv};

\nextgroupplot[
            view={0}{90},
            xlabel={\textsc{Aleatoric}},
            ylabel={$~$},
            minor tick num=4,
            yticklabels = {},
            colormap/bluecolormap,
            point meta min=-1,
            point meta max=1.5,
            colorbar style={
                ylabel={},
                ytick={-1,0,...,1},
                width=0.03\linewidth, 
                at={(1.05,1)}, 
                font=\scriptsize,
                yticklabel=$10^{\pgfmathparse{\tick}\pgfmathprintnumber\pgfmathresult}$,
                ylabel style={yshift=0pt,},
            },
            width=3.4cm, 
            height=3.4cm, 
            xmin=0, xmax=1.8, ymin=0, ymax=2,  
            title={ambig. MNIST},every axis title/.style={below right,at={(0,1)},draw=black,fill=white}, 
            font=\scriptsize,
                ]

            \addplot3[
                surf,
                shader=flat corner,
                mesh/cols=50,
                mesh/ordering=rowwise,
                point meta={log10(abs(\thisrow{weight}))},
            ] table[x=xcenter, y=ycenter] {sections/graphs/data/2d_histograms/2d_histogram_ambiguous_mnist_softmax_en.csv};

\nextgroupplot[
            view={0}{90},
            ylabel={$~$},
            minor tick num=4,
            yticklabels = {},
            colorbar,
            colormap/bluecolormap,
            point meta min=-1,
            point meta max=1.5,
            colorbar style={
                ylabel={},
                ytick={-1,0,...,1},
                width=0.15cm, 
                at={(1.05,1)}, 
                font=\scriptsize,
                yticklabel=$10^{\pgfmathparse{\tick}\pgfmathprintnumber\pgfmathresult}$,
                ylabel style={yshift=0pt,},
            },
            width=3.4cm, 
            height=3.4cm, 
            xmin=0, xmax=1.8, ymin=0, ymax=2,  
            title={fashionMNIST},every axis title/.style={below right,at={(0,1)},draw=black,fill=white},
            font=\scriptsize,
                ]

            \addplot3[
                surf,
                shader=flat corner,
                mesh/cols=50,
                mesh/ordering=rowwise,
                point meta={log10(abs(\thisrow{weight}))},
            ] table[x=xcenter, y=ycenter] {sections/graphs/data/2d_histograms/2d_histogram_fashion_mnist_softmax_en.csv};

    \end{groupplot}
\end{tikzpicture}

\caption{\acrshort{DE}-5}
\end{subfigure}
\hspace{1cm}
\begin{subfigure}{.5\textwidth}
\centering

\begin{tikzpicture}

\begin{groupplot}[
    group style={
        group size=3 by 1,
        horizontal sep=0.3cm,
    },
    height=3.4cm,
    width=3.0cm,
    legend style={
        at={(0.2,1.25)}, 
        anchor=west,
        draw=none,      
        fill=none,       
        legend columns=8
    },
]

\nextgroupplot[
            view={0}{90},
            ylabel={\textsc{Epistemic}},
            minor tick num=4,
            colormap/bluecolormap,
            point meta min=-1,
            point meta max=1.5,
            colorbar style={
                ylabel={},
                ytick={-1,0,...,1},
                width=0.03\linewidth, 
                at={(1.05,1)}, 
                font=\scriptsize,
                yticklabel=$10^{\pgfmathparse{\tick}\pgfmathprintnumber\pgfmathresult}$,
                ylabel style={yshift=0pt,},
            },
            width=3.4cm, 
            height=3.4cm, 
            xmin=0, xmax=1.8, ymin=0, ymax=2,  
            title={MNIST},every axis title/.style={below right,at={(0,1)},draw=black,fill=white}, 
            font=\scriptsize,
                ]

            \addplot3[
                surf,
                shader=flat corner,
                mesh/cols=50,
                mesh/ordering=rowwise,
                point meta={log10(abs(\thisrow{weight}))},
            ] table[x=xcenter, y=ycenter] {sections/graphs/data/2d_histograms/2d_histogram_mnist_softmax_POVI_SGD.csv};

\nextgroupplot[
            view={0}{90},
            xlabel={\textsc{Aleatoric}},
            ylabel={$~$},
            minor tick num=4,
            colormap/bluecolormap,
            point meta min=-1,
            point meta max=1.5,
            colorbar style={
                ylabel={},
                ytick={-1,0,...,1},
                width=0.03\linewidth, 
                at={(1.05,1)}, 
                font=\scriptsize,
                yticklabel=$10^{\pgfmathparse{\tick}\pgfmathprintnumber\pgfmathresult}$,
                ylabel style={yshift=0pt,},
            },
            width=3.4cm, 
            height=3.4cm, 
            xmin=0, xmax=1.8, ymin=0, ymax=2,  
            title={ambig. MNIST},every axis title/.style={below right,at={(0,1)},draw=black,fill=white},  
            font=\scriptsize,
            yticklabels = {}
                ]

            \addplot3[
                surf,
                shader=flat corner,
                mesh/cols=50,
                mesh/ordering=rowwise,
                point meta={log10(abs(\thisrow{weight}))},
            ] table[x=xcenter, y=ycenter] {sections/graphs/data/2d_histograms/2d_histogram_ambiguous_mnist_softmax_POVI_SGD.csv};

\nextgroupplot[
            view={0}{90},
            ylabel={$~$},
            minor tick num=4,
            yticklabels = {},
            colorbar,
            colormap/bluecolormap,
            point meta min=-1,
            point meta max=1.5,
            colorbar style={
                ylabel={},
                ytick={-1,0,...,1},
                width=0.15cm, 
                at={(1.05,1)}, 
                font=\scriptsize,
                yticklabel=$10^{\pgfmathparse{\tick}\pgfmathprintnumber\pgfmathresult}$,
                ylabel style={yshift=0pt,},
            },
            width=3.4cm, 
            height=3.4cm, 
            xmin=0, xmax=1.8, ymin=0, ymax=2,  
            title={fashionMNIST},every axis title/.style={below right,at={(0,1)},draw=black,fill=white},  
            font=\scriptsize,
                ]

            \addplot3[
                surf,
                shader=flat corner,
                mesh/cols=50,
                mesh/ordering=rowwise,
                point meta={log10(abs(\thisrow{weight}))},
            ] table[x=xcenter, y=ycenter] {sections/graphs/data/2d_histograms/2d_histogram_fashion_mnist_softmax_POVI_SGD.csv};

    \end{groupplot}
\end{tikzpicture}
\caption{\acrshort{LLPOVI}}

\end{subfigure}

\begin{subfigure}{.5\textwidth}
\centering

\begin{tikzpicture}

\begin{groupplot}[
    group style={
        group size=3 by 1,
        horizontal sep=0.3cm,
    },
    height=3.4cm,
    width=3.0cm,
    legend style={
        at={(0.2,1.25)}, 
        anchor=west,
        draw=none,      
        fill=none,       
        legend columns=8
    },
]

\nextgroupplot[
            view={0}{90},
            ylabel={\textsc{Epistemic}},
            minor tick num=4,
            colormap/bluecolormap,
            point meta min=-1,
            point meta max=1.5,
            colorbar style={
                ylabel={},
                ytick={-1,0,...,1},
                width=0.03\linewidth, 
                at={(1.05,1)}, 
                font=\scriptsize,
                yticklabel=$10^{\pgfmathparse{\tick}\pgfmathprintnumber\pgfmathresult}$,
                ylabel style={yshift=0pt,},
            },
            width=3.4cm, 
            height=3.4cm, 
            xmin=0, xmax=1.8, ymin=0, ymax=2,  
            title={MNIST},every axis title/.style={below right,at={(0,1)},draw=black,fill=white},
            font=\scriptsize,
                ]

            \addplot3[
                surf,
                shader=flat corner,
                mesh/cols=50,
                mesh/ordering=rowwise,
                point meta={log10(abs(\thisrow{weight}))},
            ] table[x=xcenter, y=ycenter] {sections/graphs/data/2d_histograms/2d_histogram_mnist_softmax_POVI_kde_f_WGD_patches_random.csv};

\nextgroupplot[
            view={0}{90},
            xlabel={\textsc{Aleatoric}},
            ylabel={$~$},
            minor tick num=4,
            yticklabels = {},
            colormap/bluecolormap,
            point meta min=-1,
            point meta max=1.5,
            colorbar style={
                ylabel={},
                ytick={-1,0,...,1},
                width=0.03\linewidth, 
                at={(1.05,1)}, 
                font=\scriptsize,
                yticklabel=$10^{\pgfmathparse{\tick}\pgfmathprintnumber\pgfmathresult}$,
                ylabel style={yshift=0pt,},
            },
            width=3.4cm, 
            height=3.4cm, 
            xmin=0, xmax=1.8, ymin=0, ymax=2,  
            title={ambig. MNIST},every axis title/.style={below right,at={(0,1)},draw=black,fill=white},
            font=\scriptsize,
                ]

            \addplot3[
                surf,
                shader=flat corner,
                mesh/cols=50,
                mesh/ordering=rowwise,
                point meta={log10(abs(\thisrow{weight}))},
            ] table[x=xcenter, y=ycenter] {sections/graphs/data/2d_histograms/2d_histogram_ambiguous_mnist_softmax_POVI_kde_f_WGD_patches_random.csv};

\nextgroupplot[
            view={0}{90},
            ylabel={$~$},
            minor tick num=4,
            colorbar,
            colormap/bluecolormap,
            point meta min=-1,
            point meta max=1.5,
            yticklabels = {}, 
            colorbar style={
                ylabel={},
                ytick={-1,0,...,1},
                width=0.15cm, 
                at={(1.05,1)}, 
                font=\scriptsize,
                yticklabel=$10^{\pgfmathparse{\tick}\pgfmathprintnumber\pgfmathresult}$,
                ylabel style={yshift=0pt,},
            },
            width=3.4cm, 
            height=3.4cm, 
            xmin=0, xmax=1.8, ymin=0, ymax=2,  
            title={fashionMNIST},every axis title/.style={below right,at={(0,1)},draw=black,fill=white},
            font=\scriptsize,
                ]

            \addplot3[
                surf,
                shader=flat corner,
                mesh/cols=50,
                mesh/ordering=rowwise,
                point meta={log10(abs(\thisrow{weight}))},
            ] table[x=xcenter, y=ycenter] 
            {sections/graphs/data/2d_histograms/2d_histogram_fashion_mnist_softmax_POVI_kde_f_WGD_patches_random.csv};

    \end{groupplot}
\end{tikzpicture}
\caption{\acrshort{fLLPOVI}}

\end{subfigure}

\caption{Histograms of aleatoric versus epistemic uncertainty on ID data (MNIST, Ambiguous MNIST) and OOD data (fashion-MNIST). We compare (a) an unregularized \acrshort{DE}-5, (b) an unregularized \acrshort{LLPOVI}, (c) and \acrshort{fLLPOVI} with augmented training data as repulsion samples ($\rep=\textit{Patches-8}$). }
\label{fig:2dhist}
\end{figure}

Table \ref{tab:OOD:mnist} summarizes the ID performance and OOD detection gain we obtain by retraining the last layer of the base network for different choices of repulsion samples.
Applying our \acrshort{LLPOVI} without any repulsion term improves on the base network (MAP) in all evaluation metrics. Imposing diversity on the parameters of the last layer particles (\acrshort{RLLPOVI}), however, does not provide any additional benefit. 
Further improvements in uncertainty decomposition are achieved with function space repulsion (\gls{fLLPOVI}) and an appropriate choice of repulsion samples (patches, eMNIST). We note that the evaluated \gls{OOD} datasets did not include samples from $\repset$, i.e. diversity for repulsion samples is able to generalize to unseen \gls{OOD} data. As discussed in Section \ref{sec:my_method}, we show that forcing diversity on the training data yields worse results than simply training an unregularized \acrshort{LLPOVI}. 
Interestingly, while uncertainty methods based on the disagreement between predictions have difficulty in discriminating between ambiguous and OOD data, it is the easier case for DDU. Since DDU relies solely on feature space density, ambiguous samples belonging to multiple classes tend to have higher density than samples with unique classes.
In summary, our \gls{fLLPOVI} achieves competitive results with DDU in terms of uncertainty decomposition and, in addition, improves uncertainty calibration on \gls{ID} data, i.e. low ECE scores. 

In Appendix B.2, we evaluate our method for base network architectures without regularization to avoid feature collapse (LeNet, VGG-16). In Appendix B.3, we analyze if repulsion samples can improve uncertainty decomposition in poorly regularized feature spaces.

\begin{figure}[]
    \centering
    \pgfplotstableread[col sep=comma]{sections/graphs/active_learning/ensemble.csv}\tableEnsemble
\pgfplotstableread[col sep=comma]{sections/graphs/active_learning/softmax.csv}\tableSoftmax
\pgfplotstableread[col sep=comma]{sections/graphs/active_learning/gmm.csv}\tableGMM
\pgfplotstableread[col sep=comma]{sections/graphs/active_learning/povi_SGD.csv}\tablePoviSGD
\pgfplotstableread[col sep=comma]{sections/graphs/active_learning/povi_kde_WGD.csv}\tablePoviWGD
\pgfplotstableread[col sep=comma]{sections/graphs/active_learning/povi_kde_f_WGD_train_gamma_0_001_pool_128.csv}\tablePoviPool
\pgfplotstableread[col sep=comma]{sections/graphs/active_learning/povi_kde_f_WGD_train_gamma_0_1_pool_128.csv}\tablePoviPool

\newcommand{\uncertaintyplot}[5]{
    \addplot [name path=upper,draw=none,forget plot] table[x expr=\thisrow{iteration}*5,y expr=\thisrow{#1_mean}+\thisrow{#1_std}] {#4};
    \addplot [name path=lower,draw=none,forget plot] table[x expr=\thisrow{iteration}*5,y expr=\thisrow{#1_mean}-\thisrow{#1_std}] {#4};
    \addplot [fill=#2,forget plot,fill opacity=0.1] fill between[of=upper and lower];
    \addplot [color=#2, line width=1.5pt, dash pattern=#5] table[x expr=\thisrow{iteration}*5,y=#1_mean] {#4};
    \addlegendentry{#3}
}

\tikzexternaldisable
\begin{tikzpicture}
\begin{axis}[
    grid=major,
    xlabel=Acquired dataset size,
    ylabel=Test Accuracy,
    legend style={
        at={(0.5,1.2)}, 
        anchor=center,
        draw=none,      
        fill=none,       
        legend columns=2,
        row sep=.15cm,     
        font=\scriptsize,
    },
    transpose legend,
    font=\scriptsize,
    xmin=0,           
    xmax=295,         
    ymin=20,           
    ymax=70,           
    width=0.45\textwidth,  
    height=0.35\textwidth,          
    ticklabel style={font=\scriptsize} 
]

\uncertaintyplot{ambiguous}{MidnightBlue}{DE-3 (PE)}{\tableEnsemble}{}
\uncertaintyplot{ambiguous_MI}{MidnightBlue}{DE-3 (MI)}{\tableEnsemble}{on 1pt off 1pt}
\uncertaintyplot{ambiguous_MI}{darkgray}{MAP}{\tableSoftmax}{}
\uncertaintyplot{ambiguous_MI}{Mulberry}{DDU}{\tableGMM}{}
\uncertaintyplot{ambiguous}{orange}{\acrshort{LLPOVI}  (PE)}{\tablePoviSGD}{}
\uncertaintyplot{ambiguous_MI}{orange}{\acrshort{LLPOVI} (MI)}{\tablePoviSGD}{on 1pt off 1pt}
\uncertaintyplot{ambiguous}{Green}{$\underset{\textit{pool}}{\text{\acrshort{fLLPOVI}}}$ (PE)}{\tablePoviPool}{}
\uncertaintyplot{ambiguous_MI}{Green}{$\underset{\textit{pool}}{\text{\acrshort{fLLPOVI}}}$ (MI)}{\tablePoviPool}{on 1pt off 1pt}
\end{axis}
\end{tikzpicture}
\tikzexternalenable
    \caption{Test accuracy of the model as a function of the data samples that are acquired using the different uncertainty estimates. Predictive entropy (PE) combines aleatoric and epistemic uncertainty. Using the mutual information (MI) of the \acrshort{LLPOVI} and \acrshort{fLLPOVI} prediction outperforms softmax entropy of the single network and performs on par with the other uncertainty baselines. The results are averaged over 5 runs.
 }    \label{fig:active_learning}
\end{figure}
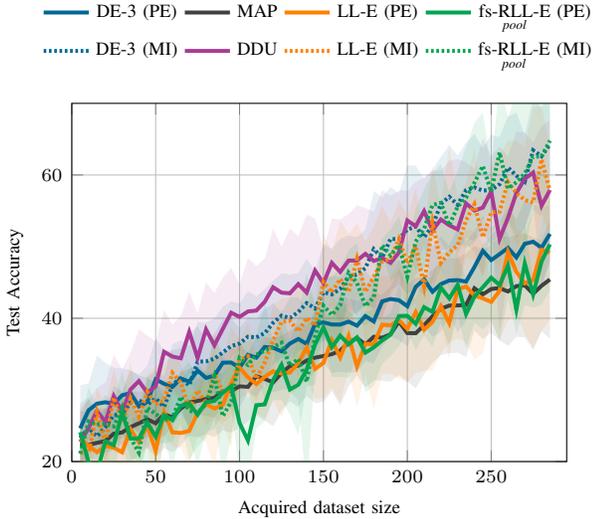

\subsection{Uncertainty decomposition for active learning} \label{sec:active learning}
Next, we evaluate the performance of our \acrshort{LLPOVI} and \acrshort{fLLPOVI} uncertainty estimates on an active learning task proposed in \cite{mukhoti2023deep}. Given a small number of initial training samples and a large pool of unlabeled data, the aim is to select the most informative data points, which are subsequently used to retrain the network. 
We start with an initial training set of 20 samples and a pool of clean and ambiguous MNIST samples. The ratio of clean to ambiguous is 1:60. In each iteration, we add the 5 samples with the highest epistemic uncertainty in the pool set. 
Thus, disentangling aleatoric and epistemic uncertainty is essential to select informative samples that improve prediction accuracy.
Fig. \ref{fig:active_learning} shows that DDU, \acrshort{LLPOVI} and \acrshort{fLLPOVI} are all able to compete with \acrshortpl{DE}. All methods achieve a similar accuracy on the test set at the end of the iterations. We clearly see the importance of using epistemic uncertainty to avoid selecting ambiguous samples with high aleatoric uncertainty.

\pgfplotstableread[col sep=comma]{sections/graphs/OOD_CIFAR10/OOD_wide_cifar10_auroc.csv}\tableOOD
\pgfplotstableread[col sep=comma]{sections/graphs/OOD_CIFAR10/ID_wide_cifar10_accuracy.csv}\tableIDaccuracy
\pgfplotstableread[col sep=comma]{sections/graphs/OOD_CIFAR10/ID_wide_cifar10_nll.csv}\tableIDnll
\pgfplotstableread[col sep=comma]{sections/graphs/OOD_CIFAR10/ID_wide_cifar10_ece.csv}\tableIDece

\pgfplotstableread[col sep=comma]{sections/graphs/OOD_CIFAR10/OOD_wide_cifar10_auroc_POVI_SGD.csv}\tableOODSGD
\pgfplotstableread[col sep=comma]{sections/graphs/OOD_CIFAR10/ID_wide_cifar10_accuracy_POVI_SGD.csv}\tableIDaccuracySGD
\pgfplotstableread[col sep=comma]{sections/graphs/OOD_CIFAR10/ID_wide_cifar10_nll_POVI_SGD.csv}\tableIDnllSGD
\pgfplotstableread[col sep=comma]{sections/graphs/OOD_CIFAR10/ID_wide_cifar10_ece_POVI_SGD.csv}\tableIDeceSGD

\pgfplotstableread[col sep=comma]{sections/graphs/OOD_CIFAR10/OOD_wide_cifar10_auroc_POVI_kde_WGD.csv}\tableOODWGD
\pgfplotstableread[col sep=comma]{sections/graphs/OOD_CIFAR10/ID_wide_cifar10_accuracy_POVI_kde_WGD.csv}\tableIDaccuracyWGD
\pgfplotstableread[col sep=comma]{sections/graphs/OOD_CIFAR10/ID_wide_cifar10_nll_POVI_kde_WGD.csv}\tableIDnllWGD
\pgfplotstableread[col sep=comma]{sections/graphs/OOD_CIFAR10/ID_wide_cifar10_ece_POVI_kde_WGD.csv}\tableIDeceWGD

\pgfplotstableread[col sep=comma]{sections/graphs/OOD_CIFAR10/OOD_wide_cifar10_auroc_POVI_kde_f_WGD_coll_cifar100.csv}\tableOODcifar
\pgfplotstableread[col sep=comma]{sections/graphs/OOD_CIFAR10/ID_wide_cifar10_accuracy_POVI_kde_f_WGD_coll_cifar100.csv}\tableIDaccuracycifar
\pgfplotstableread[col sep=comma]{sections/graphs/OOD_CIFAR10/ID_wide_cifar10_nll_POVI_kde_f_WGD_coll_cifar100.csv}\tableIDnllcifar
\pgfplotstableread[col sep=comma]{sections/graphs/OOD_CIFAR10/ID_wide_cifar10_ece_POVI_kde_f_WGD_coll_cifar100.csv}\tableIDececifar

\pgfplotstableread[col sep=comma]{sections/graphs/OOD_CIFAR10/OOD_wide_cifar10_auroc_POVI_kde_f_WGD_coll_cifar10_.csv}\tableOODcifarTen
\pgfplotstableread[col sep=comma]{sections/graphs/OOD_CIFAR10/ID_wide_cifar10_accuracy_POVI_kde_f_WGD_coll_cifar10_.csv}\tableIDaccuracycifarTen
\pgfplotstableread[col sep=comma]{sections/graphs/OOD_CIFAR10/ID_wide_cifar10_nll_POVI_kde_f_WGD_coll_cifar10_.csv}\tableIDnllcifarTen
\pgfplotstableread[col sep=comma]{sections/graphs/OOD_CIFAR10/ID_wide_cifar10_ece_POVI_kde_f_WGD_coll_cifar10_.csv}\tableIDececifarTen

\pgfplotstableread[col sep=comma]{sections/graphs/OOD_CIFAR10/OOD_wide_cifar10_auroc_POVI_kde_f_WGD_coll_tinyimagenet.csv}\tableOODFtiny
\pgfplotstableread[col sep=comma]{sections/graphs/OOD_CIFAR10/ID_wide_cifar10_accuracy_POVI_kde_f_WGD_coll_tinyimagenet.csv}\tableIDaccuracyFtiny
\pgfplotstableread[col sep=comma]{sections/graphs/OOD_CIFAR10/ID_wide_cifar10_nll_POVI_kde_f_WGD_coll_tinyimagenet.csv}\tableIDnllFtiny
\pgfplotstableread[col sep=comma]{sections/graphs/OOD_CIFAR10/ID_wide_cifar10_ece_POVI_kde_f_WGD_coll_tinyimagenet.csv}\tableIDeceFtiny

\pgfplotstableread[col sep=comma]{sections/graphs/OOD_CIFAR10/OOD_wide_cifar10_auroc_POVI_kde_f_WGD_coll_texture.csv}\tableOODFtext
\pgfplotstableread[col sep=comma]{sections/graphs/OOD_CIFAR10/ID_wide_cifar10_accuracy_POVI_kde_f_WGD_coll_texture.csv}\tableIDaccuracyFtext
\pgfplotstableread[col sep=comma]{sections/graphs/OOD_CIFAR10/ID_wide_cifar10_nll_POVI_kde_f_WGD_coll_texture.csv}\tableIDnllFtext
\pgfplotstableread[col sep=comma]{sections/graphs/OOD_CIFAR10/ID_wide_cifar10_ece_POVI_kde_f_WGD_coll_texture.csv}\tableIDeceFtext

\pgfplotstableread[col sep=comma]{sections/graphs/OOD_CIFAR10/OOD_wide_cifar10_auroc_POVI_kde_f_WGD_coll_cifar10__patches_std_16.csv}\tableOODPatchsixteen
\pgfplotstableread[col sep=comma]{sections/graphs/OOD_CIFAR10/ID_wide_cifar10_accuracy_POVI_kde_f_WGD_coll_cifar10__patches_std_16.csv}\tableIDaccuracyPatchsixteen
\pgfplotstableread[col sep=comma]{sections/graphs/OOD_CIFAR10/ID_wide_cifar10_nll_POVI_kde_f_WGD_coll_cifar10__patches_std_16.csv}\tableIDnllPatchsixteen
\pgfplotstableread[col sep=comma]{sections/graphs/OOD_CIFAR10/ID_wide_cifar10_ece_POVI_kde_f_WGD_coll_cifar10__patches_std_16.csv}\tableIDecePatchsixteen

\pgfplotstableread[col sep=comma]{sections/graphs/OOD_CIFAR10/OOD_wide_cifar10_auroc_POVI_kde_f_WGD_coll_cifar10__patches_std_32.csv}\tableOODPatchthirtytwo
\pgfplotstableread[col sep=comma]{sections/graphs/OOD_CIFAR10/ID_wide_cifar10_accuracy_POVI_kde_f_WGD_coll_cifar10__patches_std_32.csv}\tableIDaccuracyPatchthirtytwo
\pgfplotstableread[col sep=comma]{sections/graphs/OOD_CIFAR10/ID_wide_cifar10_nll_POVI_kde_f_WGD_coll_cifar10__patches_std_32.csv}\tableIDnllPatchthirtytwo
\pgfplotstableread[col sep=comma]{sections/graphs/OOD_CIFAR10/ID_wide_cifar10_ece_POVI_kde_f_WGD_coll_cifar10__patches_std_32.csv}\tableIDecePatchthirtytwo

\pgfplotstableread[col sep=comma]{sections/graphs/OOD_CIFAR10/OOD_wide_cifar10_auroc_POVI_kde_f_WGD_coll_cifar10__patches_std_8.csv}\tableOODPatcheight
\pgfplotstableread[col sep=comma]{sections/graphs/OOD_CIFAR10/ID_wide_cifar10_accuracy_POVI_kde_f_WGD_coll_cifar10__patches_std_8.csv}\tableIDaccuracyPatcheight
\pgfplotstableread[col sep=comma]{sections/graphs/OOD_CIFAR10/ID_wide_cifar10_nll_POVI_kde_f_WGD_coll_cifar10__patches_std_8.csv}\tableIDnllPatcheight
\pgfplotstableread[col sep=comma]{sections/graphs/OOD_CIFAR10/ID_wide_cifar10_ece_POVI_kde_f_WGD_coll_cifar10__patches_std_8.csv}\tableIDecePatcheight

\renewcommand{\fetchAndStore}[4]{%
    \pgfplotstablegetelem{#1}{#2}\of{#3}
    \expandafter\xdef\csname #2Mean#1#4\endcsname{\pgfplotsretval}
    \pgfplotstablegetelem{#1}{#2_std}\of{#3}
    \expandafter\xdef\csname #2Std#1#4\endcsname{\pgfplotsretval}
}

\foreach \method in {softmax,logits,logits_GMM,softmax_laplace_MI,softmax_SNGP_PE,softmax_en_PE,softmax_en_MI} {
    \foreach \intensity in {0,1,2,3,4,5} { 
        \fetchAndStore{\intensity}{\method}{\tableOOD}{OOD}
    }
}

\foreach \method in {softmax,logits,logits_GMM,softmax_laplace_MI,softmax_SNGP_PE,softmax_en_PE,softmax_en_MI} {
    \fetchAndStore{0}{\method}{\tableIDaccuracy}{accuracy}
    \fetchAndStore{0}{\method}{\tableIDnll}{nll}
    \fetchAndStore{0}{\method}{\tableIDece}{ece}
}

\foreach \method in {softmax_POVI_MI} {
    \foreach \intensity in {0,1,2,3,4,5} { 
        \fetchAndStore{\intensity}{\method}{\tableOODSGD}{OODSGD}
        \fetchAndStore{\intensity}{\method}{\tableOODWGD}{OODWGD}
        \fetchAndStore{\intensity}{\method}{\tableOODFtiny}{OODFtiny}
        \fetchAndStore{\intensity}{\method}{\tableOODcifar}{OODcifar}
        \fetchAndStore{\intensity}{\method}{\tableOODcifarTen}{OODcifarTen}
        \fetchAndStore{\intensity}{\method}{\tableOODFtext}{OODFtext}
        \fetchAndStore{\intensity}{\method}{\tableOODPatchsixteen}{OODPatchsixteen}
        \fetchAndStore{\intensity}{\method}{\tableOODPatchthirtytwo}{OODPatchthirtytwo}
        \fetchAndStore{\intensity}{\method}{\tableOODPatcheight}{OODPatcheight}
    }
}
\foreach \method in {softmax_POVI_MI} {
    \fetchAndStore{0}{\method}{\tableIDaccuracySGD}{accuracySGD}
    \fetchAndStore{0}{\method}{\tableIDnllSGD}{nllSGD}
    \fetchAndStore{0}{\method}{\tableIDeceSGD}{eceSGD}

    \fetchAndStore{0}{\method}{\tableIDaccuracyWGD}{accuracyWGD}
    \fetchAndStore{0}{\method}{\tableIDnllWGD}{nllWGD}
    \fetchAndStore{0}{\method}{\tableIDeceWGD}{eceWGD}
    
    \fetchAndStore{0}{\method}{\tableIDaccuracycifar}{accuracycifar}
    \fetchAndStore{0}{\method}{\tableIDnllcifar}{nllcifar}
    \fetchAndStore{0}{\method}{\tableIDececifar}{ececifar}
    
    \fetchAndStore{0}{\method}{\tableIDaccuracycifarTen}{accuracycifarTen}
    \fetchAndStore{0}{\method}{\tableIDnllcifarTen}{nllcifarTen}
    \fetchAndStore{0}{\method}{\tableIDececifarTen}{ececifarTen}    
    
    \fetchAndStore{0}{\method}{\tableIDaccuracyFtiny}{accuracyFtiny}
    \fetchAndStore{0}{\method}{\tableIDnllFtiny}{nllFtiny}
    \fetchAndStore{0}{\method}{\tableIDeceFtiny}{eceFtiny}
    
    \fetchAndStore{0}{\method}{\tableIDaccuracyFtext}{accuracyFtext}
    \fetchAndStore{0}{\method}{\tableIDnllFtext}{nllFtext}
    \fetchAndStore{0}{\method}{\tableIDeceFtext}{eceFtext}

    \fetchAndStore{0}{\method}{\tableIDaccuracyPatchsixteen}{accuracyPatchsixteen}
    \fetchAndStore{0}{\method}{\tableIDnllPatchsixteen}{nllPatchsixteen}
    \fetchAndStore{0}{\method}{\tableIDecePatchsixteen}{ecePatchsixteen}
    
    \fetchAndStore{0}{\method}{\tableIDaccuracyPatchthirtytwo}{accuracyPatchthirtytwo}
    \fetchAndStore{0}{\method}{\tableIDnllPatchthirtytwo}{nllPatchthirtytwo}
    \fetchAndStore{0}{\method}{\tableIDecePatchthirtytwo}{ecePatchthirtytwo}
    
    \fetchAndStore{0}{\method}{\tableIDaccuracyPatcheight}{accuracyPatcheight}
    \fetchAndStore{0}{\method}{\tableIDnllPatcheight}{nllPatcheight}
    \fetchAndStore{0}{\method}{\tableIDecePatcheight}{ecePatcheight}
}

\addtolength{\tabcolsep}{-0.4em}

\begin{table*}[]
\centering
\caption{Comparison of uncertainty estimation for \acrshort{ID} calibration, and \acrshort{OOD} detection on CIFAR10. Mean and standard deviation are computed over 10 runs. Best results are in bold, second best are underlined.}
\label{tab:my-table}
\resizebox{\textwidth}{!}{%
\begin{tabular}{@{}lcccccccccc@{}}
\toprule
\multirow{2}{*}{\textbf{Method}} & \multirow{2}{*}{\textsc{Acc.} $\uparrow$ [\%]} & \multirow{2}{*}{\textsc{NLL} $\downarrow$ [\%]} & \multirow{2}{*}{\textsc{ECE} $\downarrow$ [\%]} & \multicolumn{6}{c}{\textsc{OOD} \textsc{Auroc} $\uparrow$ [\%]} &  \\ \cmidrule(lr){5-10}
 &  &  &  & \textit{Cifar100} & \textit{TinyIm.} & \textit{Places365} & \textit{Texture} & \textit{SVHN} & \textit{FakeData} &  \\ \midrule
MAP & $\underline{\csname softmaxMean0accuracy\endcsname _{\pm \csname softmaxStd0accuracy\endcsname}}$ & ${\csname softmaxMean0nll\endcsname _{\pm \csname softmaxStd0nll\endcsname}}$ & ${\csname softmaxMean0ece\endcsname _{\pm \csname softmaxStd0ece\endcsname}}$ & ${\csname softmaxMean0OOD\endcsname _{\pm \csname softmaxStd0OOD\endcsname}}$ & ${\csname softmaxMean1OOD\endcsname _{\pm \csname softmaxStd1OOD\endcsname}}$ & ${\csname softmaxMean2OOD\endcsname _{\pm \csname softmaxStd2OOD\endcsname}}$ & ${\csname softmaxMean3OOD\endcsname _{\pm \csname softmaxStd3OOD\endcsname}}$ & ${\csname softmaxMean4OOD\endcsname _{\pm \csname softmaxStd4OOD\endcsname}}$ & ${\csname softmaxMean5OOD\endcsname _{\pm \csname softmaxStd5OOD\endcsname}}$ &  \\ \cmidrule(r){1-10}
DDU & $\underline{\csname softmaxMean0accuracy\endcsname _{\pm \csname softmaxStd0accuracy\endcsname}}$ & ${\csname softmaxMean0nll\endcsname _{\pm \csname softmaxStd0nll\endcsname}}$ & ${\csname softmaxMean0ece\endcsname _{\pm \csname softmaxStd0ece\endcsname}}$ & ${\csname logits_GMMMean0OOD\endcsname _{\pm \csname logits_GMMStd0OOD\endcsname}}$ & ${\csname logits_GMMMean1OOD\endcsname _{\pm \csname logits_GMMStd1OOD\endcsname}}$ & ${\csname logits_GMMMean2OOD\endcsname _{\pm \csname logits_GMMStd2OOD\endcsname}}$ & $\mathbf{\csname logits_GMMMean3OOD\endcsname _{\pm \csname logits_GMMStd3OOD\endcsname}}$ & $\mathbf{\csname logits_GMMMean4OOD\endcsname _{\pm \csname logits_GMMStd4OOD\endcsname}}$ & $\mathbf{\csname logits_GMMMean5OOD\endcsname _{\pm \csname logits_GMMStd5OOD\endcsname}}$ &  \\
SNGP & ${\csname softmax_SNGP_PEMean0accuracy\endcsname _{\pm \csname softmax_SNGP_PEStd0accuracy\endcsname}}$ & ${\csname softmax_SNGP_PEMean0nll\endcsname _{\pm \csname softmax_SNGP_PEStd0nll\endcsname}}$ & ${\csname softmax_SNGP_PEMean0ece\endcsname _{\pm \csname softmax_SNGP_PEStd0ece\endcsname}}$ & ${\csname softmax_SNGP_PEMean0OOD\endcsname _{\pm \csname softmax_SNGP_PEStd0OOD\endcsname}}$ & ${\csname softmax_SNGP_PEMean1OOD\endcsname _{\pm \csname softmax_SNGP_PEStd1OOD\endcsname}}$ & $\mathbf{\csname softmax_SNGP_PEMean2OOD\endcsname _{\pm \csname softmax_SNGP_PEStd2OOD\endcsname}}$ & ${\csname softmax_SNGP_PEMean3OOD\endcsname _{\pm \csname softmax_SNGP_PEStd3OOD\endcsname}}$ & ${\csname softmax_SNGP_PEMean4OOD\endcsname _{\pm \csname softmax_SNGP_PEStd4OOD\endcsname}}$ & ${\csname softmax_SNGP_PEMean5OOD\endcsname _{\pm \csname softmax_SNGP_PEStd5OOD\endcsname}}$ &  \\
LL-Laplace & ${\csname softmax_laplace_MIMean0accuracy\endcsname _{\pm \csname softmax_laplace_MIStd0accuracy\endcsname}}$ & ${\csname softmax_laplace_MIMean0nll\endcsname _{\pm \csname softmax_laplace_MIStd0nll\endcsname}}$ & ${\csname softmax_laplace_MIMean0ece\endcsname _{\pm \csname softmax_laplace_MIStd0ece\endcsname}}$ & ${\csname softmax_laplace_MIMean0OOD\endcsname _{\pm \csname softmax_laplace_MIStd0OOD\endcsname}}$ & ${\csname softmax_laplace_MIMean1OOD\endcsname _{\pm \csname softmax_laplace_MIStd1OOD\endcsname}}$ & ${\csname softmax_laplace_MIMean2OOD\endcsname _{\pm \csname softmax_laplace_MIStd2OOD\endcsname}}$ & ${\csname softmax_laplace_MIMean3OOD\endcsname _{\pm \csname softmax_laplace_MIStd3OOD\endcsname}}$ & ${\csname softmax_laplace_MIMean4OOD\endcsname _{\pm \csname softmax_laplace_MIStd4OOD\endcsname}}$ & ${\csname softmax_laplace_MIMean5OOD\endcsname _{\pm \csname softmax_laplace_MIStd5OOD\endcsname}}$ &  \\ \midrule
\acrshort{LLPOVI} \emph{(ours)} & ${\csname softmax_POVI_MIMean0accuracySGD\endcsname _{\pm \csname softmax_POVI_MIStd0accuracySGD\endcsname}}$ & $\underline{\csname softmax_POVI_MIMean0nllSGD\endcsname _{\pm \csname softmax_POVI_MIStd0nllSGD\endcsname}}$ & ${\csname softmax_POVI_MIMean0eceSGD\endcsname _{\pm \csname softmax_POVI_MIStd0eceSGD\endcsname}}$ & ${\csname softmax_POVI_MIMean0OODSGD\endcsname _{\pm \csname softmax_POVI_MIStd0OODSGD\endcsname}}$ & ${\csname softmax_POVI_MIMean1OODSGD\endcsname _{\pm \csname softmax_POVI_MIStd1OODSGD\endcsname}}$ & ${\csname softmax_POVI_MIMean2OODSGD\endcsname _{\pm \csname softmax_POVI_MIStd2OODSGD\endcsname}}$ & ${\csname softmax_POVI_MIMean3OODSGD\endcsname _{\pm \csname softmax_POVI_MIStd3OODSGD\endcsname}}$ & ${\csname softmax_POVI_MIMean4OODSGD\endcsname _{\pm \csname softmax_POVI_MIStd4OODSGD\endcsname}}$ & ${\csname softmax_POVI_MIMean5OODSGD\endcsname _{\pm \csname softmax_POVI_MIStd5OODSGD\endcsname}}$ &  \\
\acrshort{RLLPOVI} \emph{(ours)}  & ${\csname softmax_POVI_MIMean0accuracyWGD\endcsname _{\pm \csname softmax_POVI_MIStd0accuracyWGD\endcsname}}$ & ${\csname softmax_POVI_MIMean0nllWGD\endcsname _{\pm \csname softmax_POVI_MIStd0nllWGD\endcsname}}$ & ${\csname softmax_POVI_MIMean0eceWGD\endcsname _{\pm \csname softmax_POVI_MIStd0eceWGD\endcsname}}$ & ${\csname softmax_POVI_MIMean0OODWGD\endcsname _{\pm \csname softmax_POVI_MIStd0OODWGD\endcsname}}$ & ${\csname softmax_POVI_MIMean1OODWGD\endcsname _{\pm \csname softmax_POVI_MIStd1OODWGD\endcsname}}$ & ${\csname softmax_POVI_MIMean2OODWGD\endcsname _{\pm \csname softmax_POVI_MIStd2OODWGD\endcsname}}$ & ${\csname softmax_POVI_MIMean3OODWGD\endcsname _{\pm \csname softmax_POVI_MIStd3OODWGD\endcsname}}$ & ${\csname softmax_POVI_MIMean4OODWGD\endcsname _{\pm \csname softmax_POVI_MIStd4OODWGD\endcsname}}$ & ${\csname softmax_POVI_MIMean5OODWGD\endcsname _{\pm \csname softmax_POVI_MIStd5OODWGD\endcsname}}$ &  \\
\acrshort{fLLPOVI} \emph{(ours)} &  &  &  &  &  &  &  &  &  &  \\
\textit{~~~+ Cifar10} & ${\csname softmax_POVI_MIMean0accuracycifarTen\endcsname _{\pm \csname softmax_POVI_MIStd0accuracycifarTen\endcsname}}$ & ${\csname softmax_POVI_MIMean0nllcifarTen\endcsname _{\pm \csname softmax_POVI_MIStd0nllcifarTen\endcsname}}$ & ${\csname softmax_POVI_MIMean0ececifarTen\endcsname _{\pm \csname softmax_POVI_MIStd0ececifarTen\endcsname}}$ & ${\csname softmax_POVI_MIMean0OODcifarTen\endcsname _{\pm \csname softmax_POVI_MIStd0OODcifarTen\endcsname}}$ & ${\csname softmax_POVI_MIMean1OODcifarTen\endcsname _{\pm \csname softmax_POVI_MIStd1OODcifarTen\endcsname}}$ & ${\csname softmax_POVI_MIMean2OODcifarTen\endcsname _{\pm \csname softmax_POVI_MIStd2OODcifarTen\endcsname}}$ & ${\csname softmax_POVI_MIMean3OODcifarTen\endcsname _{\pm \csname softmax_POVI_MIStd3OODcifarTen\endcsname}}$ & ${\csname softmax_POVI_MIMean4OODcifarTen\endcsname _{\pm \csname softmax_POVI_MIStd4OODcifarTen\endcsname}}$ & ${\csname softmax_POVI_MIMean5OODcifarTen\endcsname _{\pm \csname softmax_POVI_MIStd5OODcifarTen\endcsname}}$ &  \\
\textit{~~~+ Cifar100} & ${\csname softmax_POVI_MIMean0accuracycifar\endcsname _{\pm \csname softmax_POVI_MIStd0accuracycifar\endcsname}}$ & ${\csname softmax_POVI_MIMean0nllcifar\endcsname _{\pm \csname softmax_POVI_MIStd0nllcifar\endcsname}}$ & ${\csname softmax_POVI_MIMean0ececifar\endcsname _{\pm \csname softmax_POVI_MIStd0ececifar\endcsname}}$ & $\mathbf{\csname softmax_POVI_MIMean0OODcifar\endcsname _{\pm \csname softmax_POVI_MIStd0OODcifar\endcsname}}$ & $\mathbf{\csname softmax_POVI_MIMean1OODcifar\endcsname _{\pm \csname softmax_POVI_MIStd1OODcifar\endcsname}}$ & ${\csname softmax_POVI_MIMean2OODcifar\endcsname _{\pm \csname softmax_POVI_MIStd2OODcifar\endcsname}}$ & ${\csname softmax_POVI_MIMean3OODcifar\endcsname _{\pm \csname softmax_POVI_MIStd3OODcifar\endcsname}}$ & ${\csname softmax_POVI_MIMean4OODcifar\endcsname _{\pm \csname softmax_POVI_MIStd4OODcifar\endcsname}}$ & ${\csname softmax_POVI_MIMean5OODcifar\endcsname _{\pm \csname softmax_POVI_MIStd5OODcifar\endcsname}}$ &  \\
\textit{~~~+ TinyImagenet} & ${\csname softmax_POVI_MIMean0accuracyFtiny\endcsname _{\pm \csname softmax_POVI_MIStd0accuracyFtiny\endcsname}}$ & ${\csname softmax_POVI_MIMean0nllFtiny\endcsname _{\pm \csname softmax_POVI_MIStd0nllFtiny\endcsname}}$ & ${\csname softmax_POVI_MIMean0eceFtiny\endcsname _{\pm \csname softmax_POVI_MIStd0eceFtiny\endcsname}}$ & ${\csname softmax_POVI_MIMean0OODFtiny\endcsname _{\pm \csname softmax_POVI_MIStd0OODFtiny\endcsname}}$ & ${\csname softmax_POVI_MIMean1OODFtiny\endcsname _{\pm \csname softmax_POVI_MIStd1OODFtiny\endcsname}}$ & $\underline{\csname softmax_POVI_MIMean2OODFtiny\endcsname _{\pm \csname softmax_POVI_MIStd2OODFtiny\endcsname}}$ & ${\csname softmax_POVI_MIMean3OODFtiny\endcsname _{\pm \csname softmax_POVI_MIStd3OODFtiny\endcsname}}$ & ${\csname softmax_POVI_MIMean4OODFtiny\endcsname _{\pm \csname softmax_POVI_MIStd4OODFtiny\endcsname}}$ & ${\csname softmax_POVI_MIMean5OODFtiny\endcsname _{\pm \csname softmax_POVI_MIStd5OODFtiny\endcsname}}$ &  \\
\textit{~~~+ Texture} & ${\csname softmax_POVI_MIMean0accuracyFtext\endcsname _{\pm \csname softmax_POVI_MIStd0accuracyFtext\endcsname}}$ & ${\csname softmax_POVI_MIMean0nllFtext\endcsname _{\pm \csname softmax_POVI_MIStd0nllFtext\endcsname}}$ & ${\csname softmax_POVI_MIMean0eceFtext\endcsname _{\pm \csname softmax_POVI_MIStd0eceFtext\endcsname}}$ & ${\csname softmax_POVI_MIMean0OODFtext\endcsname _{\pm \csname softmax_POVI_MIStd0OODFtext\endcsname}}$ & ${\csname softmax_POVI_MIMean1OODFtext\endcsname _{\pm \csname softmax_POVI_MIStd1OODFtext\endcsname}}$ & ${\csname softmax_POVI_MIMean2OODFtext\endcsname _{\pm \csname softmax_POVI_MIStd2OODFtext\endcsname}}$ & $\underline{\csname softmax_POVI_MIMean3OODFtext\endcsname _{\pm \csname softmax_POVI_MIStd3OODFtext\endcsname}}$ & ${\csname softmax_POVI_MIMean4OODFtext\endcsname _{\pm \csname softmax_POVI_MIStd4OODFtext\endcsname}}$ & ${\csname softmax_POVI_MIMean5OODFtext\endcsname _{\pm \csname softmax_POVI_MIStd5OODFtext\endcsname}}$ &  \\
\textit{~~~+ Patches-32} & ${\csname softmax_POVI_MIMean0accuracyPatchthirtytwo\endcsname _{\pm \csname softmax_POVI_MIStd0accuracyPatchthirtytwo\endcsname}}$ & ${\csname softmax_POVI_MIMean0nllPatchthirtytwo\endcsname _{\pm \csname softmax_POVI_MIStd0nllPatchthirtytwo\endcsname}}$ & $\underline{\csname softmax_POVI_MIMean0ecePatchthirtytwo\endcsname _{\pm \csname softmax_POVI_MIStd0ecePatchthirtytwo\endcsname}}$ & ${\csname softmax_POVI_MIMean0OODPatchthirtytwo\endcsname _{\pm \csname softmax_POVI_MIStd0OODPatchthirtytwo\endcsname}}$ & ${\csname softmax_POVI_MIMean1OODPatchthirtytwo\endcsname _{\pm \csname softmax_POVI_MIStd1OODPatchthirtytwo\endcsname}}$ & ${\csname softmax_POVI_MIMean2OODPatchthirtytwo\endcsname _{\pm \csname softmax_POVI_MIStd2OODPatchthirtytwo\endcsname}}$ & ${\csname softmax_POVI_MIMean3OODPatchthirtytwo\endcsname _{\pm \csname softmax_POVI_MIStd3OODPatchthirtytwo\endcsname}}$ & ${\csname softmax_POVI_MIMean4OODPatchthirtytwo\endcsname _{\pm \csname softmax_POVI_MIStd4OODPatchthirtytwo\endcsname}}$ & ${\csname softmax_POVI_MIMean5OODPatchthirtytwo\endcsname _{\pm \csname softmax_POVI_MIStd5OODPatchthirtytwo\endcsname}}$ &  \\
\textit{~~~+ Patches-16} & ${\csname softmax_POVI_MIMean0accuracyPatchsixteen\endcsname _{\pm \csname softmax_POVI_MIStd0accuracyPatchsixteen\endcsname}}$ & ${\csname softmax_POVI_MIMean0nllPatchsixteen\endcsname _{\pm \csname softmax_POVI_MIStd0nllPatchsixteen\endcsname}}$ & $\mathbf{\csname softmax_POVI_MIMean0ecePatchsixteen\endcsname _{\pm \csname softmax_POVI_MIStd0ecePatchsixteen\endcsname}}$ & ${\csname softmax_POVI_MIMean0OODPatchsixteen\endcsname _{\pm \csname softmax_POVI_MIStd0OODPatchsixteen\endcsname}}$ & ${\csname softmax_POVI_MIMean1OODPatchsixteen\endcsname _{\pm \csname softmax_POVI_MIStd1OODPatchsixteen\endcsname}}$ & ${\csname softmax_POVI_MIMean2OODPatchsixteen\endcsname _{\pm \csname softmax_POVI_MIStd2OODPatchsixteen\endcsname}}$ & ${\csname softmax_POVI_MIMean3OODPatchsixteen\endcsname _{\pm \csname softmax_POVI_MIStd3OODPatchsixteen\endcsname}}$ & ${\csname softmax_POVI_MIMean4OODPatchsixteen\endcsname _{\pm \csname softmax_POVI_MIStd4OODPatchsixteen\endcsname}}$ & ${\csname softmax_POVI_MIMean5OODPatchsixteen\endcsname _{\pm \csname softmax_POVI_MIStd5OODPatchsixteen\endcsname}}$ &  \\ 
\midrule
DE-5 & $\mathbf{\csname softmax_en_MIMean0accuracy\endcsname _{\pm \csname softmax_en_MIStd0accuracy\endcsname}}$ & $\mathbf{\csname softmax_en_MIMean0nll\endcsname _{\pm \csname softmax_en_MIStd0nll\endcsname}}$ & ${\csname softmax_en_MIMean0ece\endcsname _{\pm \csname softmax_en_MIStd0ece\endcsname}}$ & $\underline{\csname softmax_en_MIMean0OOD\endcsname _{\pm \csname softmax_en_MIStd0OOD\endcsname}}$ & $\underline{\csname softmax_en_MIMean1OOD\endcsname _{\pm \csname softmax_en_MIStd1OOD\endcsname}}$ & ${\csname softmax_en_MIMean2OOD\endcsname _{\pm \csname softmax_en_MIStd2OOD\endcsname}}$ & ${\csname softmax_en_MIMean3OOD\endcsname _{\pm \csname softmax_en_MIStd3OOD\endcsname}}$ & $\underline{\csname softmax_en_MIMean4OOD\endcsname _{\pm \csname softmax_en_MIStd4OOD\endcsname}}$ & $\underline{\csname softmax_en_MIMean5OOD\endcsname _{\pm \csname softmax_en_MIStd5OOD\endcsname}}$ &  \\ \bottomrule
\end{tabular}%
}\label{tab:OOD:cifar10}
\end{table*}
\pgfplotstableread[col sep=comma]{sections/graphs/OOD_CIFAR100/OOD_cifar100_auroc.csv}\tableOOD
\pgfplotstableread[col sep=comma]{sections/graphs/OOD_CIFAR100/ID_cifar100_accuracy.csv}\tableIDaccuracy
\pgfplotstableread[col sep=comma]{sections/graphs/OOD_CIFAR100/ID_cifar100_nll.csv}\tableIDnll
\pgfplotstableread[col sep=comma]{sections/graphs/OOD_CIFAR100/ID_cifar100_ece.csv}\tableIDece

\pgfplotstableread[col sep=comma]{sections/graphs/OOD_CIFAR100/OOD_cifar100_auroc_POVI_SGD.csv}\tableOODSGD
\pgfplotstableread[col sep=comma]{sections/graphs/OOD_CIFAR100/ID_cifar100_accuracy_POVI_SGD.csv}\tableIDaccuracySGD
\pgfplotstableread[col sep=comma]{sections/graphs/OOD_CIFAR100/ID_cifar100_nll_POVI_SGD.csv}\tableIDnllSGD
\pgfplotstableread[col sep=comma]{sections/graphs/OOD_CIFAR100/ID_cifar100_ece_POVI_SGD.csv}\tableIDeceSGD

\pgfplotstableread[col sep=comma]{sections/graphs/OOD_CIFAR100/OOD_cifar100_auroc_POVI_kde_WGD.csv}\tableOODWGD
\pgfplotstableread[col sep=comma]{sections/graphs/OOD_CIFAR100/ID_cifar100_accuracy_POVI_kde_WGD.csv}\tableIDaccuracyWGD
\pgfplotstableread[col sep=comma]{sections/graphs/OOD_CIFAR100/ID_cifar100_nll_POVI_kde_WGD.csv}\tableIDnllWGD
\pgfplotstableread[col sep=comma]{sections/graphs/OOD_CIFAR100/ID_cifar100_ece_POVI_kde_WGD.csv}\tableIDeceWGD

\pgfplotstableread[col sep=comma]{sections/graphs/OOD_CIFAR100/OOD_cifar100_auroc_POVI_kde_f_WGD_coll_tinyimagenet.csv}\tableOODFtiny
\pgfplotstableread[col sep=comma]{sections/graphs/OOD_CIFAR100/ID_cifar100_accuracy_POVI_kde_f_WGD_coll_tinyimagenet.csv}\tableIDaccuracyFtiny
\pgfplotstableread[col sep=comma]{sections/graphs/OOD_CIFAR100/ID_cifar100_nll_POVI_kde_f_WGD_coll_tinyimagenet.csv}\tableIDnllFtiny
\pgfplotstableread[col sep=comma]{sections/graphs/OOD_CIFAR100/ID_cifar100_ece_POVI_kde_f_WGD_coll_tinyimagenet.csv}\tableIDeceFtiny

\pgfplotstableread[col sep=comma]{sections/graphs/OOD_CIFAR100/OOD_cifar100_auroc_POVI_kde_f_WGD_coll_texture.csv}\tableOODFtext
\pgfplotstableread[col sep=comma]{sections/graphs/OOD_CIFAR100/ID_cifar100_accuracy_POVI_kde_f_WGD_coll_texture.csv}\tableIDaccuracyFtext
\pgfplotstableread[col sep=comma]{sections/graphs/OOD_CIFAR100/ID_cifar100_nll_POVI_kde_f_WGD_coll_texture.csv}\tableIDnllFtext
\pgfplotstableread[col sep=comma]{sections/graphs/OOD_CIFAR100/ID_cifar100_ece_POVI_kde_f_WGD_coll_texture.csv}\tableIDeceFtext

\pgfplotstableread[col sep=comma]{sections/graphs/OOD_CIFAR100/OOD_wide_cifar100_auroc_POVI_kde_f_WGD_coll_cifar100.csv}\tableOODFcifar
\pgfplotstableread[col sep=comma]{sections/graphs/OOD_CIFAR100/ID_wide_cifar100_accuracy_POVI_kde_f_WGD_coll_cifar100.csv}\tableIDaccuracyFcifar
\pgfplotstableread[col sep=comma]{sections/graphs/OOD_CIFAR100/ID_wide_cifar100_nll_POVI_kde_f_WGD_coll_cifar100.csv}\tableIDnllFcifar
\pgfplotstableread[col sep=comma]{sections/graphs/OOD_CIFAR100/ID_wide_cifar100_ece_POVI_kde_f_WGD_coll_cifar100.csv}\tableIDeceFcifar

\pgfplotstableread[col sep=comma]{sections/graphs/OOD_CIFAR100/OOD_cifar100_auroc_POVI_kde_f_WGD_coll_cifar100_patches_std_16.csv}\tableOODPatchsixteen
\pgfplotstableread[col sep=comma]{sections/graphs/OOD_CIFAR100/ID_cifar100_accuracy_POVI_kde_f_WGD_coll_cifar100_patches_std_16.csv}\tableIDaccuracyPatchsixteen
\pgfplotstableread[col sep=comma]{sections/graphs/OOD_CIFAR100/ID_cifar100_nll_POVI_kde_f_WGD_coll_cifar100_patches_std_16.csv}\tableIDnllPatchsixteen
\pgfplotstableread[col sep=comma]{sections/graphs/OOD_CIFAR100/ID_cifar100_ece_POVI_kde_f_WGD_coll_cifar100_patches_std_16.csv}\tableIDecePatchsixteen

\pgfplotstableread[col sep=comma]{sections/graphs/OOD_CIFAR100/OOD_cifar100_auroc_POVI_kde_f_WGD_coll_cifar100_patches_std_32.csv}\tableOODPatchthirtytwo
\pgfplotstableread[col sep=comma]{sections/graphs/OOD_CIFAR100/ID_cifar100_accuracy_POVI_kde_f_WGD_coll_cifar100_patches_std_32.csv}\tableIDaccuracyPatchthirtytwo
\pgfplotstableread[col sep=comma]{sections/graphs/OOD_CIFAR100/ID_cifar100_nll_POVI_kde_f_WGD_coll_cifar100_patches_std_32.csv}\tableIDnllPatchthirtytwo
\pgfplotstableread[col sep=comma]{sections/graphs/OOD_CIFAR100/ID_cifar100_ece_POVI_kde_f_WGD_coll_cifar100_patches_std_32.csv}\tableIDecePatchthirtytwo

\pgfplotstableread[col sep=comma]{sections/graphs/OOD_CIFAR100/OOD_cifar100_auroc_POVI_kde_f_WGD_coll_cifar100_patches_std_8.csv}\tableOODPatcheight
\pgfplotstableread[col sep=comma]{sections/graphs/OOD_CIFAR100/ID_cifar100_accuracy_POVI_kde_f_WGD_coll_cifar100_patches_std_8.csv}\tableIDaccuracyPatcheight
\pgfplotstableread[col sep=comma]{sections/graphs/OOD_CIFAR100/ID_cifar100_nll_POVI_kde_f_WGD_coll_cifar100_patches_std_8.csv}\tableIDnllPatcheight
\pgfplotstableread[col sep=comma]{sections/graphs/OOD_CIFAR100/ID_cifar100_ece_POVI_kde_f_WGD_coll_cifar100_patches_std_8.csv}\tableIDecePatcheight

\renewcommand{\fetchAndStore}[4]{%
    \pgfplotstablegetelem{#1}{#2}\of{#3}
    \expandafter\xdef\csname #2Mean#1#4\endcsname{\pgfplotsretval}
    \pgfplotstablegetelem{#1}{#2_std}\of{#3}
    \expandafter\xdef\csname #2Std#1#4\endcsname{\pgfplotsretval}
}

\foreach \method in {softmax,logits,logits_GMM,softmax_SNGP_PE,softmax_en_PE,softmax_en_MI, softmax_laplace_MI} {
    \foreach \intensity in {0,1,2,3,4,5} { 
        \fetchAndStore{\intensity}{\method}{\tableOOD}{OOD}
    }
}

\foreach \method in {softmax,logits,logits_GMM,softmax_SNGP_PE,softmax_en_PE,softmax_en_MI, softmax_laplace_MI} {
    \fetchAndStore{0}{\method}{\tableIDaccuracy}{accuracy}
    \fetchAndStore{0}{\method}{\tableIDnll}{nll}
    \fetchAndStore{0}{\method}{\tableIDece}{ece}
}

\foreach \method in {softmax_POVI_MI} {
    \foreach \intensity in {0,1,2,3,4,5} { 
        \fetchAndStore{\intensity}{\method}{\tableOODSGD}{OODSGD}
        \fetchAndStore{\intensity}{\method}{\tableOODWGD}{OODWGD}
        \fetchAndStore{\intensity}{\method}{\tableOODFtiny}{OODFtiny}
        \fetchAndStore{\intensity}{\method}{\tableOODcifar}{OODcifar}
        \fetchAndStore{\intensity}{\method}{\tableOODFtext}{OODFtext}
        \fetchAndStore{\intensity}{\method}{\tableOODFcifar}{OODFcifar}
        \fetchAndStore{\intensity}{\method}{\tableOODPatchsixteen}{OODPatchsixteen}
        \fetchAndStore{\intensity}{\method}{\tableOODPatchthirtytwo}{OODPatchthirtytwo}
        \fetchAndStore{\intensity}{\method}{\tableOODPatcheight}{OODPatcheight}
    }
}
\foreach \method in {softmax_POVI_MI} {
    \fetchAndStore{0}{\method}{\tableIDaccuracySGD}{accuracySGD}
    \fetchAndStore{0}{\method}{\tableIDnllSGD}{nllSGD}
    \fetchAndStore{0}{\method}{\tableIDeceSGD}{eceSGD}

    \fetchAndStore{0}{\method}{\tableIDaccuracyWGD}{accuracyWGD}
    \fetchAndStore{0}{\method}{\tableIDnllWGD}{nllWGD}
    \fetchAndStore{0}{\method}{\tableIDeceWGD}{eceWGD}
    
    \fetchAndStore{0}{\method}{\tableIDaccuracycifar}{accuracycifar}
    \fetchAndStore{0}{\method}{\tableIDnllcifar}{nllcifar}
    \fetchAndStore{0}{\method}{\tableIDececifar}{ececifar}
    
    \fetchAndStore{0}{\method}{\tableIDaccuracyFtiny}{accuracyFtiny}
    \fetchAndStore{0}{\method}{\tableIDnllFtiny}{nllFtiny}
    \fetchAndStore{0}{\method}{\tableIDeceFtiny}{eceFtiny}
    
    \fetchAndStore{0}{\method}{\tableIDaccuracyFtext}{accuracyFtext}
    \fetchAndStore{0}{\method}{\tableIDnllFtext}{nllFtext}
    \fetchAndStore{0}{\method}{\tableIDeceFtext}{eceFtext}
    
    \fetchAndStore{0}{\method}{\tableIDaccuracyFcifar}{accuracyFcifar}
    \fetchAndStore{0}{\method}{\tableIDnllFcifar}{nllFcifar}
    \fetchAndStore{0}{\method}{\tableIDeceFcifar}{eceFcifar}
    
    \fetchAndStore{0}{\method}{\tableIDaccuracyPatchsixteen}{accuracyPatchsixteen}
    \fetchAndStore{0}{\method}{\tableIDnllPatchsixteen}{nllPatchsixteen}
    \fetchAndStore{0}{\method}{\tableIDecePatchsixteen}{ecePatchsixteen}
    
    \fetchAndStore{0}{\method}{\tableIDaccuracyPatchthirtytwo}{accuracyPatchthirtytwo}
    \fetchAndStore{0}{\method}{\tableIDnllPatchthirtytwo}{nllPatchthirtytwo}
    \fetchAndStore{0}{\method}{\tableIDecePatchthirtytwo}{ecePatchthirtytwo}
    
    \fetchAndStore{0}{\method}{\tableIDaccuracyPatcheight}{accuracyPatcheight}
    \fetchAndStore{0}{\method}{\tableIDnllPatcheight}{nllPatcheight}
    \fetchAndStore{0}{\method}{\tableIDecePatcheight}{ecePatcheight}
}

\begin{table*}[]
\centering
\caption{Comparison of uncertainty estimation for \acrshort{ID} calibration, and \acrshort{OOD} detection on CIFAR100. Mean and standard deviation are computed over 10 runs. Best results are in bold, second best are underlined.}
\label{tab:my-table}
\resizebox{\textwidth}{!}
{
\begin{tabular}{@{}lccccccccc@{}}
\toprule
\multirow{2}{*}{\textbf{Method}} & \multirow{2}{*}{\textsc{Acc.} $\uparrow$ [\%]} & \multirow{2}{*}{\textsc{NLL} $\downarrow$ [\%]} & \multirow{2}{*}{\textsc{ECE} $\downarrow$ [\%]} & \multicolumn{6}{c}{\textsc{OOD} \textsc{Auroc} $\uparrow$ [\%]} \\ \cmidrule(lr){5-9}
 &  &  &  & \textit{TinyIm.} & \textit{Places365} & \textit{Texture} & \textit{SVHN} & \textit{FakeData} &  \\ \midrule
MAP & $\underline{\csname softmaxMean0accuracy\endcsname _{\pm \csname softmaxStd0accuracy\endcsname}}$ & ${\csname softmaxMean0nll\endcsname _{\pm \csname softmaxStd0nll\endcsname}}$ & ${\csname softmaxMean0ece\endcsname _{\pm \csname softmaxStd0ece\endcsname}}$ & $\underline{\csname softmaxMean1OOD\endcsname _{\pm \csname softmaxStd1OOD\endcsname}}$ & ${\csname softmaxMean2OOD\endcsname _{\pm \csname softmaxStd2OOD\endcsname}}$ & ${\csname softmaxMean3OOD\endcsname _{\pm \csname softmaxStd3OOD\endcsname}}$ & ${\csname softmaxMean4OOD\endcsname _{\pm \csname softmaxStd4OOD\endcsname}}$ & ${\csname softmaxMean5OOD\endcsname _{\pm \csname softmaxStd5OOD\endcsname}}$ &  \\ \cmidrule(r){1-9}
DDU & $\underline{\csname softmaxMean0accuracy\endcsname _{\pm \csname softmaxStd0accuracy\endcsname}}$ & ${\csname softmaxMean0nll\endcsname _{\pm \csname softmaxStd0nll\endcsname}}$ & ${\csname softmaxMean0ece\endcsname _{\pm \csname softmaxStd0ece\endcsname}}$ & ${\csname logits_GMMMean1OOD\endcsname _{\pm \csname logits_GMMStd1OOD\endcsname}}$ & ${\csname logits_GMMMean2OOD\endcsname _{\pm \csname logits_GMMStd2OOD\endcsname}}$ & ${\csname logits_GMMMean3OOD\endcsname _{\pm \csname logits_GMMStd3OOD\endcsname}}$ & ${\csname logits_GMMMean4OOD\endcsname _{\pm \csname logits_GMMStd4OOD\endcsname}}$ & $\underline{\csname logits_GMMMean5OOD\endcsname _{\pm \csname logits_GMMStd5OOD\endcsname}}$ &  \\
SNGP & ${\csname softmax_SNGP_PEMean0accuracy\endcsname _{\pm \csname softmax_SNGP_PEStd0accuracy\endcsname}}$ & ${\csname softmax_SNGP_PEMean0nll\endcsname _{\pm \csname softmax_SNGP_PEStd0nll\endcsname}}$ & ${\csname softmax_SNGP_PEMean0ece\endcsname _{\pm \csname softmax_SNGP_PEStd0ece\endcsname}}$ & ${\csname softmax_SNGP_PEMean1OOD\endcsname _{\pm \csname softmax_SNGP_PEStd1OOD\endcsname}}$ & $\mathbf{\csname softmax_SNGP_PEMean2OOD\endcsname _{\pm \csname softmax_SNGP_PEStd2OOD\endcsname}}$ & $\underline{\csname softmax_SNGP_PEMean3OOD\endcsname _{\pm \csname softmax_SNGP_PEStd3OOD\endcsname}}$ & ${\csname softmax_SNGP_PEMean4OOD\endcsname _{\pm \csname softmax_SNGP_PEStd4OOD\endcsname}}$ & ${\csname softmax_SNGP_PEMean5OOD\endcsname _{\pm \csname softmax_SNGP_PEStd5OOD\endcsname}}$ &  \\
LL-Laplace & ${\csname softmax_laplace_MIMean0accuracy\endcsname _{\pm \csname softmax_laplace_MIStd0accuracy\endcsname}}$ & ${\csname softmax_laplace_MIMean0nll\endcsname _{\pm \csname softmax_laplace_MIStd0nll\endcsname}}$ & ${\csname softmax_laplace_MIMean0ece\endcsname _{\pm \csname softmax_laplace_MIStd0ece\endcsname}}$ & $\mathbf{\csname softmax_laplace_MIMean1OOD\endcsname _{\pm \csname softmax_laplace_MIStd1OOD\endcsname}}$ & ${\csname softmax_laplace_MIMean2OOD\endcsname _{\pm \csname softmax_laplace_MIStd2OOD\endcsname}}$ & ${\csname softmax_laplace_MIMean3OOD\endcsname _{\pm \csname softmax_laplace_MIStd3OOD\endcsname}}$ & ${\csname softmax_laplace_MIMean4OOD\endcsname _{\pm \csname softmax_laplace_MIStd4OOD\endcsname}}$ & ${\csname softmax_laplace_MIMean5OOD\endcsname _{\pm \csname softmax_laplace_MIStd5OOD\endcsname}}$ &  \\ 
\midrule
\acrshort{LLPOVI} \emph{(ours)} & ${\csname softmax_POVI_MIMean0accuracySGD\endcsname _{\pm \csname softmax_POVI_MIStd0accuracySGD\endcsname}}$ & $\underline{\csname softmax_POVI_MIMean0nllSGD\endcsname _{\pm \csname softmax_POVI_MIStd0nllSGD\endcsname}}$ & ${\csname softmax_POVI_MIMean0eceSGD\endcsname _{\pm \csname softmax_POVI_MIStd0eceSGD\endcsname}}$ & ${\csname softmax_POVI_MIMean1OODSGD\endcsname _{\pm \csname softmax_POVI_MIStd1OODSGD\endcsname}}$ & ${\csname softmax_POVI_MIMean2OODSGD\endcsname _{\pm \csname softmax_POVI_MIStd2OODSGD\endcsname}}$ & ${\csname softmax_POVI_MIMean3OODSGD\endcsname _{\pm \csname softmax_POVI_MIStd3OODSGD\endcsname}}$ & ${\csname softmax_POVI_MIMean4OODSGD\endcsname _{\pm \csname softmax_POVI_MIStd4OODSGD\endcsname}}$ & ${\csname softmax_POVI_MIMean5OODSGD\endcsname _{\pm \csname softmax_POVI_MIStd5OODSGD\endcsname}}$ &  \\
\acrshort{RLLPOVI} \emph{(ours)} & ${\csname softmax_POVI_MIMean0accuracyWGD\endcsname _{\pm \csname softmax_POVI_MIStd0accuracyWGD\endcsname}}$ & ${\csname softmax_POVI_MIMean0nllWGD\endcsname _{\pm \csname softmax_POVI_MIStd0nllWGD\endcsname}}$ & $\underline{\csname softmax_POVI_MIMean0eceWGD\endcsname _{\pm \csname softmax_POVI_MIStd0eceWGD\endcsname}}$ & ${\csname softmax_POVI_MIMean1OODWGD\endcsname _{\pm \csname softmax_POVI_MIStd1OODWGD\endcsname}}$ & ${\csname softmax_POVI_MIMean2OODWGD\endcsname _{\pm \csname softmax_POVI_MIStd2OODWGD\endcsname}}$ & ${\csname softmax_POVI_MIMean3OODWGD\endcsname _{\pm \csname softmax_POVI_MIStd3OODWGD\endcsname}}$ & ${\csname softmax_POVI_MIMean4OODWGD\endcsname _{\pm \csname softmax_POVI_MIStd4OODWGD\endcsname}}$ & ${\csname softmax_POVI_MIMean5OODWGD\endcsname _{\pm \csname softmax_POVI_MIStd5OODWGD\endcsname}}$ &  \\
\acrshort{fLLPOVI} \emph{(ours)} &  &  &  &  &  &  &  &  &  \\
\textit{~~~+ Cifar100} & ${\csname softmax_POVI_MIMean0accuracyFcifar\endcsname _{\pm \csname softmax_POVI_MIStd0accuracyFcifar\endcsname}}$ & ${\csname softmax_POVI_MIMean0nllFcifar\endcsname _{\pm \csname softmax_POVI_MIStd0nllFcifar\endcsname}}$ & ${\csname softmax_POVI_MIMean0eceFcifar\endcsname _{\pm \csname softmax_POVI_MIStd0eceFcifar\endcsname}}$ & ${\csname softmax_POVI_MIMean1OODFcifar\endcsname _{\pm \csname softmax_POVI_MIStd1OODFcifar\endcsname}}$ & ${\csname softmax_POVI_MIMean2OODFcifar\endcsname _{\pm \csname softmax_POVI_MIStd2OODFcifar\endcsname}}$ & ${\csname softmax_POVI_MIMean3OODFcifar\endcsname _{\pm \csname softmax_POVI_MIStd3OODFcifar\endcsname}}$ & ${\csname softmax_POVI_MIMean4OODFcifar\endcsname _{\pm \csname softmax_POVI_MIStd4OODFcifar\endcsname}}$ & ${\csname softmax_POVI_MIMean5OODFcifar\endcsname _{\pm \csname softmax_POVI_MIStd5OODFcifar\endcsname}}$ &  \\
\textit{~~~+ TinyImagenet} & ${\csname softmax_POVI_MIMean0accuracyFtiny\endcsname _{\pm \csname softmax_POVI_MIStd0accuracyFtiny\endcsname}}$ & ${\csname softmax_POVI_MIMean0nllFtiny\endcsname _{\pm \csname softmax_POVI_MIStd0nllFtiny\endcsname}}$ & ${\csname softmax_POVI_MIMean0eceFtiny\endcsname _{\pm \csname softmax_POVI_MIStd0eceFtiny\endcsname}}$ & $\underline{\csname softmax_POVI_MIMean1OODFtiny\endcsname _{\pm \csname softmax_POVI_MIStd1OODFtiny\endcsname}}$ & $\underline{\csname softmax_POVI_MIMean2OODFtiny\endcsname _{\pm \csname softmax_POVI_MIStd2OODFtiny\endcsname}}$ & ${\csname softmax_POVI_MIMean3OODFtiny\endcsname _{\pm \csname softmax_POVI_MIStd3OODFtiny\endcsname}}$ & $\underline{\csname softmax_POVI_MIMean4OODFtiny\endcsname _{\pm \csname softmax_POVI_MIStd4OODFtiny\endcsname}}$ & $\underline{\csname softmax_POVI_MIMean5OODFtiny\endcsname _{\pm \csname softmax_POVI_MIStd5OODFtiny\endcsname}}$ &  \\
\textit{~~~+ Texture} & ${\csname softmax_POVI_MIMean0accuracyFtext\endcsname _{\pm \csname softmax_POVI_MIStd0accuracyFtext\endcsname}}$ & ${\csname softmax_POVI_MIMean0nllFtext\endcsname _{\pm \csname softmax_POVI_MIStd0nllFtext\endcsname}}$ & ${\csname softmax_POVI_MIMean0eceFtext\endcsname _{\pm \csname softmax_POVI_MIStd0eceFtext\endcsname}}$ & ${\csname softmax_POVI_MIMean1OODFtext\endcsname _{\pm \csname softmax_POVI_MIStd1OODFtext\endcsname}}$ & ${\csname softmax_POVI_MIMean2OODFtext\endcsname _{\pm \csname softmax_POVI_MIStd2OODFtext\endcsname}}$ & $\mathbf{\csname softmax_POVI_MIMean3OODFtext\endcsname _{\pm \csname softmax_POVI_MIStd3OODFtext\endcsname}}$ & $\mathbf{\csname softmax_POVI_MIMean4OODFtext\endcsname _{\pm \csname softmax_POVI_MIStd4OODFtext\endcsname}}$ & ${\csname softmax_POVI_MIMean5OODFtext\endcsname _{\pm \csname softmax_POVI_MIStd5OODFtext\endcsname}}$ &  \\
\textit{~~~+ Patches-32} & ${\csname softmax_POVI_MIMean0accuracyPatchthirtytwo\endcsname _{\pm \csname softmax_POVI_MIStd0accuracyPatchthirtytwo\endcsname}}$ & $\underline{\csname softmax_POVI_MIMean0nllPatchthirtytwo\endcsname _{\pm \csname softmax_POVI_MIStd0nllPatchthirtytwo\endcsname}}$ & ${\csname softmax_POVI_MIMean0ecePatchthirtytwo\endcsname _{\pm \csname softmax_POVI_MIStd0ecePatchthirtytwo\endcsname}}$ & ${\csname softmax_POVI_MIMean1OODPatchthirtytwo\endcsname _{\pm \csname softmax_POVI_MIStd1OODPatchthirtytwo\endcsname}}$ & ${\csname softmax_POVI_MIMean2OODPatchthirtytwo\endcsname _{\pm \csname softmax_POVI_MIStd2OODPatchthirtytwo\endcsname}}$ & ${\csname softmax_POVI_MIMean3OODPatchthirtytwo\endcsname _{\pm \csname softmax_POVI_MIStd3OODPatchthirtytwo\endcsname}}$ & ${\csname softmax_POVI_MIMean4OODPatchthirtytwo\endcsname _{\pm \csname softmax_POVI_MIStd4OODPatchthirtytwo\endcsname}}$ & $\mathbf{\csname softmax_POVI_MIMean5OODPatchthirtytwo\endcsname _{\pm \csname softmax_POVI_MIStd5OODPatchthirtytwo\endcsname}}$ &  \\
\textit{~~~+ Patches-16} & ${\csname softmax_POVI_MIMean0accuracyPatchsixteen\endcsname _{\pm \csname softmax_POVI_MIStd0accuracyPatchsixteen\endcsname}}$ & ${\csname softmax_POVI_MIMean0nllPatchsixteen\endcsname _{\pm \csname softmax_POVI_MIStd0nllPatchsixteen\endcsname}}$ & ${\csname softmax_POVI_MIMean0ecePatchsixteen\endcsname _{\pm \csname softmax_POVI_MIStd0ecePatchsixteen\endcsname}}$ & ${\csname softmax_POVI_MIMean1OODPatchsixteen\endcsname _{\pm \csname softmax_POVI_MIStd1OODPatchsixteen\endcsname}}$ & ${\csname softmax_POVI_MIMean2OODPatchsixteen\endcsname _{\pm \csname softmax_POVI_MIStd2OODPatchsixteen\endcsname}}$ & ${\csname softmax_POVI_MIMean3OODPatchsixteen\endcsname _{\pm \csname softmax_POVI_MIStd3OODPatchsixteen\endcsname}}$ & ${\csname softmax_POVI_MIMean4OODPatchsixteen\endcsname _{\pm \csname softmax_POVI_MIStd4OODPatchsixteen\endcsname}}$ & $\mathbf{\csname softmax_POVI_MIMean5OODPatchsixteen\endcsname _{\pm \csname softmax_POVI_MIStd5OODPatchsixteen\endcsname}}$ &  \\
\midrule
DE-5  & $\mathbf{\csname softmax_en_MIMean0accuracy\endcsname _{\pm \csname softmax_en_MIStd0accuracy\endcsname}}$ & $\mathbf{\csname softmax_en_MIMean0nll\endcsname _{\pm \csname softmax_en_MIStd0nll\endcsname}}$ & $\mathbf{\csname softmax_en_MIMean0ece\endcsname _{\pm \csname softmax_en_MIStd0ece\endcsname}}$ & ${\csname softmax_en_MIMean1OOD\endcsname _{\pm \csname softmax_en_MIStd1OOD\endcsname}}$ & ${\csname softmax_en_MIMean2OOD\endcsname _{\pm \csname softmax_en_MIStd2OOD\endcsname}}$ & ${\csname softmax_en_MIMean3OOD\endcsname _{\pm \csname softmax_en_MIStd3OOD\endcsname}}$ & ${\csname softmax_en_MIMean4OOD\endcsname _{\pm \csname softmax_en_MIStd4OOD\endcsname}}$ & ${\csname softmax_en_MIMean5OOD\endcsname _{\pm \csname softmax_en_MIStd5OOD\endcsname}}$ &  \\ \bottomrule
\end{tabular}%
}
\label{tab:OOD:cifar100}
\end{table*}
\subsection{Semantic shift detection}

To further verify the epistemic uncertainty estimates of our method, we perform OOD detection on larger image classification tasks. The results are summarized in Tables \ref{tab:OOD:cifar10} and \ref{tab:OOD:cifar100}. We train a Wide-Resnet-28-10 for CIFAR10 and CIFAR100. Again, our \acrshort{LLPOVI} head consists of 10 particles with linear layers only. \\
In terms of accuracy and \gls{ID} calibration, \glspl{DE} perform the best, followed by our \gls{fLLPOVI}. As we only retrain the last layer of the base network, we do not expect to improve the prediction accuracy. We emphasize that our aim is to introduce computationally cheap uncertainty estimates that are informative about erroneous predictions and OOD data. For both CIFAR10 and CIFAR100, retraining the last layer leads to improvements over the base network (MAP) for all examined uncertainty scores. While last-layer retraining does not lead to improved accuracy, NLL and ECE are consistently improved, which indicates better calibration of the uncertainty estimates. Also, function space diversity on augmented training data can lead to further improvements for \gls{ID} uncertainty estimates. 

Additionally, we note improvements in \gls{OOD} detection. Using informative repulsion samples for the function space loss results in better performance than unregularized \acrshort{LLPOVI} retraining. Particularly for CIFAR100, the epistemic uncertainty of \gls{fLLPOVI} with diverse repulsion samples (e.g., TinyImagenet) surpasses full \glspl{DE} on most \gls{OOD} datasets. While DDU effectively detects far \gls{OOD} data (Texture, FakeData), it does not outperform the base network (MAP) on \gls{OOD} samples close to the training set (CIFAR100, TinyImagenet). As these images may have similar features to the training set, considering feature density alone may not be sufficient. In such cases, leveraging function space diversity with informative repulsion samples can overcome the limitations of spurious shared features and enhance improve uncertainty estimates. For instance, diversity on CIFAR100 or TinyImagenet generalizes well to unseen \gls{OOD} samples. On the other hand, diversity on the training data itself leads to worse \gls{ID} calibration and \gls{OOD} detection.

In Appendix B.4, we show that spectral normalization has minimal impact when the base network includes residual connections. In Appendix B.5, we evaluate our method across different base networks with varying feature space sizes. Additionally, we test various repulsion samples and the generalization to detect new \gls{OOD} distributions.

\tikzexternaldisable
\begin{figure*}[t!]
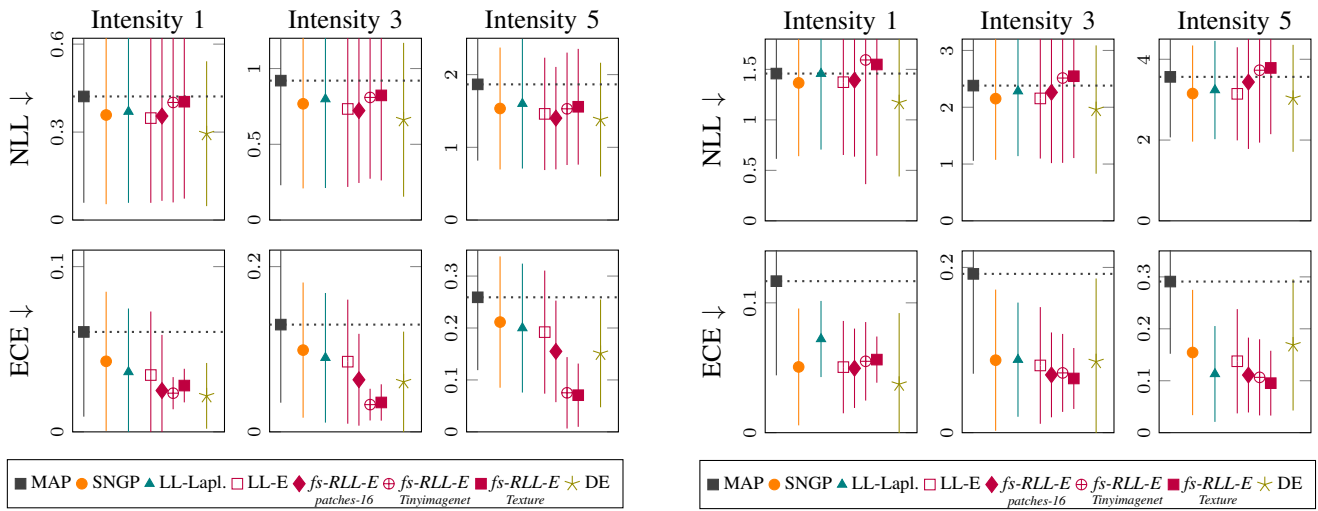

  \begin{minipage}{0.45\textwidth}
   \input{sections/graphs/barplot_cifar10_corrupted}
  \end{minipage}
  \hspace*{.8cm}
 \begin{minipage}{0.45\textwidth}
    \vspace*{-.4cm}
    \input{sections/graphs/barplot_cifar100_corrupted}
  \end{minipage}
    \caption{NLL and uncertainty calibration of the different methods on CIFAR10-C (left) and CIFAR100-C (right), for different levels of corruption intensity, averaged over all corruption types. By retraining the last linear layer only, our method \acrshort{fLLPOVI} improves NLL, and achieves similar ECE scores as \acrshortpl{DE}. }
  \label{fig:corruption:cifar}
\end{figure*}

\tikzexternalenable
\subsection{Covariate shift calibration}

Finally, we analyze the behavior of our model when presented with corrupted data (CIFAR10-C and CIFAR100-C \cite{hendrycks2018benchmarking}) for the same base network, Wide-Resnet-28-10. These datasets contain 19 types of corruptions, each with 5 different severity levels. 
Figure \ref{fig:corruption:cifar} shows the NLL and ECE results averaged over all corruption types.
Retraining the last layer without any regularization (\acrshort{LLPOVI}) improves the uncertainty calibration of the single network (MAP) for both datasets. Function space repulsion further improves calibration on CIFAR10-C, outperforming LL-Laplace and SNGP, and achieving competitive uncertainty calibration to \glspl{DE}. 
If the final layer of the network is sensitive to features that are not relevant to the target class, corrupted data might activate these irrelevant features, resulting in overconfident and inaccurate predictions. Enforcing predictive diversity on specified repulsion samples can help to down-weight the influence of non-robust features (through the repulsion term) and focus on robust features for prediction, thereby improving the model's calibration and performance on corrupted data. Interestingly, for CIFAR100-C, the benefits of performing inference in function space are not as evident as for CIFAR10-C. Here, random initialization and retraining of the last layer (\acrshort{LLPOVI}) achieves comparable results to the other uncertainty methods at minimal parameter and computational cost. Results for the individual corruption types are summarized in Appendix B.1.

\section{Conclusion}
We have shown that particle optimization in function space is not limited to \gls{DE} architectures. A significant number of parameters can be saved by exploring different network architectures to parameterize the function space. We proposed a hybrid approach using a multi-headed network. The shared base network acts as a feature extractor for the repulsive ensemble head. This offers a principled way to provide already trained networks with retrospective uncertainty estimates, and to incorporate prior functional knowledge into the training procedure.
Additionally, we highlighted the inherent limitations of enforcing diversity on training data alone. By utilizing augmented training data, or unlabeled \gls{OOD} data, we achieve significant improvements on \gls{OOD} detection without harming classification accuracy.    
We empirically demonstrate that our method successfully disentangles aleatoric and epistemic uncertainty, improves \gls{OOD} detection, provides calibrated uncertainty estimates under distribution shifts, and performs well in active learning. At the same time, we significantly reduce the computational and memory requirements compared to \glspl{DE}.

\subsection*{Future work}
The function space formulation of the inference problem requires a selection of repulsion samples. If those repulsion samples do not cover the domain of interest during deployment of the model, the function space repulsion term will not lead to improvements. 
As future work, it is of interest to give more rigorous statements about the selection of repulsion samples and the implications for uncertainty estimates. We plan to further explore data augmentation schemes for the generation of repulsion samples for different tasks and their respective limitations.

\section*{Acknowledgments}
The authors gratefully acknowledge the financial support under the scope of the COMET program within
the K2 Center “Integrated Computational Material, Process and Product Engineering (IC-MPPE)” (Project No
886385). This program is supported by the Austrian Federal Ministries for Climate Action, Environment,
Energy, Mobility, Innovation and Technology (BMK) and for Labour and Economy (BMAW), represented by
the Austrian Research Promotion Agency (FFG), and the federal states of Styria, Upper Austria and Tyrol. 

The financial support by the Austrian Federal Ministry of Labour and Economy, the National Foundation for Research, Technology and Development and the Christian Doppler Research Association is gratefully acknowledged. 


\bibliographystyle{IEEEtran}

\vspace{-33pt}

\begin{IEEEbiography}[{\includegraphics[width=1in,height=1.25in,clip,keepaspectratio]{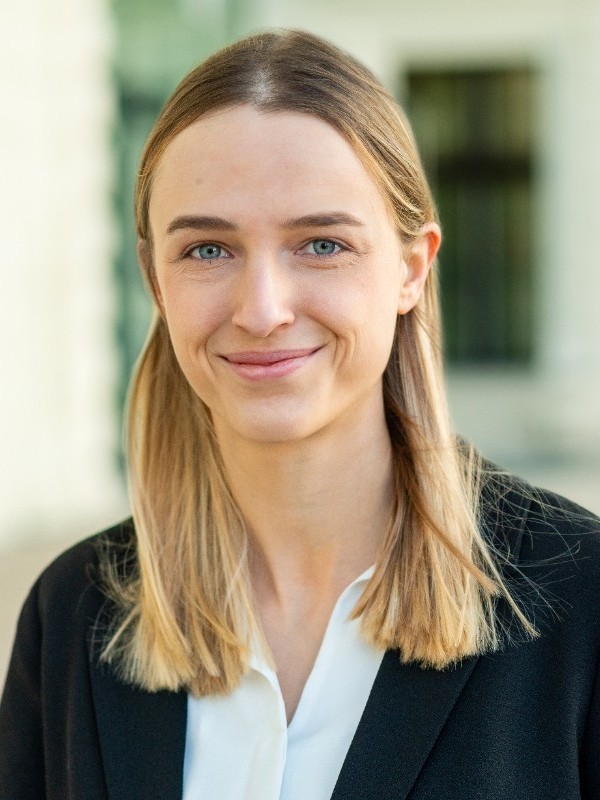}}]{Sophie Steger}
received her BSc and MSc (Dipl. Ing.) degree in Electrical Engineering at Graz University of Technology in in 2020 and 2022, respectively. She is currently working towards the PhD degree with the Laboratory of Signal Processing and Speech Communication, at Graz University of Technology. Her research interests include Bayesian deep learning, probabilistic machine learning, and predictive uncertainty quantification.
\end{IEEEbiography}


\begin{IEEEbiography}[{\includegraphics[width=1in,height=1.25in,clip,keepaspectratio]{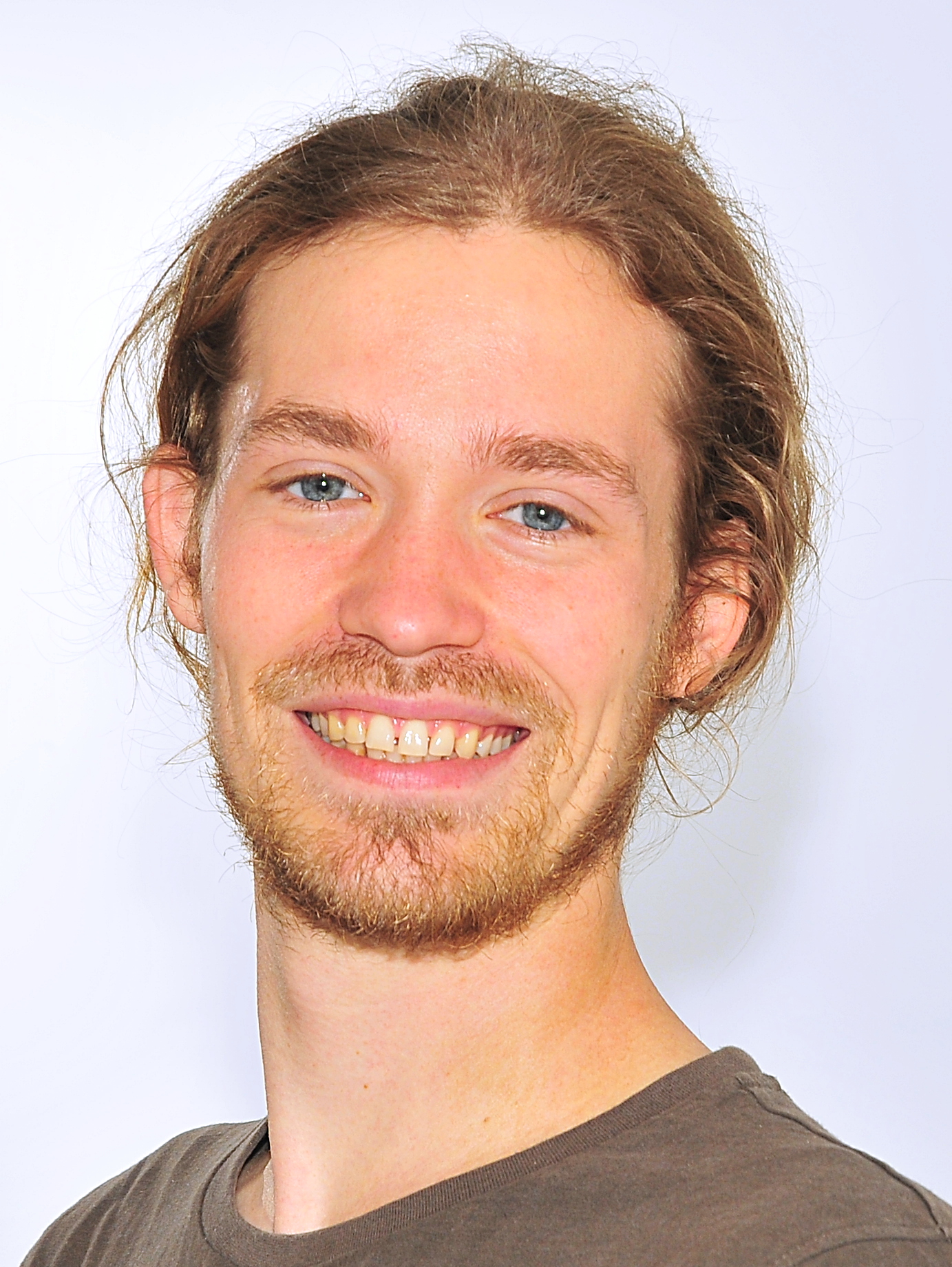}}]{Christian Knoll}
received the BSc and MSc degrees in information and computer engineering, in 2011 and 2014, respectively, and the PhD degree from the Graz University of Technology, in 2019. In 2022 he was awarded the Josef Krainer Förderungspreis. He is a co-founder of Levata, an R\&D consulting company. His research interests include graphical models, statistical signal processing, message passing methods, hybrid-, and  probabilistic machine learning.
\end{IEEEbiography}

\begin{IEEEbiography}[{\includegraphics[width=1in,height=1.25in,clip,keepaspectratio]{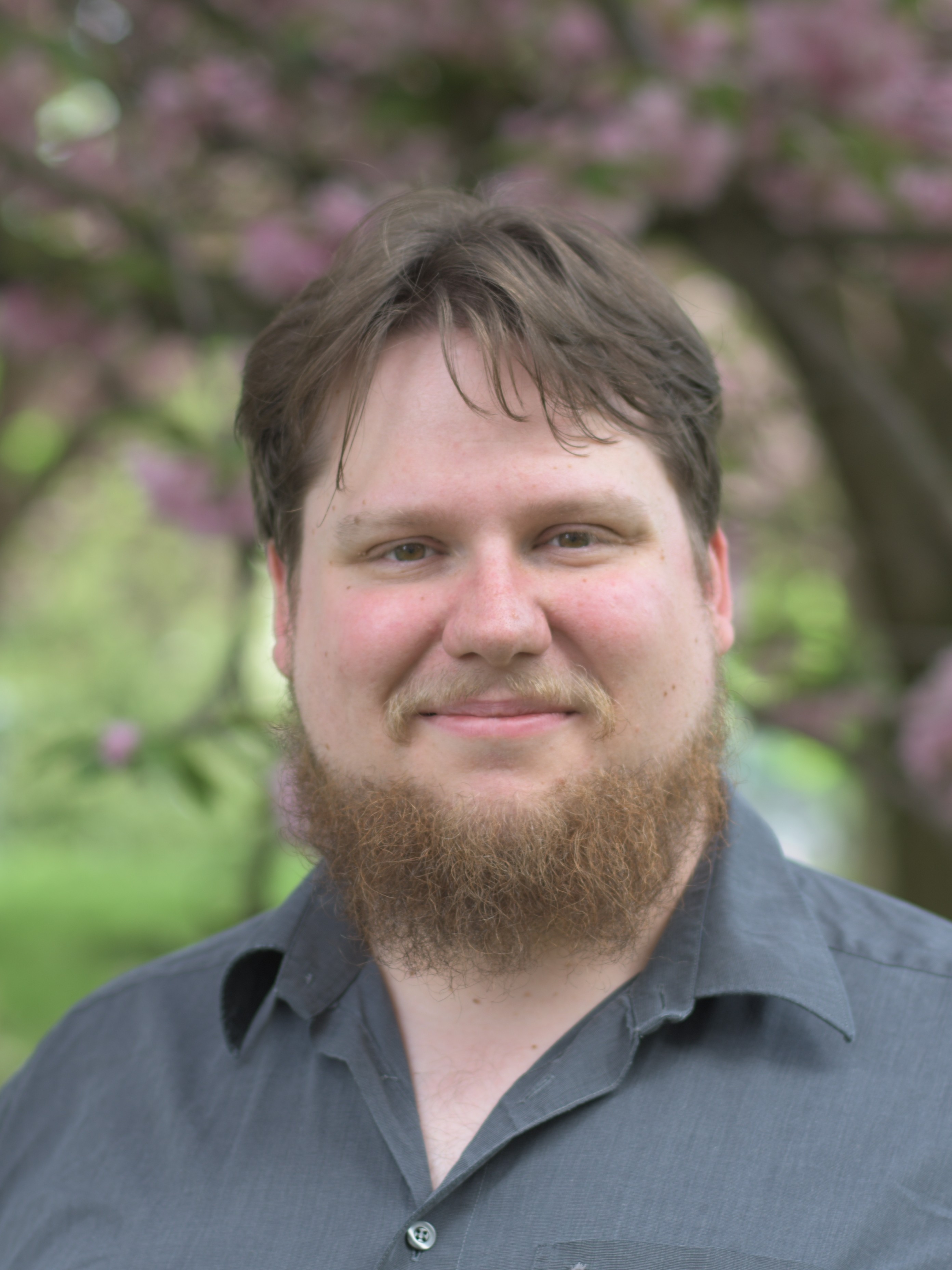}}]{
Bernhard Klein}
is a fifth-year PhD student at the Computing Systems Group lead by Prof. Dr. Holger Fröning at Heidelberg University. He completed his Master of Science at the Institute for Environmental Physics in 2018. His research interests include code generation, machine learning, probabilistic, approximate and noisy computing and all means of closing the gap between machine learning and embedded devices. Current research projects include automatic compression of neural networks for embedded systems, finding robust neural architectures and methods to increase robustness against analog noisy hardware, and efficient hardware-aware, inference of Bayesian neural networks on resource-constrained embedded systems. 
\end{IEEEbiography}

\begin{IEEEbiography}[{\includegraphics[width=1in,height=1.25in,clip,keepaspectratio]{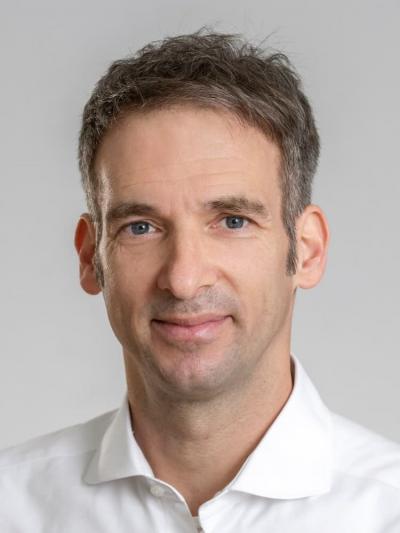}}]{
Holger Fr\"oning}
is a full professor at Heidelberg University, where he leads the Computing Systems Group within the Institute of Computer Engineering. From 2011 to 2018, he was associate professor at the same university. In 2016, he was with NVIDIA Research (Santa Clara, CA, US) as visiting scientist, sponsored by Bill Dally. Early 2015 he was visiting professor at the Graz University of Technology (Austria), sponsored by Gernot Kubin. From 2008 to 2011 he reported to Jose Duato from the Technical University of Valencia (Spain). He has received his PhD and MSc degrees 2007 respectively 2001 from University of Mannheim, Germany. His research centers on embedded machine learning and high-performance computing, and the intersection with hardware architectures.
\end{IEEEbiography}

\begin{IEEEbiography}[{\includegraphics[width=1in,height=1.25in,clip,keepaspectratio]{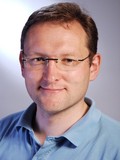}}]{Franz Pernkopf}
received his MSc (Dipl. Ing.) degree in Electrical Engineering at Graz University of Technology, and a PhD degree from the University of Leoben. In 2002 he was awarded the Erwin Schr\"odinger Fellowship. He was a Research Associate at the University of Washington, Seattle, from 2004 to 2006. From 2010-2019 he was an Associate Professor at the Laboratory of Signal Processing and Speech Communication; since 2019, he is a Professor for Intelligent Systems, at Graz University of Technology. His research is focused on pattern recognition, machine learning, and computational data analytics with applications in various fields ranging from signal processing to medical data analysis and industrial applications.qu
\end{IEEEbiography}

\onecolumn

\makeatletter
\renewcommand\appendix{\par
  \setcounter{section}{0}%
  \setcounter{subsection}{0}%
  \gdef\thesection{\@Alph\c@section}}
\makeatother
\appendix

\vspace{-.5cm}

\renewcommand{\thesection}{\textsc{Appendix} \Alph{section}}
\renewcommand{\thesubsection}{\thesection.\arabic{subsection}}

\section{Training details} \label{app:training_details}
For particle-based inference in function space \cite{wang2019function,dangelo2021repulsive}, we relied on the implementation available at \url{https://github.com/ratschlab/repulsive_ensembles}, and for DDU \cite{mukhoti2023deep} at \url{https://github.com/omegafragger/DDU}.
Table \ref{tab:implementation} summarizes relevant hyperparameters for training the base networks and our \gls{RLLPOVI}.

\begin{table}[h!]
\centering
\caption{Implementation details and hyperparameter for the different experiments.}
\label{tab:my-table}
\resizebox{1\columnwidth}{!}{%
\begin{tabular}{@{}cccc@{}}
\toprule
\textsc{Task} & \textsc{Architecture} & \textsc{Hyperparameter} & \textsc{Value} \\ \midrule
\multirow{8}{*}{\begin{tabular}[c]{@{}c@{}}\textsc{Image classification} \\  \textsc{base network}\end{tabular}} & \multirow{8}{*}{\begin{tabular}[c]{@{}c@{}}\textsc{Resnet-18} \\  \textsc{Wide-Resnet-28-10} \end{tabular}} & \multirow{2}{*}{\textsc{Epochs}} & 50 (DirtyMNIST) \\
 &  &  & 300 (CIFAR10/100) \\ \cmidrule(l){3-4} 
 &  & \textsc{Optimizer} & SGD \\ \cmidrule(l){3-4} 
 &  & \multirow{3}{*}{\textsc{Learning rate}} & 0.1 \\
 &  &  & 0.01 -- epoch 25 (DirtyMNIST),  epoch 150 (CIFAR10/100) \\
 &  &  & 0.001 -- epoch 40 (DirtyMNIST),  epoch 250 (CIFAR10/100) \\ \cmidrule(l){3-4} 
 &  & \textsc{Momentum} & 0.9 \\ \midrule  
\multirow{3}{*}{\begin{tabular}[c]{@{}c@{}}\textsc{Active learning} \\ \textsc{base network}\end{tabular}} & \multirow{3}{*}{\textsc{Resnet-18}} & \textsc{Epochs} & 20 \\ \cmidrule(l){3-4} 
 &  & \textsc{Optimizer} & Adam \\ \cmidrule(l){3-4} 
 &  & \textsc{Learning rate} & 0.001 \\ \midrule
\multirow{5}{*}{\textsc{Last-Layer-Ensemble}} & \multirow{5}{*}{\textsc{Fully connected}} & \textsc{Epochs} & 30 \\ \cmidrule(l){3-4} 
 &  & \# \textsc{hidden layer} & 0 \\
 &  & \# \textsc{neurons per layer} & 10 \\ \cmidrule(l){3-4} 
 &  & \textsc{Learning rate} & 0.0001 \\ \cmidrule(l){3-4} 
 &  & \# \textsc{batch size training data} & 128 \\ 
 &  & \# \textsc{batch size repulsion samples} & 128 \\ 
 \bottomrule
\end{tabular}%
} 
\label{tab:implementation}
\end{table}

\paragraph*{Spectral normalization}
We follow the implementation of \cite{mukhoti2023deep} and utilize base networks with residual connections and spectral normalization. As proposed in SNGP and DDU \cite{liu2023simple,mukhoti2023deep}, spectral normalization is used to bound the Lipschitz constant of the network. Online spectral normalization with a one step power iteration is applied to convolutional weights, and exact spectral normalization is applied to 1x1 convolutional layers \cite{mukhoti2023deep}. \\

\paragraph*{Computational overhead}
In all experiments, we use a pretrained base network as the feature extractor for the uncertainty estimation. 
For the image classification experiments, we do not modify the base network and add an ensemble head consisting of linear layers only. Thus, the number of trainable parameters of our method is determined by the dimension of the feature space of the base network $d$, the number of classes $K$, and the number of particles $n$, i.e., $(d \times K + K)\times n$. The feature space dimension for various base networks is shown in Table \ref{tab:feature_dim}.
\setlength{\tabcolsep}{7pt} 
\begin{table}[H]
\centering
\caption{Feature space dimension of different base network architectures.}
\label{tab:feature_dim}
\begin{tabular}{ll}
\hline
\textsc{LeNet}   & $d=84$  \\
\textsc{VGG-16}    & $d=512$  \\
\textsc{Resnet-18} & $d=512$ \\
\textsc{WideResNet-28-10} & $d=640$  \\
\textsc{Resnet-50}        & $d=2048$ \\
\textsc{Resnet-101}       & $d=2048$ \\
\hline
\end{tabular}
\end{table}
\pagebreak

\renewcommand{\arraystretch}{1.3}

\newpage

\section{Additional experimental results}

\subsection{Covariate shift calibration} \label{sec:individual_corruptions}
\begin{figure}[b!]
    \centering
    \includegraphics[width=1\linewidth]{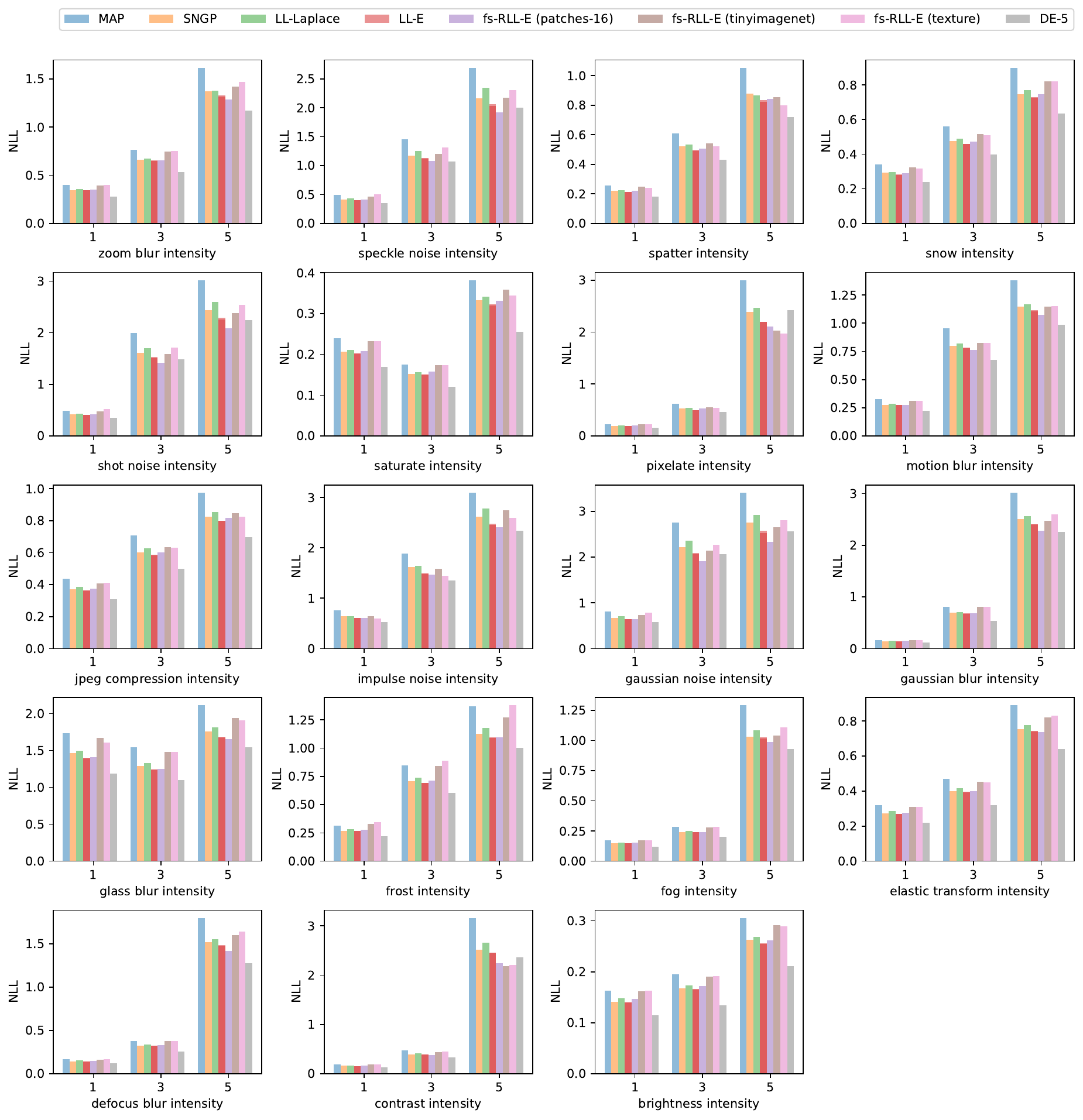}
    \caption{Negative log-likelihood (NLL) on CIFAR10-C.}
    \label{fig:cifar10c-nll-individual}
\end{figure}

\begin{figure}
    \centering
    \includegraphics[width=1\linewidth]{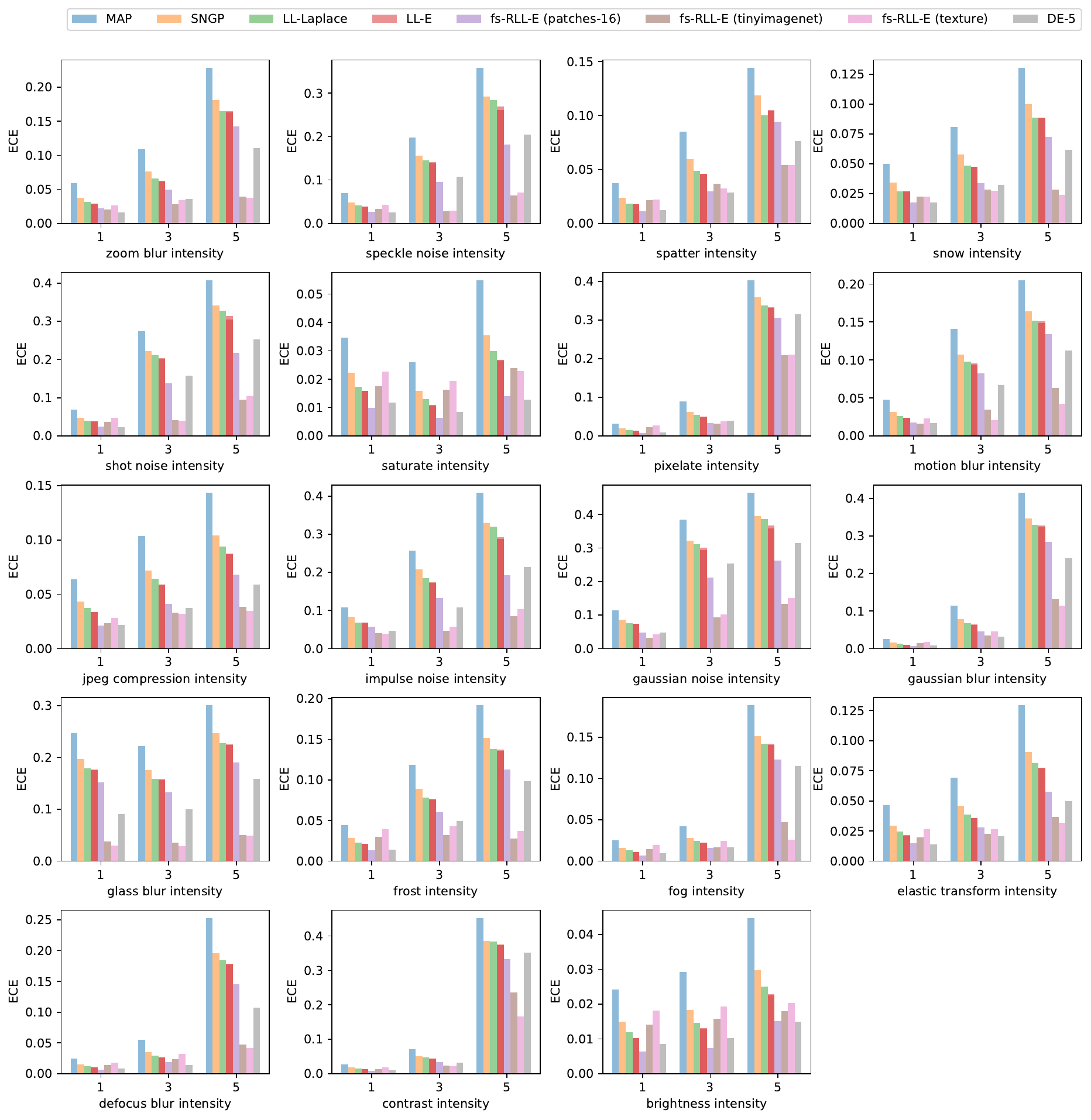}
    \caption{Expected calibration error (ECE) on CIFAR10-C.}
    \label{fig:cifar10c-ece-individual}
\end{figure}

\begin{figure}[b!]
    \centering
    \includegraphics[width=1\linewidth]{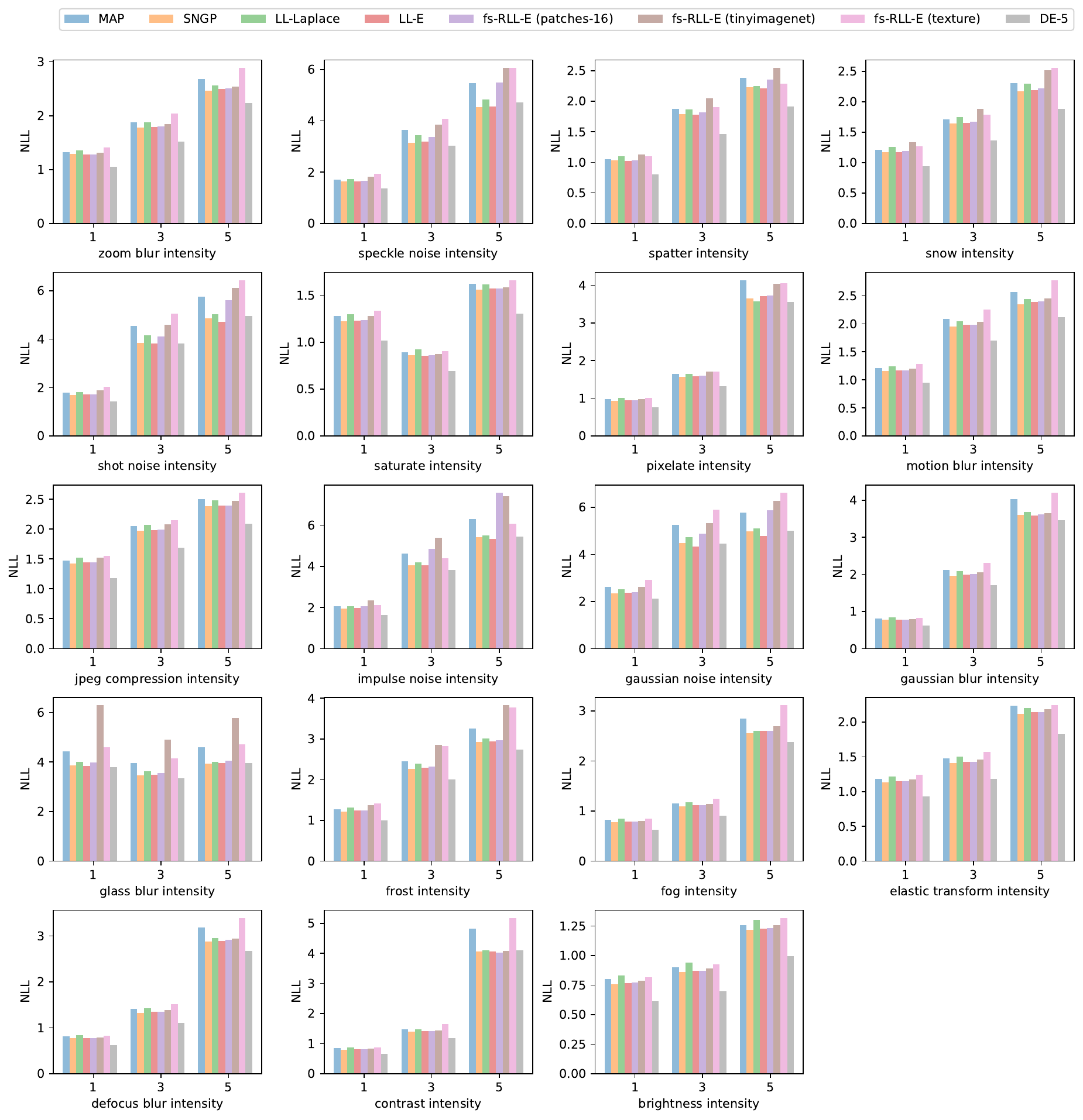}
    \caption{Negative log-likelihood (NLL) on CIFAR100-C.}
    \label{fig:cifar100c-nll-individual}
\end{figure}

\begin{figure}[b!]
    \centering
    \includegraphics[width=1\linewidth]{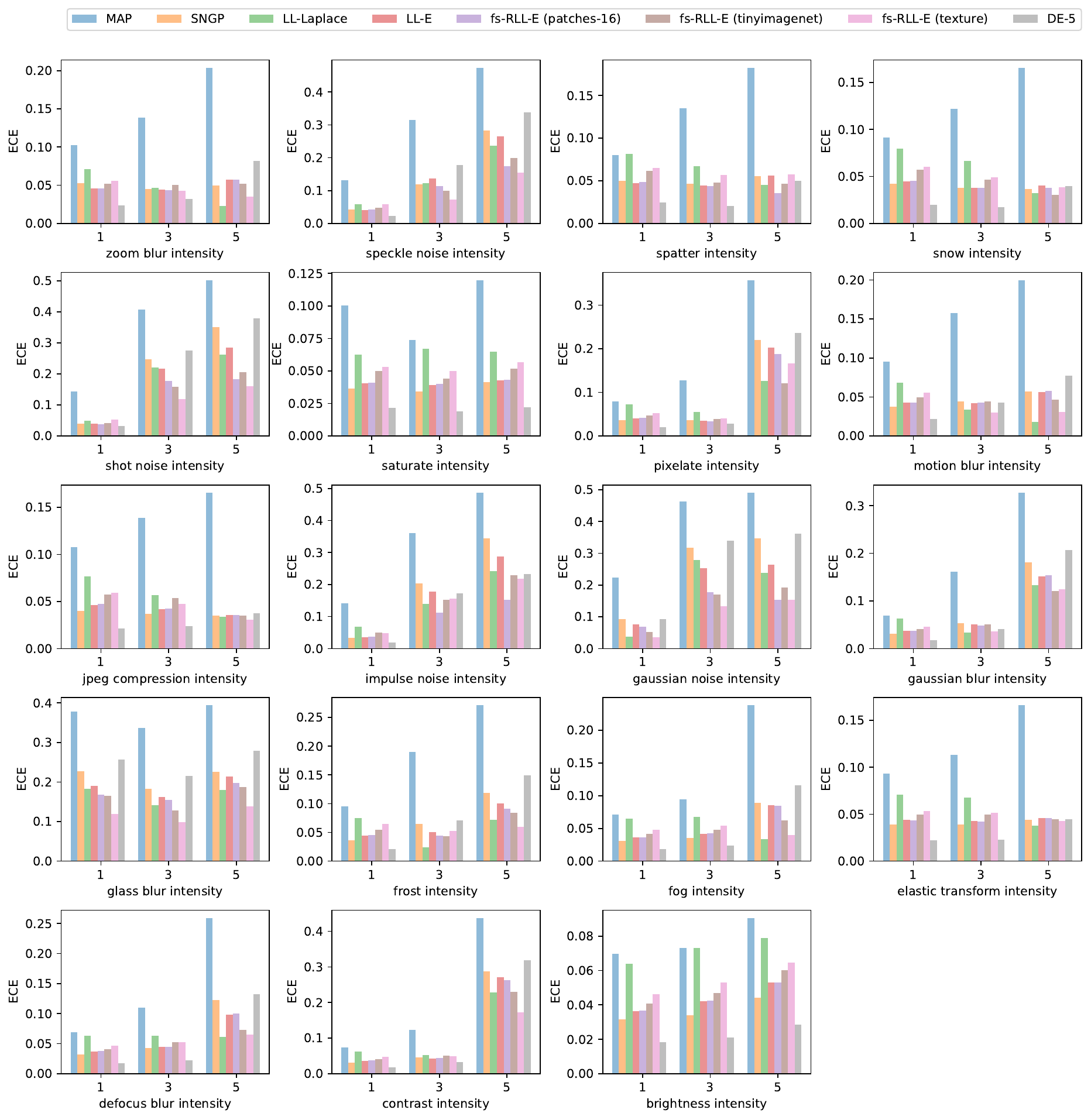}
    \caption{Expected calibration error (ECE) on CIFAR100-C.}
    \label{fig:cifar100c-ece-individual}
\end{figure}

We evaluated the uncertainty calibration of our method on the CIFAR-10-C and CIFAR-100-C datasets \cite{hendrycks2018benchmarking}. 
Figure \ref{fig:cifar10c-nll-individual}, \ref{fig:cifar10c-ece-individual}, \ref{fig:cifar100c-nll-individual}, and \ref{fig:cifar100c-ece-individual} show the results for the individual corruption types and severity levels 1, 3, and 5. 
Last-layer retraining improves calibration of the uncertainty estimates across almost all corruption types. 

\clearpage
\clearpage
\clearpage
\clearpage
\pagebreak
\subsection{Inductive biases of the base network}\label{sec:inductive_biases}
Deterministic uncertainty methods such as DDU \cite{mukhoti2023deep} and SNGP \cite{liu2023simple} emphasized the importance of the inductive biases of the base network. The feature extractor needs to be sufficiently sensitive to distances in the input space to provide meaningful uncertainty estimates. We evaluate the performance of our method with different base network architectures. We repeat the uncertainty decomposition experiments on the DirtyMNIST dataset in Section VIII-B with a LeNet-5, VGG-16, and Resnet-18 architecture. Table \ref{tab:mnist_inductive_biases} summarizes the results, where we report the ID accuracy and OOD detection performance. Similar to ablation studies in DDU \cite{mukhoti2023deep}, we observe that the base network architecture plays an important role in the disentanglement of aleatoric and epistemic uncertainty. However, \acrshort{fLLPOVI} provides more robust results across different network architectures compared to DDU and SNGP.
\def\arraystretch{1.5}%

\setlength{\tabcolsep}{10pt} 
\begin{table}[H]
\centering
\caption{Classification and uncertainty calibration on DirtyMNIST for different base network architectures. Results are averaged over 5 runs.}
\label{tab:mnist_inductive_biases}
\begin{tabular}{llccccccc}
\hline
& \textbf{Base network} & MAP & DDU & SNGP & LL-Laplace & DE-5 & \acrshort{LLPOVI} \textit{(ours)} &  $\underset{+\textit{eMNIST}}{\text{\acrshort{fLLPOVI}}}$ \textit{(ours)}\\ \hline
\multirow{4}{*}{\rotatebox[origin=c]{90}{Accuracy $\uparrow$}} & LeNet     &      77.94 & 77.94 & \underline{82.24} & 77.98 & \textbf{82.48} & 81.13 & 81.09 \\
                                                               & VGG-11    &      80.12 & 80.12 & 83.31 & 80.05 & 83.23 & \textbf{83.49} & \underline{83.45} \\
                                                               & VGG-16    &      79.17 & 79.17 & \textbf{83.17} & 79.10 & \underline{83.12} & 83.07 & 83.04 \\
                                                               & Resnet-18 &      80.76 & 80.76 & \textbf{83.64} & 80.73 & 83.23 & \underline{83.54} & \underline{83.54} \\
                                                               \hline
\multirow{4}{*}{\rotatebox[origin=c]{90}{ECE $\downarrow$}} 
                                                               & LeNet        &    3.53 & 3.53 & \underline{1.63} & 3.42 & 9.49 & \textbf{1.47} & 1.82 \\
                                                               & VGG-11       &    2.15 & 2.15 & 3.98 & 2.03 & 4.85 & \textbf{0.95} & \underline{1.13} \\
                                                               & VGG-16       &    3.12 & 3.12 & 3.65 & 2.75 & 5.05 & \textbf{1.07} & \underline{1.26} \\
                                                               & Resnet-18    &    3.28 & 3.28 & 3.04 & 2.96 & 5.50 & \textbf{0.93} & \underline{1.14} \\
                                                               \hline
\multirow{4}{*}{\rotatebox[origin=c]{90}{$\underset{\text{MNIST vs ambig.}}{\textsc{Auroc} \uparrow}$ }} 
                                                               & LeNet        &     93.13 & 52.96 & 92.76 & 93.10 & \textbf{95.80} & \underline{95.25} & 94.77 \\
                                                               & VGG-11       &     95.17 & 59.96 & 92.06 & 95.03 & \underline{96.81} & \textbf{97.27} & 96.05 \\
                                                               & VGG-16       &     93.47 & 73.31 & 92.22 & 93.39 & \underline{96.57} & \textbf{96.85} & 95.64 \\
                                                               & Resnet-18    &     93.25 & 52.85 & 88.61 & nan & \underline{96.21} & \textbf{96.58} & 95.01 \\
                                                               \hline 
\multirow{4}{*}{\rotatebox[origin=c]{90}{$\underset{\text{MNIST vs OOD}}{\textsc{Auroc} \uparrow}$ }} 
                                                               & LeNet        &      \textbf{97.85} & 84.11 & 96.62 & 95.27 & 85.03 & \underline{96.86} & 96.68 \\
                                                               & VGG-11       &      \underline{98.60} & 97.42 & 96.78 & 96.42 & 97.57 & \textbf{99.04} & 97.53 \\
                                                               & VGG-16       &      \underline{98.01} & 97.78 & 95.62 & 95.00 & 97.63 & \textbf{98.87} & 96.48 \\
                                                               & Resnet-18    &      97.66 & \textbf{99.78} & 93.41 & nan & 99.11 & 99.12 & \underline{99.33} \\
                                                               \hline
\multirow{4}{*}{\rotatebox[origin=c]{90}{$\underset{\text{ambig. vs OOD}}{\textsc{Auroc} \uparrow}$ }} 
                                                               & LeNet        &      80.47 & 83.10 & 72.29 & 81.14 & 81.45 & \underline{85.68} & \textbf{92.68} \\
                                                               & VGG-11       &      81.39 & \textbf{98.22} & 75.63 & 79.34 & 80.77 & 94.59 & \underline{96.64} \\
                                                               & VGG-16       &      79.69 & \textbf{96.43} & 71.94 & 72.10 & 80.28 & 92.03 & \underline{95.54} \\
                                                               & Resnet-18    &      77.41 & \textbf{99.96} & 69.61 & nan & 94.65 & 94.38 & \underline{99.55} \\
                                                               \hline
\end{tabular}
\end{table}

\newpage

\subsection{Feature collapse and repulsion samples} \label{sec:feature_collapse} 
Following our evaluation of the base network architecture, we analyze whether the \acrshort{fLLPOVI} with repulsion samples from the \acrshort{OOD} training set is capable to detect the corresponding \acrshort{OOD} test set. Table \ref{tab:OOD_mnist_oracle} summarizes the results for the DirtyMNIST dataset. We observe that \acrshort{fLLPOVI} with repulsion samples from the \acrshort{OOD} training set is improving the \acrshort{OOD} detection performance without compromising \acrshort{ID} classification accuracy. 
Linear layers with informative repulsion samples are able to detect the corresponding \acrshort{OOD} test set with high accuracy for all network architectures. Especially, the Resnet-18 architecture with bi-Lipschitz constraints that aim to avoid feature collapse, performs significantly better in decomposing aleatoric and epistemic uncertainty. The lack of inherent inductive biases in the base network architecture can be compensated by the repulsion in function space to a certain degree. However, it remains challenging to find such informative repulsion samples prior to training.  
\setlength{\tabcolsep}{7pt} 

\begin{table}[H]
\centering
\caption{\textsc{Auroc} of ambiguous MNIST vs OOD data. Evaluation of epistemic uncertainty for distinguishing ambiguous MNIST from specified OOD datasets. We test the influence of repulsion samples from the \acrshort{OOD} training set for different base network architectures. Informative repulsion samples can improve \acrshort{OOD} detection on base models without an explicitly regularized feature space. Results are averaged over 5 runs.}
\label{tab:OOD_mnist_oracle}
\resizebox{\textwidth}{!}{%
\begin{tabular}{llcccccccc}
\hline
 &
  \textbf{OOD dataset} &
  \multicolumn{1}{l}{MAP} &
  \multicolumn{1}{l}{DDU} &
  \multicolumn{1}{l}{\acrshort{LLPOVI} \textit{(ours)}} &
  \multicolumn{1}{l}{\acrshort{fLLPOVI} \textit{(ours)}} &
   &
   &
   &
   \\
 &
  \textbf{} &
  \multicolumn{1}{l}{} &
  \multicolumn{1}{l}{} &
  \multicolumn{1}{l}{} &
  \multicolumn{1}{l}{+\textit{dirtyMNIST}} &
  \multicolumn{1}{l}{+\textit{eMNIST}} &
  \multicolumn{1}{l}{+\textit{kMNIST}} &
  \multicolumn{1}{l}{+\textit{fashionMNIST}} &
  +\textit{Omniglot} \\ \hline
\multirow{5}{*}{\rotatebox[origin=c]{90}{LeNet}}     & eMNIST       &   60.19 & 84.76 & 78.99 & 78.94 & \underline{87.10} & \textbf{88.17} & 81.52 & 80.39 \\
                                                     & kMNIST       &   72.48 & 86.52 & 86.10 & 86.06 & \underline{91.07} & \textbf{94.31} & 86.08 & 87.79 \\
                                                     & fashionMNIST &   88.61 & 79.84 & 88.96 & 88.85 & 93.48 & \underline{96.45} & \textbf{98.07} & 93.94 \\
                                                     & Omniglot     &   87.76 & 83.38 & 86.77 & 86.69 & 92.69 & \underline{94.14} & 92.00 & \textbf{98.89} \\ \cline{2-10} 
\multirow{5}{*}{\rotatebox[origin=c]{90}{VGG-11}}    & eMNIST       & 63.44 & \textbf{96.51} & 82.60 & 82.64 & \underline{96.49} & 95.50 & 86.42 & 79.62  \\
                                                     & kMNIST       & 76.59 & \underline{99.49} & 93.57 & 93.60 & 98.72 & \textbf{99.85} & 94.13 & 96.89  \\
                                                     & fashionMNIST & 89.38 & 97.84 & 97.11 & 97.12 & 97.53 & \underline{98.16} & \textbf{99.74} & 91.65  \\
                                                     & Omniglot     & 85.56 & 96.99 & 95.00 & 95.03 & 94.04 & \underline{99.26} & 95.23 & \textbf{99.99}  \\ \cline{2-10} 
\multirow{5}{*}{\rotatebox[origin=c]{90}{VGG-16}}    & eMNIST       & 61.13 & 94.09 & 78.70 & 78.71 & \textbf{96.23} & \underline{94.17} & 86.54 & 78.09 \\  
                                                     & kMNIST       & 72.51 & \underline{98.10} & 89.31 & 89.30 & 97.44 & \textbf{99.16} & 91.85 & 92.13 \\  
                                                     & fashionMNIST & 87.37 & 97.32 & 95.35 & 95.35 & \underline{98.11} & 98.09 & \textbf{99.68} & 94.19 \\  
                                                     & Omniglot     & 87.05 & 94.44 & 94.36 & 94.33 & 91.47 & \underline{98.17} & 96.00 & \textbf{98.86} \\  \cline{2-10} 
\multirow{5}{*}{\rotatebox[origin=c]{90}{Resnet-18}} & eMNIST       & 65.39 & \textbf{98.58} & 84.41 & 84.41 & \underline{98.14} & 97.35 & 93.55 & 88.40 \\
                                                     & kMNIST       & 77.08 & \underline{99.92} & 93.32 & 93.33 & 99.15 & \textbf{99.93} & 98.96 & 98.21 \\
                                                     & fashionMNIST & 82.71 & \underline{99.89} & 98.15 & 98.14 & 99.59 & 99.75 & \textbf{99.98} & 98.76 \\
                                                     & Omniglot     & 77.79 & \textbf{100.00} & 97.03 & 97.02 & \underline{99.94} & \textbf{100.00} & \textbf{100.00} & \textbf{100.00} \\ \hline
\end{tabular}}
\end{table}

\newpage

\subsection{Sensitivity to spectral normalization}
Next, we test the sensitivity of our method to spectral normalization. As proposed in SNGP and DDU \cite{liu2023simple,mukhoti2023deep}, residual connections are used to make the base network more sensitive to distances in the input space and spectral normalization is applied to bound the Lipschitz constant of the network. 
We compare the performance of \acrshort{fLLPOVI} with and without spectral normalization on the CIFAR-10 dataset in terms of uncertainty calibration and \acrshort{OOD} detection. 

Table \ref{tab:spectral_normalization} summarizes the results. Similar to ablation studies in DDU \cite{mukhoti2023deep}, we observe little difference in the performance of \acrshort{fLLPOVI} with and without spectral normalization. The experimental results justify the use of post-hoc uncertainty methods on top of pretrained residual networks, even if they are not specifically trained with spectral normalization constraints. 
\setlength{\tabcolsep}{7pt} 

\begin{table}[H]
\centering
\caption{Sensitivity of uncertainty methods to spectral normalization for a Resnet-18 architecture. Results are averaged over 5 runs.}
\label{tab:spectral_normalization}
\resizebox{\textwidth}{!}{%
\begin{tabular}{llcccccccc}
\hline
\multirow{2}{*}{\textbf{Method}} &
  \multirow{2}{*}{\textbf{SN}} &
  \multirow{2}{*}{\textsc{Acc.} $\uparrow$ [\%]} &
  \multirow{2}{*}{\textsc{ECE} $\downarrow$ [\%]} &
  \multicolumn{6}{c}{\textsc{Auroc} $\uparrow$ [\%]} \\ \cline{5-10} 
 &
   &
   &
   &
  \textit{Cifar100} &
  \multicolumn{1}{c}{\textit{TinyIm.}} &
  Places365 &
  \textit{Texture} &
  \textit{SVHN} &
  FakeData \\ \hline
\multirow{4}{*}{MAP}                         & \xmark                 & 95.04  &    2.58      &        88.39 & 88.29 & 89.09 & 90.39 & 94.51 & 94.34 \\
                                             & \cmark (coeff. = 3.0)  & 95.06  &    2.60      &        88.43 & 88.25 & 89.21 & 90.23 & 94.10 & 94.30 \\
                                             & \cmark (coeff. = 7.0)  & 95.05  &    2.61      &        88.49 & 88.31 & 89.21 & 90.69 & 94.12 & 94.15 \\ \hline
\multirow{4}{*}{DDU}                         & \xmark                 & 95.04  &     2.58      &         88.96 & 89.23 & 90.70 & 95.62 & 93.61 & 99.99 \\ 
                                             & \cmark (coeff. = 3.0)  & 95.06  &     2.60      &         89.03 & 89.23 & 90.66 & 95.56 & 94.59 & 99.99 \\ 
                                             & \cmark (coeff. = 7.0)  & 95.05  &     2.61      &         88.96 & 89.23 & 90.58 & 95.68 & 94.81 & 99.98 \\ \hline
\multirow{4}{*}{LL-Laplace}                  & \xmark                 & 95.03   &  1.03  &     88.14 & 88.27 & 89.13 & 90.86 & 93.04 & 95.74  \\
                                             & \cmark (coeff. = 3.0)  &   95.02    &   1.09    &     87.50 & 87.59 & 88.58 & 89.98 & 92.43 & 94.01  \\
                                             & \cmark (coeff. = 7.0)  &   95.02    &   1.04    &     88.28 & 88.37 & 89.21 & 91.14 & 93.14 & 95.31  \\ \hline
\multirow{4}{*}{SNGP      }                  & \xmark                 & 95.02  &    1.25       &     86.47 & 83.90 & 92.32 & 98.18 & 87.03 & 88.59   \\
                                             & \cmark (coeff. = 3.0)  & 95.04  &    0.96       &     88.42 & 87.97 & 91.28 & 96.05 & 94.23 & 94.09   \\
                                             & \cmark (coeff. = 7.0)  & 95.02  &    1.25       &     86.49 & 83.89 & 92.47 & 98.17 & 86.31 & 87.39   \\ \hline
\multirow{4}{*}{\acrshort{LLPOVI} \emph{(ours)}}  & \xmark                 &  95.05  &  1.81   &    88.73 & 88.97 & 90.04 & 91.41 & 93.58 & 95.76  \\
                                             & \cmark (coeff. = 3.0)  &  95.05  &  1.84   &    88.82 & 88.97 & 90.30 & 91.19 & 93.12 & 95.38  \\
                                             & \cmark (coeff. = 7.0)  &  95.06  &  1.83   &    88.85 & 88.98 & 90.23 & 91.56 & 93.65 & 96.32  \\ \cline{2-10} 
    \multirow{4}{*}{$\underset{+\textit{Cifar100}}{\text{\acrshort{fLLPOVI}}}$ \emph{(ours)}} & \xmark                 &  94.99  &   1.94  &   91.24 & 91.42 & 91.89 & 94.80 & 96.29 & 98.08   \\
                                                                                         & \cmark (coeff. = 3.0)  &  95.00  &   2.02  &   91.27 & 91.38 & 91.83 & 94.63 & 96.30 & 98.03   \\
                                                                                         & \cmark (coeff. = 7.0)  &  95.02 &    2.03 &   91.27 & 91.39 & 91.74 & 95.00 & 96.61 & 97.28   \\ \cline{2-10} 
\multirow{4}{*}{$\underset{+\textit{TinyImagenet}}{\text{\acrshort{fLLPOVI}}}$ \emph{(ours)}} & \xmark                 &  95.00 &   1.10  &  87.38 & 89.31 & 90.18 & 93.94 & 91.52 & 99.74 \\
                                                                                         & \cmark (coeff. = 3.0)  &  95.02 &   1.14  &  88.12 & 89.66 & 90.60 & 93.94 & 92.41 & 99.11 \\
                                                                                         & \cmark (coeff. = 7.0)  &  95.03 &   1.13  &  87.70 & 89.47 & 90.16 & 94.14 & 92.50 & 99.43 \\ \hline
\multirow{4}{*}{DE-5}   & \xmark                 & 95.98  & 0.59  &    90.54 & 90.00 & 90.15 & 93.49 & 94.06 & 99.15 \\
                        & \cmark (coeff. = 3.0)  & 96.02  & 0.52  &    90.83 & 90.25 & 90.38 & 93.53 & 95.02 & 99.26 \\
                        & \cmark (coeff. = 7.0)  & 95.96  & 0.55  &    90.52 & 90.00 & 90.20 & 93.45 & 93.98 & 98.94 \\ \hline
\end{tabular}}
\end{table}

\newpage

\subsection{Sensitivity to base network architecture}\label{sec:base_network_size}
We evaluate the performance of \acrshort{fLLPOVI} for various base network architectures with varying feature space dimensionality on the CIFAR-10 dataset. For each base network, we evaluate the performance of \acrshort{fLLPOVI} with repulsion samples from the \acrshort{OOD} training set in Table \ref{tab:base_network_size}. \\ 
All networks analyzed in this section include residual connections and exhibit sensitivity to distances in the input space. Results are consistent across networks. Importantly, retraining the last layer with function space regularization improves the calibration of the \acrshort{ID} uncertainty and improves the epistemic uncertainty estimates for the \acrshort{OOD} detection.
In Table \ref{tab:OOD:cifar10_oracle} we analyze the effect of using different repulsion samples on the \acrshort{OOD} detection capabilities. As in Section \ref{sec:feature_collapse} we test whether \acrshort{fLLPOVI} with repulsion samples from the \acrshort{OOD} training set is capable to detect the corresponding \acrshort{OOD} test set. Additionally, we can observe whether diversity on specific \acrshort{OOD} datasets generalizes to other \acrshort{OOD} datasets. In most cases, the prediction diversity of linear layers is sufficiently flexible and can compete with or outperform the density-based DDU method. 
However, we observe that diversity not not generalize in all cases. While diversity on the \textit{Texture} dataset improves \acrshort{OOD} detection on the \textit{SVHN} and \textit{FakeData} datasets, it does not apply vice versa. 
\setlength{\tabcolsep}{7pt} 

\begin{table}[H]
\centering
\caption{Influence of the base network architecture on the performance of single model uncertainty methods and deep ensembles. Results are averaged over 5 runs.}
\label{tab:base_network_size}
\resizebox{\textwidth}{!}{%
\begin{tabular}{llcccccccc}
\hline
\multirow{2}{*}{\textbf{Method}} &
  \multirow{2}{*}{\textbf{Base network}} &
  \multirow{2}{*}{\textsc{Acc.} $\uparrow$ [\%]} &
  \multirow{2}{*}{\textsc{ECE} $\downarrow$ [\%]} &
  \multicolumn{6}{c}{\textsc{Auroc} $\uparrow$ [\%]} \\ \cline{5-10} 
 &
   &
   &
   &
  \textit{Cifar100} &
  \multicolumn{1}{c}{\textit{TinyIm.}} &
  Places365 &
  \textit{Texture} &
  \textit{SVHN} &
  FakeData \\ \hline
\multirow{4}{*}{MAP}                         & Resnet-18                & 95.06  & 2.60  &  88.43 & 88.25 & 89.21 & 90.23 & 94.10 & 94.30   \\
                                             & Wide-Resnet-28-10        & 95.91  & 2.33  &  89.71 & 89.25 & 90.00 & 89.21 & 93.46 & 97.06   \\
                                             & Resnet-50                & 94.99  & 2.85  &  87.98 & 87.59 & 88.18 & 90.38 & 92.17 & 95.52   \\
                                             & Resnet-101               & 94.82  & 2.80  &  88.15 & 87.68 & 88.35 & 90.24 & 93.97 & 96.33   \\ \hline
\multirow{4}{*}{DDU}                         & Resnet-18                & 95.06  & 2.60  &  89.03 & 89.23 & 90.66 & 95.56 & 94.59 & 99.99   \\
                                             & Wide-Resnet-28-10        & 95.91  & 2.33  &  89.00 & 89.14 & 90.81 & 97.62 & 98.62 & 100.00   \\
                                             & Resnet-50                & 94.99  & 2.85  &  87.35 & 87.55 & 89.09 & 95.23 & 93.92 & 100.00   \\
                                             & Resnet-101               & 94.82  & 2.80  &  86.92 & 87.03 & 88.57 & 94.98 & 94.26 & 100.00   \\ \hline
\multirow{4}{*}{LL-Laplace}                  & Resnet-18                & 95.02  & 1.09  & 87.50 & 87.59 & 88.58 & 89.98 & 92.43 & 94.01  \\
                                             & Wide-Resnet-28-10        & 95.89  & 1.13  & 88.77 & 88.57 & 89.63 & 89.15 & 92.30 & 96.74  \\
                                             & Resnet-50                & 94.94  & 1.39  & 87.46 & 87.25 & 88.01 & 90.10 & 90.88 & 96.16  \\
                                             \hline
\multirow{4}{*}{SNGP}                        & Resnet-18                & 95.04  & 0.96  &  88.42 & 87.97 & 91.28 & 96.05 & 94.23 & 94.09   \\
                                             & Wide-Resnet-28-10        & 95.90  & 1.25  &  89.44 & 88.73 & 91.85 & 95.30 & 93.15 & 97.13   \\
                                             & Resnet-50                & 94.99  & 1.34  &  87.36 & 86.86 & 90.52 & 95.82 & 91.22 & 95.52   \\
                                             & Resnet-101               & 94.80  & 1.06  &  87.99 & 87.34 & 90.76 & 95.77 & 93.79 & 96.50  \\ \hline
\multirow{4}{*}{\acrshort{LLPOVI} \emph{(ours)}}  & Resnet-18                & 95.05  & 1.84  &  88.82 & 88.97 & 90.30 & 91.19 & 93.12 & 95.38   \\
                                             & Wide-Resnet-28-10        & 95.88  & 0.97  &  89.72 & 89.64 & 91.02 & 89.87 & 92.39 & 97.77   \\
                                             & Resnet-50                & 95.01  & 2.26  &  88.33 & 88.23 & 89.32 & 91.43 & 92.48 & 96.98   \\
                                             & Resnet-101               & 94.83  & 2.16  &  88.61 & 88.42 & 89.77 & 91.50 & 94.00 & 98.37   \\ \cline{2-10} 
\multirow{4}{*}{$\underset{+\textit{TinyImagenet}}{\text{\acrshort{fLLPOVI}}}$ \emph{(ours)}}
                                             & Resnet-18                & 95.02  & 1.14  &  88.12 & 89.66 & 90.60 & 93.94 & 92.41 & 99.11    \\
                                             & Wide-Resnet-28-10        & 95.83  & 1.62  &  89.51 & 90.40 & 91.57 & 94.24 & 94.27 & 99.62    \\
                                             & Resnet-50                & 94.95  & 1.49  &  89.58 & 90.34 & 91.51 & 94.77 & 94.96 & 99.92    \\
                                             & Resnet-101               & 94.77  & 1.46  &  89.73 & 90.60 & 91.77 & 94.89 & 95.42 & 99.98    \\ \hline
\multirow{4}{*}{DE-5      }                  & Resnet-18                & 95.96  & 0.59  & 90.59 & 90.00 & 90.17 & 93.22 & 94.85 & 98.94    \\
                                             & Wide-Resnet-28-10        & 96.59  & 0.77  & 91.81 & 90.84 & 91.09 & 94.54 & 97.24 & 99.84    \\
                                             & Resnet-50                & 96.12  & 0.75  & 90.32 & 89.53 & 89.60 & 93.78 & 94.37 & 99.99    \\
                                             & Resnet-101               & 96.00  & 0.71  & 89.95 & 89.17 & 89.21 & 93.09 & 94.98 & 99.14    \\ \hline
\end{tabular}
}
\end{table}

\clearpage

\begin{table}[H]
\centering
\caption{\textsc{Auroc} of Cifar10 vs OOD data. Function space uncertainty methods with repulsion samples from the \acrshort{OOD} training set for different base network architectures. Results are averaged over 5 runs.}
\label{tab:OOD:cifar10_oracle}
\begin{tabular}{llcccccccc}
\hline
 &
  \textbf{OOD dataset} &
  \multicolumn{1}{l}{MAP} &
  \multicolumn{1}{l}{DDU} &
  \multicolumn{1}{l}{\acrshort{LLPOVI}} &
  \multicolumn{1}{l}{\acrshort{fLLPOVI}} &
   &
   &
   &
  \multicolumn{1}{l}{} \\
 &
  \textbf{} &
  \multicolumn{1}{l}{} &
  \multicolumn{1}{l}{} &
  \multicolumn{1}{l}{} &
  \multicolumn{1}{l}{+\textit{Cifar100}} &
  \multicolumn{1}{l}{+\textit{TinyIm.}} &
  \multicolumn{1}{l}{+\textit{Texture}} &
  \multicolumn{1}{l}{+\textit{SVHN}} &
  \multicolumn{1}{l}{+\textit{FakeData}} \\ \hline
\multirow{6}{*}{\rotatebox[origin=c]{90}{Resnet-18}}        & \textit{Cifar100}  & 88.43 & 89.03 & 88.82 & \textbf{91.27} & 88.12 & \underline{89.37} & 87.08 & 85.72 \\
                                                            & \textit{TinyIm.}   & 88.25 & 89.23 & 88.97 & \textbf{91.38} & 89.66 & \underline{89.88} & 85.74 & 86.28 \\
                                                            & \textit{Places365} & 89.21 & \underline{90.66} & 90.30 & \textbf{91.83} & 90.60 & 88.54 & 84.60 & 84.95 \\
                                                            & \textit{Texture}   & 90.23 & \underline{95.56} & 91.19 & 94.63 & 93.94 & \textbf{95.72} & 90.53 & 90.98 \\
                                                            & \textit{SVHN}      & 94.10 & 94.59 & 93.12 & \underline{96.30} & 92.41 & 95.94 & \textbf{99.09} & 89.72 \\
                                                            & \textit{FakeData}  & 94.30 & \underline{99.99} & 95.38 & 98.03 & 99.11 & 98.21 & 89.74 & \textbf{100.00} \\ \hline
\multirow{6}{*}{\rotatebox[origin=c]{90}{WideResnet-28-10}} & \textit{Cifar100}  & 89.71 & 89.00 & \underline{89.72} & \textbf{91.93} & 89.51 & 89.42 & 86.83 & 85.67 \\
                                                            & \textit{TinyIm.}   & 89.25 & 89.14 & 89.64 & \textbf{91.46} & \underline{90.40} & 89.82 & 85.06 & 86.33 \\
                                                            & \textit{Places365} & 90.00 & 90.81 & 91.02 & \underline{91.45} & \textbf{91.57} & 89.39 & 83.31 & 85.74 \\
                                                            & \textit{Texture}   & 89.21 & \textbf{97.62} & 89.87 & 94.06 & 94.24 & \underline{95.66} & 88.77 & 90.47 \\
                                                            & \textit{SVHN}      & 93.46 & \underline{98.62} & 92.39 & 96.34 & 94.27 & 95.55 & \textbf{99.13} & 89.09 \\
                                                            & \textit{FakeData}  & 97.06 & \textbf{100.00} & 97.77 & 98.45 & \underline{99.62} & 98.49 & 93.11 & \textbf{100.00} \\ \hline
\multirow{6}{*}{\rotatebox[origin=c]{90}{Resnet-50}}        & \textit{Cifar100}  & 87.98 & 87.35 & 88.33 & \textbf{90.96} & \underline{89.58} & 89.11 & 85.72 & 86.03 \\
                                                            & \textit{TinyIm.}   & 87.59 & 87.55 & 88.23 & \textbf{90.83} & \underline{90.34} & 89.51 & 84.30 & 85.81 \\
                                                            & \textit{Places365} & 88.18 & 89.09 & 89.32 & \underline{91.35} & \textbf{91.51} & 88.64 & 82.73 & 84.33 \\
                                                            & \textit{Texture}   & 90.38 & \underline{95.23} & 91.43 & 93.84 & 94.77 & \textbf{96.03} & 89.74 & 90.79 \\
                                                            & \textit{SVHN}      & 92.17 & 93.92 & 92.48 & 95.16 & 94.96 & \underline{96.17} & \textbf{99.41} & 89.97 \\
                                                            & \textit{FakeData}  & 95.52 & \textbf{100.00} & 96.98 & 98.37 & \underline{99.92} & 99.43 & 93.49 & \textbf{100.00} \\ \hline
\multirow{6}{*}{\rotatebox[origin=c]{90}{Resnet-101}}       & \textit{Cifar100}  & 88.15 & 86.92 & 88.61 & \textbf{91.00} & \underline{89.82} & 89.47 & 86.95 & 86.79 \\
                                                            & \textit{TinyIm.}   & 87.68 & 87.03 & 88.42 & \textbf{90.75} & \underline{90.69} & 89.76 & 85.23 & 86.63 \\
                                                            & \textit{Places365} & 88.35 & 88.57 & 89.77 & \underline{91.26} & \textbf{91.91} & 89.06 & 84.83 & 85.18 \\
                                                            & \textit{Texture}   & 90.24 & \underline{94.98} & 91.50 & 93.69 & 94.77 & \textbf{96.26} & 90.49 & 91.89 \\
                                                            & \textit{SVHN}      & 93.97 & 94.26 & 94.00 & 96.06 & 95.39 & \underline{96.49} & \textbf{99.57} & 92.07 \\
                                                            & \textit{FakeData}  & 96.33 & \textbf{100.00} & 98.37 & 98.77 & \underline{99.98} & 99.66 & 94.23 & \textbf{100.00} \\ \hline
\end{tabular}
\end{table}

\end{document}